%% file: main.tex
\pdfoutput=1
\documentclass[10pt,twocolumn,letterpaper]{article}
\usepackage[pagenumbers]{cvpr}

\input{preamble}

\usepackage{booktabs}
\usepackage{multirow}
\usepackage{graphicx}
\usepackage{amsmath,amssymb,amsthm}
\newtheorem{proposition}{Proposition}
\usepackage{xspace}
\usepackage{algorithm}
\usepackage{algpseudocode}
\usepackage{caption}
\usepackage{microtype}
\definecolor{cvprblue}{rgb}{0.21,0.49,0.74}
\usepackage[pagebackref,breaklinks,colorlinks,allcolors=cvprblue]{hyperref}
\hypersetup{pdftitle={Search or Scoring? Diagnosing Why Visual World-Model Planners Fail}}

\newcommand{\W}{\textsc{obj}\xspace}
\newcommand{\J}{\textsc{full}\xspace}
\newcommand{\DProp}{DemoProp\xspace}
\newcommand{\DPSeed}{warm-start CEM\xspace}
\newcommand{\DPExp}{Expand\xspace}
\newcommand{\DPRes}{Restrict\xspace}
\newcommand{\DPAvg}{Average\xspace}

\title{Search or Scoring?\\Diagnosing Why Visual World-Model Planners Fail}
\author{%
Haoxuan Jia$^{1}$\hspace{0.45em}%
Yang Liu$^{2}$\hspace{0.45em}%
Shang Luo$^{2}$\hspace{0.45em}%
Chenhao Wu$^{3}$\hspace{0.45em}%
Zixiang Ni$^{4}$\hspace{0.45em}%
Yingguang Yang$^{2}$\hspace{0.45em}%
Feng Zhang$^{5}$\\
Chongyang Zhang$^{6}$\hspace{0.45em}%
Hao Zheng$^{6}$\hspace{0.45em}%
Kefu Xu$^{2}$\hspace{0.45em}%
Congjing Ran$^{7}$\hspace{0.45em}%
Jifeng Zhu$^{8}$\hspace{0.45em}%
Bin Chong$^{2,*}$\\[0.6em]
{\normalsize $^{1}$Nanyang Technological University\quad
$^{2}$Peking University\quad
$^{3}$University of Chinese Academy of Sciences}\\
{\normalsize $^{4}$Xi'an Jiaotong University\quad
$^{5}$Fuzhou University\quad
$^{6}$Fullive-AI\quad
$^{7}$Wuhan University}\\
{\normalsize $^{8}$National University of Singapore}\\[0.4em]
{\small $^{*}$Corresponding author: \texttt{chongbin@pku.edu.cn}}%
}

\begin{document}
\twocolumn[{%
\renewcommand\twocolumn[1][]{#1}%
\maketitle
\vspace{-1.2em}
\begin{center}
\includegraphics[width=\linewidth]{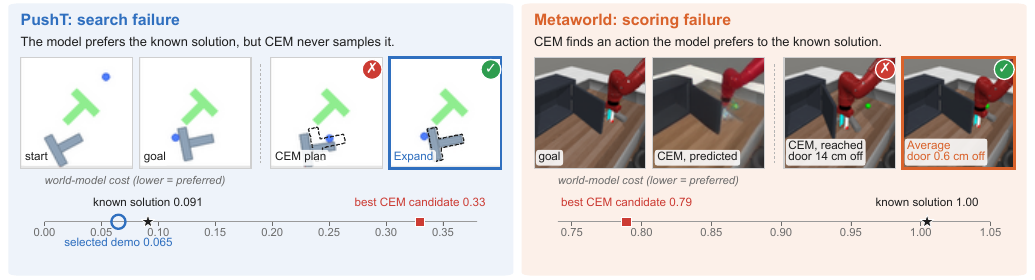}
\captionof{figure}{\textbf{Two ways a world-model planner fails.} Left, a search failure on PushT
(development task): the model's cost for the known solution (0.091) is far below that of the best of the 900 candidates recorded from the
cross-entropy method (CEM; 0.33), and adding retrieved demonstrations to the candidates (Expand) lets the cost
pick one that works. Right, a scoring failure on Metaworld door-open: the model assigns CEM's action a lower cost than the known solution and predicts
it to open the door (decoded prediction), yet the door stays half open;
Average, the mean of the 30 nearest demonstration windows, opens it. Dashed outline: the block's goal pose.
\checkmark/$\times$: full success.}
\label{fig:teaser}
\end{center}
\vspace{0.6em}
}]
\input{sections/abstract}
\input{sections/introduction}
\input{sections/related}
\input{sections/setup}
\input{sections/diagnosis}
\input{sections/method}
\input{sections/experiments}
\input{sections/predict}
\input{sections/discussion}

\section*{LLM Usage Statement}
\input{sections/llm_statement}

{\small
\bibliographystyle{ieeenat_fullname}
\bibliography{references}
}

\clearpage
\setcounter{section}{0}\setcounter{figure}{0}\setcounter{table}{0}\setcounter{proposition}{0}
\renewcommand{\theproposition}{S\arabic{proposition}}
\renewcommand{\thesection}{S\arabic{section}}
\renewcommand{\thetable}{S\arabic{table}}
\renewcommand{\thefigure}{S\arabic{figure}}
\renewcommand{\topfraction}{0.95}\renewcommand{\bottomfraction}{0.95}\renewcommand{\textfraction}{0.05}
\renewcommand{\floatpagefraction}{0.85}\renewcommand{\dbltopfraction}{0.95}\renewcommand{\dblfloatpagefraction}{0.85}
\def\theHsection{S\arabic{section}}\def\theHfigure{S\arabic{figure}}
\def\theHtable{S\arabic{table}}\def\theHproposition{S\arabic{proposition}}
\twocolumn[{\centering{\Large\bf Search or Scoring? Supplementary Material\par}\vspace{2.2em}}]
\section{DemoProp pseudocode}
\label{sec:supp-alg}
\begin{algorithm}[h]
\caption{\DProp with the known-solution probe}
\label{alg:demoprop}
\small
\begin{algorithmic}[1]
\Require cost $c$, CEM, demonstration indices $\mathcal{D}_{\textsc{Warm}}$ and $\mathcal{D}_{\textsc{Avg}}$, development tasks with known solutions
\Statex \quad ($\mathcal{D}_{\textsc{Warm}}=\mathcal{D}_{\textsc{Avg}}$ except on Metaworld: the task's own demonstrations for the warm start
\Statex \quad and the controls, those of all 42 tasks for the average; the matched controls of \cref{sec:supp-crossenv} use $\mathcal{D}_{\textsc{Avg}}$ for all)
\State \textbf{Probe:} on each development failure, compare $c$(known solution) with the cheapest of all candidates CEM evaluated
\State $s \gets$ share of development tasks CEM fails with the scoring label; $p_{\mathrm{avg}} \gets$ share the 30-window average solves
\State use $\gets$ \textsc{Warm} if $s<1-p_{\mathrm{avg}}$, else \textsc{Average}
\For{each planning step}
  \State $\mathcal{P}_{\mathrm{demo}} \gets K$ windows of $\mathcal{D}_{\text{use}}$ nearest to the key of (current, goal) proprioception
  \If{use $=$ \textsc{Warm}}
    \State warm start: $a^\star \gets$ the output of CEM with $\mathcal{P}_{\mathrm{demo}}$ in its first population
  \Else
    \State Average: $a^\star \gets$ mean of the 30 windows of $\mathcal{P}_{\mathrm{demo}}$ nearest in key (no cost)
  \EndIf
  \State execute $a^\star$
\EndFor
\State \textbf{Controls} (the first protocols): Expand, $a^\star \gets \arg\min c$ over the final CEM population $\cup\ \mathcal{P}_{\mathrm{demo}}$
\State \quad (in closed loop at every replanning step, CEM's output unless a window is the cheapest);
\State \quad Restrict, $a^\star \gets \arg\min c$ over $\mathcal{P}_{\mathrm{demo}}$
\end{algorithmic}
\end{algorithm}

\section{Endpoint-constrained CEM}
\label{sec:supp-endpoint}
The pusher of PushT tracks its commanded target through a fixed second-order
proportional--derivative response, so after the 30 simulator steps of a plan its
position is an affine function of the plan: with $z\in\mathbb{R}^{60}$ the
normalized action sequence and $s_0$ the pusher's start position and velocity,
$p_{30} = A z + b(s_0)$ with a known $A\in\mathbb{R}^{2\times 60}$. Endpoint-constrained
CEM samples only plans with $A z = p_g - b(s_0)$, where $p_g$ is the goal
proprioception that the planner already receives. It uses a fixed orthonormal
basis $N\in\mathbb{R}^{60\times58}$ of the null space of $A$ (Householder
construction, independent of the task), writes $z = z^\star + N u$ with a
particular solution $z^\star$, and runs the official diagonal-Gaussian CEM on the
58 intrinsic coordinates $u$ with the same random draws, objective, elite count
and budget ($300\times30$). Every sample therefore ends with the pusher exactly
at its goal position; the latent objective still decides where the block goes.
This variant is a strong baseline for the full-success criterion, which also
requires the pusher position; it is a baseline, not one of \DProp's uses.

\section{Tasks, splits and protocols}
\label{sec:supp-tasks}
\begin{table*}[tp]
\centering\footnotesize
\setlength{\tabcolsep}{3pt}
\begin{tabular}{@{}p{0.075\textwidth}p{0.10\textwidth}p{0.15\textwidth}p{0.15\textwidth}p{0.165\textwidth}p{0.11\textwidth}p{0.065\textwidth}p{0.085\textwidth}@{}}
\toprule
Setting & World models & Official planner & Tasks & Goal and known solution & Success: \W\ / \J & Search failures & Use that works\tabularnewline
\midrule
\raggedright PushT & \raggedright DINO-WM, JEPA-WM & \raggedright CEM $300{\times}30$, 10 elites, horizon 6, frame skip 5 & \raggedright 96 official episodes from the 21 validation trajectories & \raggedright replayed 30-step window of a demonstration; its actions & \raggedright block within 20 units and $20^{\circ}$ / also the pusher & \raggedright 29 of 34, 23 of 28 & \raggedright warm start (first protocols: \DPExp)\tabularnewline
\raggedright Wall & \raggedright DINO-WM & \raggedright CEM $300{\times}30$, 10 elites, horizon 6, frame skip 5 & \raggedright 64 tasks, goals sampled by the environment & \raggedright scripted path through the door & \raggedright final distance below 4.5 (one criterion) & \raggedright 19 of 25 & \raggedright warm start (first protocol: \DPExp)\tabularnewline
\raggedright Metaworld & \raggedright DINO-WM, JEPA-WM & \raggedright CEM $300{\times}15$, 10 elites, horizon 6, frame skip 5 & \raggedright 112 window tasks (7 tasks, 16 episodes each); \DPAvg: 168 new tasks (4 per family, all 42) & \raggedright 30-step window of the expert's replay in which the object moves most; the expert's actions & \raggedright object within 5\,cm / also the hand & \raggedright 0 of 67, 0 of 85 & \raggedright \DPAvg (first protocols: \DPRes)\tabularnewline
\raggedright RoboCasa & \raggedright DINO-WM, V-JEPA-2-AC (trained on DROID) & \raggedright CEM $300{\times}15$, 10 elites, horizon 3, latent distance only & \raggedright 90 window tasks (15 steps, reach and pick phases of 14 demonstrations) & \raggedright demonstration window converted to planner actions & \raggedright HAND: hand within 5\,cm / FULL: also the object & \raggedright 0 of 15, 0 of 10 & \raggedright \DPAvg zero-shot; or adapt the predictor\tabularnewline
\bottomrule
\end{tabular}
\caption{\textbf{Settings at a glance} (main evaluations; one plan executed open loop per task). Search failures:
failures of the official CEM output in \J\ (Wall: of CEM's cheapest final candidate) in which the known solution
costs less than every candidate CEM evaluated (Wall: than its final population), per model. Warm-start CEM, which needs no
diagnosis, matches or beats CEM in every setting except zero-shot RoboCasa with V-JEPA-2-AC. On RoboCasa
the two criteria are the hand alone and the hand with the object.}
\label{tab:supp-settings}
\end{table*}
\Cref{tab:supp-settings} summarizes the four settings.

\paragraph{Order of the work.} The study proceeded in rounds, and the protocols of
\cref{tab:supp-protocols} record each one. (i) 2026-10-29/30: the probe on PushT development tasks, and
Expand and Restrict as the registered uses, evaluated on held-out PushT tasks, Wall and the first
Metaworld window tasks. (ii) 10-31: the official PushT episodes, the Metaworld decomposition
into predicted and real-outcome costs, and the warm start as a baseline, which beat our
predictions (M1, M2). (iii) 11-01 to 11-05: RoboCasa, the goal term of the key, more compute,
iCEM, planner seeds and image keys, mostly on reused evaluation tasks. (iv) 11-07/08: the
model's choice among all tasks' windows held on 112 new Metaworld tasks and failed on the
first 168 tasks of all 42 families, which then served as development tasks on which the
average was found (\cref{sec:supp-sprint}). (v) 11-08 to 11-10, tests on new tasks: the
average on 168 newly generated tasks, the within-PushT test on 96 new episodes; on reused
evaluation tasks: the cross-environment protocol (PushT official episodes, Wall, RoboCasa),
the policy reference and the start-key average. (vi) 11-12: matched controls on the 168 new
tasks (\cref{tab:supp-matched}) and the failure-level test on the 96 new episodes of the
within-PushT test (\cref{sec:supp-flip}). The regime of Wall and RoboCasa, which have no
development tasks, was read from their evaluation tasks before the cross-environment
protocol. The rule $s<1-p_{\mathrm{avg}}$ was written in three steps on development data,
and DINO-WM's tie at 5 bits was resolved after that setting's development outcomes were
seen (\cref{sec:supp-flip}).

\emph{What confirms what.} Tasks or episodes never used before confirm: Expand on the
official PushT episodes (both models never trained on their goals); \DPRes and the average
on the new Metaworld tasks, with the matched controls on the same tasks; the flip of the
better use on the new PushT episodes. The failure-level test reran CEM on those episodes, where
the outcomes of the warm start and the average were known; only CEM's outcomes and labels were
new. New interventions on
reused evaluation tasks: the cross-environment test (PushT official episodes, Wall,
RoboCasa), more compute, iCEM, image keys and the start-key tests. Development results:
the probe on the 64 PushT development tasks, the first 168 tasks of all 42 Metaworld
families (on which the average was found), and the rule. The directional predictions of
\cref{tab:supp-ledger} are not independent tests: many share tasks, models and
comparisons. The analyses by success criterion (\cref{tab:supp-criteria-mw,tab:supp-flip-criteria}), the
choice from more candidates of one source (\cref{tab:supp-pool}), the labels under other references
(\cref{sec:supp-diag,sec:supp-robocasa,sec:supp-flip}), the cluster intervals of the cross-environment test
(\cref{sec:supp-crossenv}) and the breakdown of the RoboCasa tasks (\cref{sec:supp-robocasa}) were computed
afterwards from the frozen records; they qualify the confirmations and are not tests.

\paragraph{Intervals.} Where tasks share goal trajectories (PushT), task families (Metaworld) or
demonstrations (RoboCasa), we resample these clusters with replacement 10{,}000 times (seed 0) and report
the 2.5th and 97.5th percentiles of the paired difference (successes of one use minus those of the other)
summed over the resampled clusters; where both models enter one statistic, they share each resample.
Clusters differ in size, so a resample need not hold the original number of tasks and a bound can exceed it;
the interval describes the uncertainty of the count. Where tasks are independent (Wall) we use exact
McNemar tests or resample tasks.
Each task is a 30-step window (offset $o$, actions $o$ to $o{+}29$) of one
trajectory of the public PushT training data used to train DINO-WM. The official
\emph{replay-expert} procedure restores the window's start state and executes the
demonstrated actions; the resulting frame, proprioception and state are the goal.
The public data contain 185 demonstration sources with 101 noise variants each.
129 sources were partitioned once, with a fixed seed and before any outcome was
seen, into 64 development tasks and two held-out sets of 32 and 33 tasks, H1 and H2
(called validation and test in the protocol files; one window per source); all
variants of a source share its split. The development tasks were used for every
analysis in \cref{sec:supp-diag} and for every design decision; H1 and H2 were used
only for frozen evaluations (\cref{tab:supp-protocols}).

\paragraph{Official evaluation episodes.}
The official PushT evaluation configurations of JEPA-WMs (one per model, with
identical planner and task settings) draw 96 episodes from the
validation split (21 trajectories, 76--167 steps; the closest start of any training
trajectory is 42 units away) on 8 processes of 12 episodes each. Process $r$ seeds
its sampler with $1{+}6000r$ and draws, per episode, a trajectory and an offset; the
goal is the replay of the segment's 30 actions. We reproduce these draws in a single
process, running the episodes of each process in order with one agent so that its
generators advance as in the official run; calling the official sampler with the same
generators returns all 96 (trajectory, offset) pairs, 91 of them distinct, covering
all 21 trajectories. Before freezing the protocol we checked on two public training
windows that the same pipeline reproduces the earlier frozen runs: CEM's final
population (same seed) within $1.5{\times}10^{-5}$ relative cost, the costs of saved
proposals within $1.1{\times}10^{-5}$, for both models; the known solution reaches the
goal state exactly in all 192 episodes. Resetting the PushT simulator to a state runs
one physics step, so the planner's start differs from the stored dataset state by the
pusher's velocity step (about one unit); retrieval uses the planner's own start and
goal proprioception. The index holds every 30-step window (stride 2) of the full
public training split, since no evaluation trajectory is in it. The probe compares
the known solution with the cheapest of all 9{,}000 candidates CEM evaluated. Five
baselines run on the same episodes with the same objective: CEM with four times the
population (1{,}200 samples); MPPI with 300 samples, 30 iterations, 64 elites and
temperature 0.5 (class defaults otherwise, as no PushT setting is released); Nevergrad
(NGOpt, $300\times30$) and gradient descent (500 steps, learning rate 1, action noise
0.003) with the released PushT settings; and CEM whose first population starts from the
retrieved windows. The first carries no prediction; for
the others, P5 predicts that \DPExp wins at least as often as it loses
(\cref{tab:supp-ledger}). The last was registered as a baseline, warm-start CEM, and is now \DProp's use in the search regime;
some later protocols call it DemoProp-Seed.

\begin{table*}[tp]
\centering\footnotesize
\begin{tabular}{p{0.2\textwidth}p{0.47\textwidth}p{0.08\textwidth}p{0.14\textwidth}}
\toprule
Protocol & Content & Frozen & SHA-256 (prefix)\\
\midrule
Method (DINO-WM) & proposal rule, $K{=}100$, key, index, pass criterion for opening H2; H1 tasks & 15:13 & \texttt{99fa2d2b}\\
H2 confirmation & the 33 tasks of H2 opened once, same method and gate & 15:50 & \texttt{16e1b7b8}\\
JEPA-WM replication & same tasks, seeds and proposals, JEPA-WM cost & 17:20 & \texttt{d133f8be}\\
Closed-loop MPC & receding horizon, same rule at every replanning step & 17:28 & \texttt{5fc23522}\\
Search baselines & best-of-2/4 CEM restarts, Gaussian control & 17:39 & \texttt{603ae48a}\\
Wall & 64 new seeds, position-only key & 18:28 & \texttt{7b575c6e}\\
Runtime amendment & unfinished MPC, search and Wall tasks on the replacement machine; rule for finished and unfinished tasks & 20:54 & \texttt{ac00a87c}\\
Metaworld & seven tasks, window and official protocols, rules, controls and predictions & 12:01$^{\ast}$ & \texttt{b5944d1e}\\
Metaworld, JEPA-WM & replication on the same 112 window tasks, same rules and predictions & 17:09$^{\ast}$ & \texttt{9d1bc729}\\
Official PushT episodes & 96 official episodes, both models; probe, \DPExp, \DPRes, nearest window and five search baselines; predictions P1--P7 & 16:46$^{\dagger}$ & \texttt{96e21957}\\
Wall probe & scripted known solutions for the 64 Wall tasks; prediction W1 & 16:52$^{\dagger}$ & \texttt{15a90414}\\
Metaworld decomposition & predicted vs real-outcome cost on the 112 DINO-WM window tasks; D1, D2 & 17:19$^{\dagger}$ & \texttt{a15d6a4c}\\
Metaworld seeding, DINO-WM & warm-start CEM on the 112 window tasks; M1, M2 (both failed) & 19:34$^{\dagger}$ & \texttt{cc1c6706}\\
Metaworld seeding, JEPA-WM & same with JEPA-WM; J1 (held), J2 (failed) & 22:44$^{\dagger}$ & \texttt{9e26d418}\\
Wall seeding & warm-start CEM on the 64 Wall tasks; S1 & 22:44$^{\dagger}$ & \texttt{df815670}\\
RoboCasa windows & 90 window tasks, DINO-WM and V-JEPA-2-AC; R1 (no direction), R2 (held), R3 (held for DINO-WM only); one launcher amendment & 00:28$^{\ddagger}$ & \texttt{99f95620}\\
\bottomrule
\end{tabular}
\caption{The frozen protocols of the reported results (continued below, where the caption explains the time marks).}
\end{table*}

\begin{table*}[tp]
\ContinuedFloat
\centering\footnotesize
\begin{tabular}{p{0.2\textwidth}p{0.47\textwidth}p{0.08\textwidth}p{0.14\textwidth}}
\toprule
Protocol & Content & Frozen & SHA-256 (prefix)\\
\midrule
RoboCasa objective & $c_\Delta$ chosen on 49 development tasks; both models on the 90 window tasks; P1 (failed), P2 (held), P3 (failed for DINO-WM) & 14:09$^{\S}$ & \texttt{a542dd7e}\\
RoboCasa adaptation, DINO-WM & predictor fine-tuned on one half of the demonstrations, tasks of the other half; A1--A3 (held) & 21:45$^{\S}$ & \texttt{17f4d3ed}\\
RoboCasa adaptation, V-JEPA-2-AC & same recipe, learning rate and early stopping fixed beforehand; V1--V3 (held) & 22:24$^{\S}$ & \texttt{4751aebb}\\
Goal information, start key & goal terms removed from the key: PushT official episodes, Metaworld and RoboCasa window tasks, both models of each; G1--G6 (7 of 7 directional tests held) & 15:15$^{\P}$ & \texttt{65609c98}\\
Goal information, endpoint CEM & CEM constrained to end the pusher at its goal position, 96 official PushT episodes, both models; E1--E3 (3 of 3 held) & 15:26$^{\P}$ & \texttt{dfcf345e}\\
Compute, PushT & predicted and real-outcome cost of the frozen $1\times$ and $4\times$ plans on the 96 official episodes, both models; C5 (held) & 00:23$^{\|}$ & \texttt{8b2cd1fd}\\
Compute, PushT addendum & rank correlation of the changes in predicted and real cost; C6 (held) & 00:27$^{\|}$ & \texttt{afc70c2b}\\
Compute, Metaworld & $1\times$ and $4\times$ CEM on the 112 window tasks, both models; C1, C2 (held), C3 (held for DINO-WM only) & 00:27$^{\|}$ & \texttt{87926f6a}\\
iCEM, PushT & $\beta$ per model chosen on the 64 development tasks; iCEM and \DPExp on iCEM on the 96 official episodes, both models; I1 (held for JEPA-WM only), I2, I3 (held) & 21:09$^{\|}$ & \texttt{783725c9}\\
Planner seeds, PushT & the main run of the 96 official episodes with two further planner seeds, both models; no predictions & 11:25$^{\ast\ast}$ & \texttt{3043ba23}\\
Image key, PushT & key built from the models' own pooled image features, $w$ chosen on the 64 development tasks; \DPExp, \DPRes and the nearest window on the 96 official episodes, both models; V1--V3 (held) & 15:36$^{\ast\ast}$ & \texttt{e21df2d0}\\
Metaworld, new tasks & \DPRes with the all-task index against warm-start CEM on 112 newly generated window tasks, both models; F1--F3 (held); written after two failed development rules (\cref{sec:supp-sprint}) & 23:51$^{\dagger\dagger}$ & \texttt{b888b87b}\\
Metaworld, 42 families & \DPRes with the all-task index against warm-start CEM on 4 new window tasks of each of the 42 task families, both models, the family as unit; G1 (failed), G2 (held for DINO-WM only), G3 (held) & 04:12$^{\ddagger\ddagger}$ & \texttt{06e115e1}\\
Metaworld, averaging & \DPAvg (mean of the 30 nearest all-task windows, chosen on the 42-family tasks) against warm-start CEM on 4 new window tasks of each of the 42 families, both models, the family as unit; H1--H3 (held) & 17:54$^{\ddagger\ddagger}$ & \texttt{fe586a2f}\\
Cross-environment & averaging the 30 nearest windows on the PushT official episodes, the Wall tasks and the RoboCasa window tasks, against the warm start in every setting; X1--X4 (held) & 11:42$^{\S\S}$ & \texttt{8526a15d}\\
Within PushT & predictor quantized to 6, 5, 4, 3 bits; rule $s<1-p_{\mathrm{avg}}$ with choices fixed on 32 development tasks per model and level; warm start and averaging on 96 new official episodes (sampler seed 2), both models; R2--R5 (held) & 19:20$^{\S\S}$ & \texttt{a382cf27}\\
BC reference & goal-conditioned MLP policy trained on all demonstration windows of PushT and Metaworld (fixed recipe, 30 epochs); 96 official PushT episodes and the 168 new Metaworld tasks; no predictions (reference line) & 16:15$^{\P\P}$ & \texttt{1514eb11}\\
Start-key averaging & \DPAvg{} with the start term of the key only (same index and $K$), its nearest window and, on Metaworld, the task's own demonstrations; the 168 new Metaworld tasks and the 90 RoboCasa window tasks, the frozen full-key average re-executed as a check; no predictions (report only) & 22:16$^{\P\P}$ & \texttt{ef52346f}\\
Matched controls & on the 168 new Metaworld tasks, both models: warm-start CEM seeded with the 30 windows that \DPAvg{} averages, and with the 100 nearest all-task windows; the model's choice among the 30 and the best of them by the cost on its real outcome; M1--M5 (held) & 00:50$^{\|\|}$ & \texttt{49b58d4b}\\
Failure-level labels & CEM with the probe on the 96 new official PushT episodes of the within-PushT test, both models at 32, 6, 5 and 4 bits; on the failures that exactly one of the warm start and \DPAvg{} solves, the label's use against labels permuted within settings; P1, P2a, P2b (held) & 14:17$^{\|\|}$ & \texttt{85b0f08c}\\
\bottomrule
\end{tabular}
\caption{The frozen protocols of the reported results, in the order they were created (the first seven on
2026-10-29, $^{\ast}$the Metaworld protocols on the next day, $^{\dagger}$on 2026-10-31, $^{\ddagger}$on 2026-11-01, $^{\S}$on 2026-11-02, $^{\P}$on 2026-11-03, $^{\|}$on 2026-11-04, $^{\ast\ast}$on 2026-11-05, $^{\dagger\dagger}$on 2026-11-07, $^{\ddagger\ddagger}$on 2026-11-08, $^{\S\S}$on 2026-11-09, $^{\P\P}$on 2026-11-10 and $^{\|\|}$on 2026-11-12; times are the server write
times of the frozen files). Each protocol was written
before any of its planning, scoring or execution, except the RoboCasa arm of the start-key
protocol, launched 34 minutes earlier, whose outcomes had not been read (\cref{sec:supp-goalinfo}). All selections were frozen from
native costs before the selected proposals were executed. \Cref{tab:supp-ledger} lists every
prediction in these protocols with its outcome. The SHA-256 prefixes refer to the original
protocol files, which we will release with our code.}
\label{tab:supp-protocols}
\end{table*}

\input{supp_predictions_table}

\section{Diagnosis details}
\label{sec:supp-diag}
\begin{table}[t]
\centering\small
\setlength{\tabcolsep}{3.6pt}
\begin{tabular}{lcccc}
\toprule
& \multicolumn{2}{c}{Standard CEM} & \multicolumn{2}{c}{Endpoint CEM}\\
\cmidrule(lr){2-3}\cmidrule(lr){4-5}
Selection rule & \W & \J & \W & \J\\
\midrule
Native cost & 53 & 44 & 56 & 56\\
Best learned reranker$^{\ast}$ & 53 & 44 & 56 & 56\\
Oracle over 900 candidates & 57 & 47 & 56 & 56\\
\midrule
Native, $+$100 expert perturb.$^{\dagger}$ & 63 & 63 & 64 & 64\\
\bottomrule
\end{tabular}
\caption{\textbf{Headroom of rescoring.} Successes on 64 development tasks (DINO-WM). The
native cost already selects nearly every success available among the recorded CEM
candidates, and rerankers trained on all 900 physical labels per task ($^{\ast}$best of
20) do not improve on it. $^{\dagger}$Diagnostic only: perturbations of the known
solution, unknown at test time, let the \emph{same} cost succeed almost everywhere.}
\label{tab:supp-support}
\end{table}

\paragraph{Candidate execution.} \Cref{tab:supp-support} gives the headroom of rescoring. For every development task and both CEM
variants we stored the populations of CEM iterations 1, 15 and 30 (300 each) and
executed all 900 sequences for the full 30 steps with the frozen replay function
(115{,}200 executions, about 8 minutes on CPU). Execution is deterministic: the
1{,}024 previously executed native top-8 candidates were reproduced exactly, and
re-executions in later figures reproduce archived block errors to the reported
precision.

\paragraph{Rerankers.} Folds are formed by source index modulo 4 (both CEM variants of
a task share a fold). Feature blocks: native cost (standardized within a task's
900 candidates); predicted visual and proprioceptive cost components; the first 16
principal components of the within-task-centered predicted patch-error map (PCA fit
on the training folds only); and action geometry relative to the start and goal
pusher position (offset of the final commanded target, path length, per-step
distances). Models: pairwise logistic ranking on (native, components),
(native, components, action geometry) and all blocks; gradient-boosted trees on
(native, components, action geometry) and all blocks; each trained for \W and for
\J, and selecting among either all 900 or the final 300 candidates
(20 variants). No variant exceeds the native choice for either CEM variant.

\begin{figure}[t]
\centering
\includegraphics[width=\linewidth]{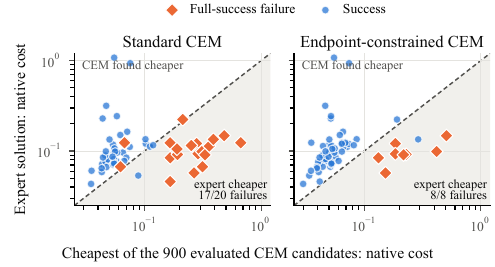}
\caption{\textbf{PushT failures are search failures} (development tasks, DINO-WM). Each
point is a task: the native cost of the cheapest of the 900 recorded CEM candidates
($x$) versus the native cost of the expert solution ($y$), for standard (left) and
endpoint-constrained CEM (right). Below the diagonal the objective prefers the expert to
every recorded CEM candidate. Nearly all full-success failures (orange diamonds) lie
below it, while most successes (blue) lie above it, where CEM found an even cheaper
action. Fig.~2 of the paper shows the same ratios next to the other settings.}
\label{fig:supp-expert}
\end{figure}

\paragraph{Expert probe.} \Cref{fig:supp-expert} shows the probe per task. The expert's normalized actions are scored with the
planner's cost function exactly as CEM scores its samples; recomputing the stored
stage-30 costs with the same call reproduces them bitwise on all 64 tasks.
Perturbations add $\sigma\cdot\epsilon$, $\epsilon\sim\mathcal N(0,I)$, to the
expert in normalized action space with a fixed seed per task, 25 each at
$\sigma\in\{0.1,0.25,0.5,1.0\}$ (100 per task); full success falls from 98\% to 4\% of
them as $\sigma$ grows. Within a task, of the pairs of perturbations that differ in full
success (156{,}176 pairs over the 64 tasks), the successful one has the lower native cost in
94.7\%. The label also survives a change of reference: replacing
the known solution by each of its executed perturbations that succeeds (34--60 per task) and
asking whether that reference is cheaper than all 900 recorded candidates, the majority of
references repeats the known solution's label on all 20 full-success failures of standard
CEM (on average 89\% of the references agree on the 17 search failures, 15\% give the search
label on the 3 scoring failures) and on 7 of the 8 failures of endpoint CEM (the eighth splits
49\%--51\%). Nor does it depend on the comparison set: on the official PushT episodes the
known solution is cheaper than every evaluated candidate, than the final population and than
CEM's executed output in exactly the same failures (29 of 34 and 23 of 28), and on the 112
Metaworld tasks (DINO-WM) it is cheaper than neither all candidates nor the final population
in any of the 67 failures (at least 1.12 times the cheapest final candidate). Metaworld's
labels under other successful references are in \cref{sec:supp-metaworld}, RoboCasa's in
\cref{sec:supp-robocasa} and those of the quantized PushT models in \cref{sec:supp-flip}.

\paragraph{Which candidates a label compares with.} Where every candidate was recorded (official PushT
episodes, quantized PushT models, Metaworld, RoboCasa), the label compares the known solution with all of
them. On the 64 PushT development tasks we recorded 900 of the 9{,}000 (iterations 1, 15 and 30), and on
Wall only the final population, so there a search label could, in principle, be a scoring label against a
cheaper candidate that CEM evaluated but did not record; we did not rerun these planners to check. On the
official episodes, comparing with all candidates, with the final population or with CEM's output gives the
same label in every failure (above), which suggests that the recorded subsets change few labels, but does
not show it for the development tasks or Wall. Nor are the labels exclusive causes: a search label does not
show that the cost ranks the other candidates correctly, and a scoring label does not show that the search
missed no other solution.

\begin{table}[t]
\centering\small
\setlength{\tabcolsep}{4pt}
\begin{tabular}{lccc}
\toprule
& \multicolumn{2}{c}{Repaired} & \\
\cmidrule(lr){2-3}
Planner & search (52) & scoring (10) & Lost\\
\midrule
\DPExp & 30 & 0 & 0\\
\DPRes & 32 & 1 & 39\\
CEM, warm start & 36 & 2 & 9\\
nearest demo. & 17 & 0 & 69\\
CEM, $4\times$ pop. & 16 & 3 & 9\\
\bottomrule
\end{tabular}
\caption{\textbf{Repairs by probe label} on the official PushT episodes (full success,
both models pooled; counted after the fact from the frozen records): failures of the
official CEM output repaired, by label, and CEM successes lost (of 130). A failure is a
search failure when the known solution costs less than every candidate CEM evaluated.}
\label{tab:supp-labelmode}
\end{table}

\paragraph{Labels and repairs within PushT.} Within one environment, the
label also tells which failures demonstrations can repair (\cref{tab:supp-labelmode}).
Every planner that uses demonstrations repairs about a third to more than two thirds of the 52 search
failures but at most 2 of the 10 scoring failures, whereas CEM with four times the
population repairs both kinds at the same rate (16 of 52 and 3 of 10). At full precision
PushT has too few scoring failures for this test; \cref{sec:supp-flip} creates them by
quantizing the predictor.

\paragraph{A farther goal (development only).} To see whether the failure type can
change within one environment, we moved the goal 60 simulator steps away on 32 development
tasks (the first development rows whose trajectories hold 60 steps; planning horizon 12,
the same CEM, cost and models, and 60-step windows retrieved with the same key). Before
running them we wrote that a protocol would be frozen only if, for both models, CEM failed
in at least 8 tasks, scoring failures made up at least half of the failures, and \DPRes won
at least as often as it lost against \DPExp. The rule was not met, so no protocol was
frozen and no official episode was run. With DINO-WM, CEM solves 7 of the 32 (full
success), and 13 of its 25 failures are scoring failures, against 3 of 20 on the 64
development tasks with 30-step goals; with JEPA-WM, CEM solves 13, and 18 of its 19 failures
remain search failures. \DPExp solves 7 and 15 ($+2/{-}2$ and $+2/{-}0$ against CEM),
\DPRes only 3 and 6, never beating \DPExp ($+0/{-}4$ and $+0/{-}9$): 60-step windows rarely
connect the start with the goal, and the nearest window solves 1 of 32 with either model.
A goal twice as far thus moves DINO-WM's failures toward scoring, as prediction errors that
grow with the horizon would, but the demonstrations do not cover such goals well enough
for Restrict to profit.
\section{Retrieval and development results}
\label{sec:supp-retrieval}
\begin{table}[t]
\centering\small
\setlength{\tabcolsep}{3pt}
\begin{tabular}{lcccc}
\toprule
& \multicolumn{2}{c}{Standard} & \multicolumn{2}{c}{Endpoint}\\
\cmidrule(lr){2-3}\cmidrule(lr){4-5}
Pool (native argmin) & \W & \J & \W & \J\\
\midrule
CEM final 300 & 53 & 44 & 56 & 56\\
+ proprio $K{=}32$ & 58 & 52 & 61 & 59\\
+ proprio $K{=}100$ (chosen) & 60 & 52 & 61 & 59\\
+ proprio $K{=}300$ & 60 & 52 & 61 & 59\\
+ random $K{=}300$ & 55 & 48 & 58 & 58\\
proprio $K{=}300$ only (no CEM) & 58 & 43 & 58 & 43\\
\midrule
+ state-keyed $K{=}32^{\dagger}$ & 59 & 49 & 59 & 56\\
\bottomrule
\end{tabular}
\caption{Development tasks (64, DINO-WM). No configuration that keeps the CEM
candidates loses a success relative to CEM; using the proposals alone, without
CEM, loses 5--16 of CEM's successes. $^{\dagger}$Privileged diagnostic that retrieves by
the simulator's start and goal \emph{states} including the block; not deployable.}
\label{tab:supp-dev}
\end{table}
The index contains every 30-step window (stride 2) of all public training trajectories,
893{,}345 windows. For development and held-out tasks, retrieval skips every variant of
every H1 and H2 source (in development folds also the held-out fold's sources); the
official episodes, whose goals come from the validation split, use the whole index. A window's actions are the stored relative pusher displacements divided by
100, the units of the environment. \Cref{tab:supp-dev} lists the development
results that led to $K{=}100$.

\section{Training-window evaluation}
\begin{table}[t]
\centering\footnotesize
\setlength{\tabcolsep}{2.4pt}
\begin{tabular}{lllcccc}
\toprule
& & & \multicolumn{2}{c}{Standard} & \multicolumn{2}{c}{Endpoint}\\
\cmidrule(lr){4-5}\cmidrule(lr){6-7}
Model & Split & Planner & \W & \J & \W & \J\\
\midrule
\multirow{6}{*}{DINO-WM} & \multirow{3}{*}{H1 (32)} & CEM & 25 & 21 & 22 & 22\\
 & & + random & 26 & 21 & 25 & 22\\
 & & \DPExp & 29 & 24 & 28 & 25\\
\cmidrule(lr){2-7}
 & \multirow{3}{*}{H2 (33)} & CEM & 30 & 26 & 27 & 27\\
 & & + random & 30 & 26 & 27 & 27\\
 & & \DPExp & 31 & 28 & 28 & 27\\
\midrule
\multirow{6}{*}{JEPA-WM} & \multirow{3}{*}{H1 (32)} & CEM & 24 & 19 & 22 & 22\\
 & & + random & 25 & 19 & 24 & 22\\
 & & \DPExp & 29 & 25 & 28 & 25\\
\cmidrule(lr){2-7}
 & \multirow{3}{*}{H2 (33)} & CEM & 29 & 26 & 30 & 30\\
 & & + random & 30 & 26 & 30 & 30\\
 & & \DPExp & 31 & 27 & 32 & 30\\
\bottomrule
\end{tabular}
\caption{Held-out results by split. No entry of \DPExp is below CEM on any task.
DINO-WM, H1 (pooled over both CEM variants): \W $+10/{-}0$ ($p{=}0.002$),
\J $+6/{-}0$ ($p{=}0.031$); H2: \W $+2/{-}0$, \J $+2/{-}0$. JEPA-WM, H1:
\W $+11/{-}0$ ($p{=}0.001$), \J $+9/{-}0$ ($p{=}0.004$); H2: \W $+4/{-}0$, \J
$+1/{-}0$.}
\label{tab:supp-splits}
\end{table}
\Cref{tab:supp-splits} gives the frozen evaluation on the 65 held-out training
windows, on which we run the closed-loop test (\cref{sec:supp-breadth}). Pooled over the two
splits, standard CEM, random windows and \DPExp reach 55/47, 56/47 and 60/52
(\W/\J) with DINO-WM and 53/45, 55/45 and 60/52 with JEPA-WM; with endpoint CEM,
49/49, 52/49, 56/52 and 52/52, 54/52, 60/55. The official output and the native
argmin reach the same outcome on every task. The proposals changed the executed action on 12 of 64 (H1) and 5
of 66 (H2) task--variant pairs with DINO-WM, and on 24 of 130 with JEPA-WM.

\section{Closed-loop MPC, search baselines and Wall}
\label{sec:supp-breadth}
\begin{table}[h]
\centering\small
\setlength{\tabcolsep}{3.5pt}
\begin{tabular}{llccc}
\toprule
Setting & Planner & Compute & \W & \J\\
\midrule
\multirow{2}{*}{\shortstack[l]{Closed-loop\\MPC (65)}}
 & MPC (official) & 1$\times$ & 52 & 50\\
 & \textbf{\DPExp} & 1.01$\times$ & \textbf{62} & \textbf{62}\\
\midrule
\multirow{4}{*}{\shortstack[l]{Wall (64)}}
 & CEM (official mean) & 1$\times$ & \multicolumn{2}{c}{39}\\
 & CEM, cheapest & 1$\times$ & \multicolumn{2}{c}{39}\\
 & + random windows & 1.01$\times$ & \multicolumn{2}{c}{42}\\
 & \DPExp & 1.01$\times$ & \multicolumn{2}{c}{47}\\
 & CEM, warm start & 1$\times$ & \multicolumn{2}{c}{\textbf{59}}\\
\bottomrule
\end{tabular}
\caption{\textbf{Closed loop and Wall} (DINO-WM). Compute is world-model evaluations
per plan (per replanning step in closed loop) relative to CEM. Wall has a single,
official success criterion.}
\label{tab:robust}
\end{table}
\paragraph{Closed-loop MPC.} We use the official receding-horizon setting with one
planning step (five simulator steps) executed per replanning step, i.e.\ six
replanning steps with remaining horizons $h=6,\dots,1$, and the same goals and
planner seeds as the open-loop runs (DINO-WM, standard CEM). At a step with $h$
planning steps left, the index holds every window of $5h$ raw actions (stride 2,
same exclusions), the query uses the current and the goal proprioception, and the
100 retrieved windows are scored with the native cost over the remaining horizon.
If the cheapest of them is cheaper than every member of the final CEM population,
its first planning step is executed, otherwise the planner's own output. The
baseline episode logs at every step whether a proposal would have won; since the
environment is deterministic and scoring consumes no planner randomness, the
method episode is run only for tasks with at least one such step, and we verify
that its executed actions equal the baseline's up to the first win.
On the 65 held-out tasks a proposal won at least one replanning step in 17
episodes (H1: 64, 70, 72, 76, 82, 83, 84, 87, 90, 91, 92; H2: 100, 106, 108,
111, 113, 122), and the method executed 27 proposal steps in total. All 17 method
episodes end in full success; the official MPC episode had failed \W in 10 and \J in
12 of them. The other 48 episodes are identical to the official MPC by construction,
and the re-run task of the smoke test reproduces its outcomes.

\paragraph{Search baselines.} Restarts rerun the official CEM with seeds
$930001+i+r\cdot10^6$ ($r=1,2,3$, task $i$) through the same loader, planner and
population capture; recomputing the stored costs of the original run ($r{=}0$)
reproduces them bitwise. Best-of-$n$ executes the native-cost argmin over the
union of the $n$ final populations ($300n$ candidates). The Gaussian control draws
100 samples around the final CEM mean with the per-dimension empirical standard
deviation of the final population (seed $20261029+i$) and executes the native
argmin over these and the 300 final candidates. A selection identical to the
original native argmin reuses its executed outcome; every other selection is
executed once with the frozen replay function.
Relative to the original native argmin, best-of-2 selected a different candidate on
30 tasks, best-of-4 on 45 and the Gaussian control on 22. Paired against CEM, best-of-2
gains one task per criterion, best-of-4 gains 4 and loses 2 (\W) and gains 4 (\J), and
the Gaussian control changes no outcome. Against \DPExp: best-of-2 $+4/{-}0$ (\W) and
$+5/{-}1$ (\J), best-of-4 $+5/{-}2$ and $+4/{-}3$, Gaussian $+5/{-}0$ and $+5/{-}0$
(\DPExp first).

\begin{figure}[t]
\centering
\includegraphics[width=\linewidth]{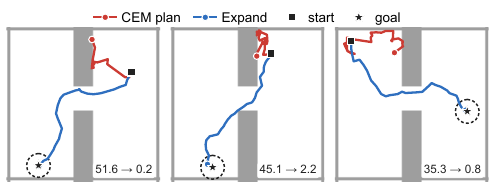}
\caption{\textbf{Wall.} Executed trajectories on three of the eight tasks in which
Expand turned a failure into a success (DINO-WM). CEM stays on the start side;
the selected demonstration window passes through the door. Dashed circle: success radius
4.5; numbers: final distance to the goal, CEM $\rightarrow$ Expand.}
\label{fig:supp-wall}
\end{figure}

\paragraph{Wall.} Typical repairs are shown in \cref{fig:supp-wall}. Of \DPExp's 8 repairs, 8 fall on search failures and 0 on scoring failures (WALL\_PROBE\_PROTOCOL\_v1); the warm start repairs 16 search and 6 scoring failures. We use the public DINO-WM Wall checkpoint (SHA-256 prefix
\texttt{ff170be5}), the official CEM configuration (horizon 6, frame skip 5,
$300\times30$ samples, 10 elites) and 64 task seeds (20261101--20261164) that
were never used before; engineering smoke tests used separate seeds.
The index contains every 30-step window (stride 2) of the 1{,}920 public Wall
training trajectories (19{,}200 windows), keyed by
$[\,0.25\,p_t,\; p_{t+30}-p_t\,]$; proposals are the windows' raw actions. The
random control draws 100 windows uniformly (seed $20261029$ plus the task seed).
Success is the official criterion, a final distance to the goal below 4.5.
CEM's output (its mean) and its cheapest candidate disagree on 4 tasks (each succeeds on two that the
other fails); all four carry the scoring label, so with the mean as the output 19 of its 25 failures are
search failures as well. Proposals were selected in 14 tasks; each ended closer to the goal
than the CEM plan (median final distance 3.9 versus 33.1), and 8 ended within the
success radius. Random windows were selected in 9 tasks and succeeded in 3. Mean
final distance: CEM mean 11.3, cheapest candidate 11.9, random windows 7.8, \DPExp 5.4.
With the closed-loop protocol's fallback to the planner output (the CEM mean) instead of the cheapest candidate, which was not pre-registered for Wall, \DPExp also reaches 47, now
$+8/{-}0$ against the CEM mean ($p{=}0.008$); random windows reach 42 ($+3/{-}0$).

\paragraph{Wall known solutions.} Wall goals are sampled by the environment, so for
the probe we script a solution: straight segments from the start to 1.5 units before
the door at its centre height, through the door to 1.5 units past it, and on to the
goal, with the 30 simulator steps allocated in proportion to segment length (the dot
moves by twice the commanded action per step). Executed in the simulator it reaches the
goal on all 64 tasks. Re-scoring the 100 retrieved windows of each task reproduces the
frozen proposal costs within $4.1{\times}10^{-5}$ (relative). The frozen run recorded
only CEM's final population, so the reference is its cheapest member. Among the 25
failures of the cheapest candidate the scripted solution is cheaper in 19 (cost ratio
0.015--0.50) and more expensive in 6 (3.1--5.3); among the 39 successes it is cheaper
in one.

\paragraph{Results by machine.} After the platform failure described in
\cref{sec:supp-log}, unfinished tasks ran on H20 instead of H200 GPUs.
\Cref{tab:supp-runtime} splits each paired comparison by machine. On the 15 Wall
tasks run on H20, proposals were selected in 4 tasks and moved the agent from 26.8,
43.4, 34.4 and 41.4 to 11.3, 5.7, 5.6 and 7.1 units from the goal, still outside the
success radius of 4.5. To rule out an effect of the machine, we re-ran on H20 the 8
H200 tasks in which a proposal turned a failure into a success: all 8 select the same
candidates, reproduce every success flag, and match the recorded final distances to
within $10^{-15}$. On the 47 search-baseline tasks run on H20, the recomputed costs of
the original CEM run differ from the stored ones by at most $3.7\times10^{-5}$
(relative) and select the same candidate on all 47.

\begin{table}[h]
\centering\small
\setlength{\tabcolsep}{3.2pt}
\begin{tabular}{llccc}
\toprule
Expand vs. & Machine & Tasks & \W & \J\\
\midrule
\multirow{2}{*}{MPC (official)} & H200 & 16 & $+1/{-}0$ & $+2/{-}0$\\
 & H20 & 49 & $+9/{-}0$ & $+10/{-}0$\\
\midrule
\multirow{2}{*}{Best-of-4 (open loop)} & H200 & 18 & $+1/{-}0$ & $+2/{-}0$\\
 & H20 & 47 & $+4/{-}2$ & $+2/{-}3$\\
\midrule
\multirow{2}{*}{CEM mean (Wall)} & H200 & 49 & \multicolumn{2}{c}{$+10/{-}1$}\\
 & H20 & 15 & \multicolumn{2}{c}{$+0/{-}1$}\\
\bottomrule
\end{tabular}
\caption{Paired gains and losses of Expand (first) split by the machine on which
each task ran. All arms of a task always ran in the same process.}
\label{tab:supp-runtime}
\end{table}

\section{Metaworld}
\label{sec:supp-metaworld}
\begin{figure*}[t]
\centering
\includegraphics[width=\linewidth]{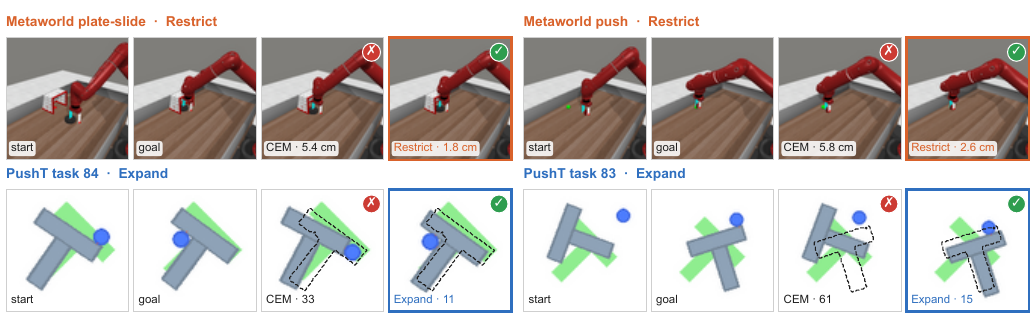}
\caption{\textbf{Repaired tasks.} Top: Metaworld window tasks with Restrict,
the first evaluation episode of plate-slide and push in which CEM fails and Restrict
succeeds. Bottom: held-out PushT training windows with Expand; dashed outline: goal pose
of the block. Numbers: final object error (cm on Metaworld, simulator units on PushT); \checkmark/$\times$: full success on Metaworld, object success on PushT.
Frames are from re-executions that reproduce the evaluation outcomes.}
\label{fig:qualitative}
\end{figure*}
\paragraph{Setup.} We use the public DINO-WM and JEPA-WM Metaworld checkpoints (SHA-256
prefixes \texttt{eab6e241} and \texttt{c3297772}; JEPA-WM with its official AdaLN
predictor, action tokens and ImageNet normalization) with the official configuration (CEM with 300 samples, 15
iterations, 10 elites, horizon 6, frame skip 5) in Metaworld 3.0.0 with MuJoCo 3.3.0,
rendered with OSMesa (the replacement machine has no EGL). The index holds every
30-step window (stride 2) of the released training demonstrations (revision
\texttt{6116f042}; 42 tasks, 300 episodes each, generated with seeds 1--3), 441{,}000
windows keyed by $[\,0.25\,p_t,\; p_{t+30}-p_t\,]$ with $p$ the hand position and
gripper opening. Evaluation seeds (2001--2007, one per task with 16 episodes each, for the
window tasks; 2101--2107 with 8 episodes each for the official protocol) never coincide with the
generation seeds; the closest training start lies 2.9\,mm from an evaluation start,
and 58 of the 112 window tasks have a training start within 1\,cm, which motivates
the exclusion control. Window-open was dropped because it is not in the training data. Two further registered controls: 100 random task windows ranked by the cost solve 66 and
58 (full success, DINO-WM and JEPA-WM), and \DPRes without training starts within 1\,cm
solves 81 and 81.

\paragraph{Window tasks.} The official evaluator warms the planning environment up
before the episode starts, while the scripted expert runs in a copy without warm-up;
we therefore replay the expert's 100 actions in the planning environment and take
start and goal from this replay, so that the expert window solves every task by
construction (112 of 112). The window $[t_0, t_0{+}30)$ is the one in which the first
object moves most, chosen from the replay alone. One CEM plan of six planning steps
is executed open loop, as on PushT. \Cref{fig:qualitative} shows two repaired tasks of each environment,
and \cref{tab:supp-mw-main} the main results.

\begin{table}[t]
\centering\small
\setlength{\tabcolsep}{3.2pt}
\begin{tabular}{lcccc}
\toprule
& \multicolumn{2}{c}{DINO-WM} & \multicolumn{2}{c}{JEPA-WM}\\
\cmidrule(lr){2-3}\cmidrule(lr){4-5}
7 families, 112 tasks & \W & \J & \W & \J\\
\midrule
CEM (official) & 80 & 45 & 73 & 27\\
\DPExp (as on PushT) & 79 & 45 & 74 & 26\\
\DPRes & 104 & 78 & 104 & \textbf{81}\\
\DPRes, all 42 tasks & 102 & 89 & 101 & 99\\
nearest window, no model & 103 & 72 & 103 & 72\\
CEM, warm start & \textbf{109} & \textbf{88} & \textbf{109} & 72\\
\midrule
known solution (reference) & 112 & 112 & 112 & 112\\
\bottomrule
\end{tabular}
\caption{\textbf{Metaworld window tasks} (evaluation seeds; the same 112 tasks for both models). The warm
start comes from later protocols; the all-task \DPRes{} row holds on new tasks of these families but not of
all 42 (\cref{sec:supp-sprint}), so bold marks the best of the other rows.}
\label{tab:supp-mw-main}
\end{table}

\paragraph{Development and protocol.} On development seeds, the official protocol
gave 12 CEM successes in 64 episodes, and in none of the 52 failures was the expert prefix
cheaper than the best CEM candidate (median cost ratio 3.4); on 28 window tasks CEM solved
13 (\J), the expert window was cheaper in none of the 15 failures, choosing among retrieved windows solved 20 (same task) and 23 (all tasks) and
the nearest window 17 and 18. Before any evaluation seed was run we froze the protocol
(\cref{tab:supp-protocols}) with five predictions for the window tasks and two for the
official protocol; \cref{tab:supp-mw-predictions} lists their outcomes and
\cref{tab:supp-mw-tasks} the window results per task.

\paragraph{Official protocol: no remedy found.} After the frozen evaluation we also
tried, on development seeds only, choosing among the retrieved demonstrations alone
at every replanning step of the official long-horizon protocol: it solved 3 (same-task
index) and 2 (all tasks) of 14 episodes against 2 for CEM, with no success on
door-open, faucet-open, plate-slide or push for any rule. With the goal 100 steps away
the objective cannot rank progress, so neither pool helps; we did not evaluate this
rule further. Subgoals inside the planning horizon, for example from retrieved
demonstrations, would be the natural next step.

\paragraph{A smaller search (development only).} To test whether the scoring failures
come from searching too much, we re-planned the 28 development window tasks with 30 and
60 instead of 300 samples per CEM iteration (15 iterations and 10 elites as before; the
tasks reproduce the frozen development records, and with DINO-WM and 300 samples so do
the costs and outcomes). A rule written before these runs would have frozen an evaluation
protocol only if search failures became the majority of CEM's failures for both models.
They did not. CEM solves fewer tasks (DINO-WM 13, 9 and 9 of 28 with 300, 60 and 30
samples; JEPA-WM 11, 8 and 8), yet none of its failures is a search failure (0 of 15, 19
and 19; 0 of 17, 20 and 20): even 450 evaluations find actions that the model prefers to
the known solution. \DPExp selects a window in at most 1 of 28 tasks, while \DPRes, which
does not depend on CEM, solves 20 and 19. On Metaworld the scoring regime does not come
from the size of the search; we froze no protocol.

\begin{table*}[tp]
\centering\small
\setlength{\tabcolsep}{3pt}
\begin{tabular}{@{}p{0.56\textwidth}p{0.19\textwidth}p{0.19\textwidth}@{}}
\toprule
Prediction (frozen) & DINO-WM & JEPA-WM\\
\midrule
A1: among CEM \J failures, the expert window is cheaper than every evaluated CEM candidate in fewer than half & holds (0 of 67) & holds (0 of 85)\\
A2: \DPExp (the PushT mode) selects a proposal in $\le$10\% of tasks and changes \W and \J by $\le$3 & holds (1 of 112; net $-1$ \W, 0 \J) & holds (0 of 112; net $+1$ \W, $-1$ \J)\\
A3: \DPRes (proposals only) beats CEM on \J (gains $>$ losses) & holds ($+43/{-}10$) & holds ($+60/{-}6$)\\
A4: world-model choice vs.\ nearest window, reported either way & $+16/{-}10$; all: $+28/{-}16$ & $+15/{-}6$; all: $+30/{-}8$\\
A5: A3 and A4 without training starts within 1\,cm & $+46/{-}10$; $+18/{-}10$ & $+60/{-}6$; $+15/{-}7$\\
\midrule
B1: official protocol, the expert prefix is cheaper than every CEM candidate in fewer than half of the failures & holds (0 of 44) & not run\\
B2: proposals win $\le$10\% of replanning steps and success changes by $\le$3 & holds (0 of 392; 12 vs.\ 12) & not run\\
\bottomrule
\end{tabular}
\caption{Metaworld predictions frozen before evaluation and their outcomes (\J unless
stated): A1--A5 on the 112 window tasks (for JEPA-WM, the same predictions in the
replication protocol), B1--B2 on the 56 official-protocol episodes (success is the
environment's flag).}
\label{tab:supp-mw-predictions}
\end{table*}

\begin{table}[t]
\centering\small
\setlength{\tabcolsep}{2.9pt}
\begin{tabular}{lcccccccc}
\toprule
& \multicolumn{4}{c}{DINO-WM} & \multicolumn{3}{c}{JEPA-WM} & \\
\cmidrule(lr){2-5}\cmidrule(lr){6-8}
Task (16 each) & CEM & +prop. & only & all & CEM & only & all & NN\\
\midrule
reach & 2 & 3 & 16 & 14 & 2 & 15 & 16 & 15\\
push & 14 & 14 & 12 & 13 & 8 & 14 & 16 & 15\\
button-press-td & 15 & 15 & 14 & 15 & 6 & 15 & 16 & 16\\
drawer-close & 1 & 2 & 1 & 12 & 2 & 0 & 16 & 0\\
door-open & 2 & 2 & 15 & 15 & 3 & 15 & 15 & 14\\
plate-slide & 2 & 2 & 12 & 7 & 0 & 10 & 5 & 9\\
faucet-open & 9 & 7 & 8 & 13 & 6 & 12 & 15 & 3\\
\midrule
total (112) & 45 & 45 & 78 & 89 & 27 & 81 & 99 & 72\\
\bottomrule
\end{tabular}
\caption{Metaworld window tasks, full successes (\J) per task. +prop.: \DPExp
(CEM population plus proposals); only: \DPRes (same-task index); all: \DPRes with the index
over all 42 tasks; NN: nearest window without any world model (identical for both
models). With the seven task families as units, exact sign-flip tests of \DPRes against CEM
give $p{=}0.25$ (DINO-WM) and $p{=}0.03$ (JEPA-WM) for the same-task index and $p{=}0.06$
and $p{=}0.016$ for the all-task index, the smallest value seven units allow; the DINO-WM
gains come from reach, door-open and plate-slide.}
\label{tab:supp-mw-tasks}
\end{table}

\begin{figure}[h]
\centering
\includegraphics[width=\linewidth]{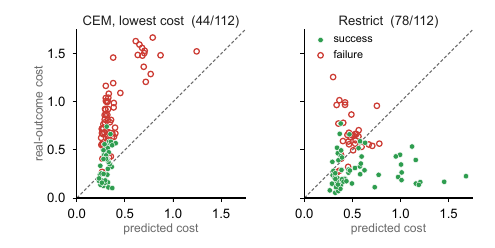}
\caption{\textbf{Metaworld: CEM's choices are overrated by the model} (DINO-WM, 112
window tasks). Predicted versus real-outcome native cost of the executed plan; above
the diagonal the real outcome is worse than predicted. Filled: full success, open:
failure. CEM's choices all have low predicted cost, which does not separate their
successes from failures; their real-outcome cost does. The windows chosen by \DPRes
are mostly predicted worse than they turn out.}
\label{fig:supp-mw-decompose}
\end{figure}

\paragraph{Predicted versus real-outcome cost.} For the 112 DINO-WM window tasks we
re-ran the frozen CEM plan (cheapest final cost, proposal costs and every outcome
reproduce the frozen record), executed five plans (CEM mean, CEM's cheapest candidate, the \DPRes window, the nearest window and the expert window), and encoded
the frame the simulator reaches with the model's own encoder. The \emph{real-outcome
cost} applies the native objective, visual and proprioceptive terms, to this encoding
and the goal encoding; the predicted cost splits the native cost into the same terms
and reproduces it for every plan. \Cref{tab:supp-decompose,fig:supp-mw-decompose}
summarize the result. The proprioceptive term contributes about $2{\times}10^{-5}$ of
the cost at $\alpha{=}0.1$, so reweighting it is not a remedy here. Per task family,
CEM's choice is optimistic in every full-success failure (reach 14, door-open 14,
drawer-close 14, plate-slide 14, faucet-open 9, push 2, button-press 1).

\begin{table}[h]
\centering\small
\setlength{\tabcolsep}{3.4pt}
\begin{tabular}{lcccc}
\toprule
Plan & \J & pred. & real & real$>$pred.\\
\midrule
CEM mean & 45 & 0.32 & 0.57 & 93\\
CEM, cheapest candidate & 44 & 0.32 & 0.63 & 94\\
\DPRes window & 78 & 0.41 & 0.30 & 38\\
nearest window & 72 & 0.63 & 0.45 & 16\\
expert window & 112 & 0.58 & 0.00 & 0\\
\bottomrule
\end{tabular}
\caption{\textbf{Metaworld: predicted vs.\ real-outcome cost} (DINO-WM, 112 window
tasks; medians of the native cost, and the number of tasks in which the real outcome
is worse than predicted). In all 68 full-success failures of CEM's cheapest
candidate the real outcome is worse than predicted (median ratio 2.3), and the expert is
always predicted worse than it is. AUROC for full success over the executed plans
without the expert: 0.54 on predicted cost, 0.94 on real-outcome cost (0.48 and 0.96
with it). On real outcomes the objective prefers the \DPRes window to CEM's choice in
58 of the 68 failures and in all 43 that \DPRes repairs. Frozen protocol
\texttt{a15d6a4c}.}
\label{tab:supp-decompose}
\end{table}

\paragraph{Within a task.} The pooled AUROC mixes tasks and plans. Reanalysing the same records
(no new runs), we compare every two executed plans of one task that differ in success and count
how often the successful one costs less (ties count half). On real outcomes the native cost
orders 97\% of the 216 pairs that differ in object success and 98\% of the 450 that differ in
full success; on predicted outcomes 14\% and 25\%, worse than chance, because CEM's choices are
the ones predicted best. Where both plans place the object and only one brings the hand within
5\,cm (243 pairs), the real-outcome cost still orders 99\% correctly and the predicted cost 34\%:
the objective sees where the hand ends, the predictions do not. Without the known solution the
shares are 96\%, 96\% and 97\% on real outcomes and 14\%, 26\% and 39\% on predictions. By
family the real-outcome cost is weakest on push (82\% of full-success pairs); the 9 of 450
pairs it orders wrongly all lie at the 5\,cm radius (e.g.\ push, seed 2002, episode 13: a
success with the object 4.7\,cm off at cost 0.64 against a failure 6.6\,cm off at 0.32). Used as
references in place of the known solution, the nearest window and \DPRes's window succeed on
40 and 43 of CEM's 68 failures; on one of the 50 failures with a successful reference that
reference is predicted cheaper than CEM's choice (by 0.3\%), so one label of 50 would change.

\begin{figure*}[t]
\centering
\includegraphics[width=\linewidth]{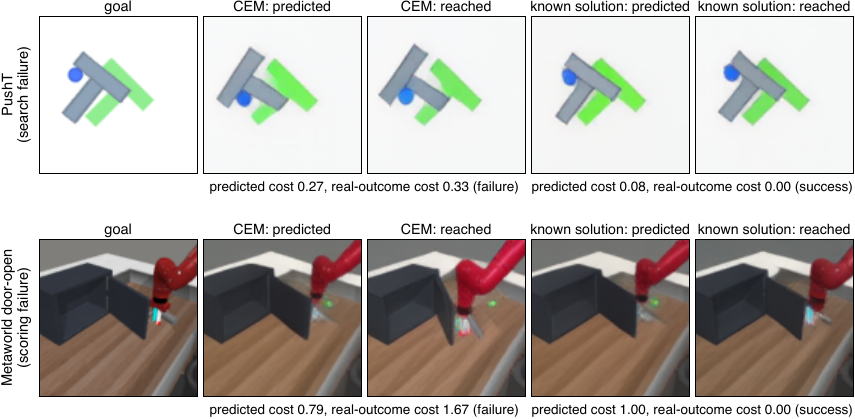}
\caption{\textbf{What the world model predicts, and what happens} (DINO-WM). Each pair
decodes, with the same released image decoder, a plan's predicted final latent
(predicted) and the encoding of the frame the simulator reaches (reached), so the two
differ only by the prediction error; crops of the $224{\times}224$ planner frames. Top,
a search failure (official PushT episode 0): the model predicts both plans correctly,
and CEM never finds the known solution. Bottom, a scoring failure (Metaworld door-open,
the task of Fig.~1 of the paper): the model predicts that CEM's choice opens the door,
which in fact opens much less, and rates the known solution worse although it reaches
the goal.}
\label{fig:supp-decode}
\end{figure*}

\paragraph{What the model predicts.} The image decoder released with JEPA-WMs maps
DINOv2 ViT-S patch features back to images. \Cref{fig:supp-decode} uses it to show a
plan's predicted final latent next to the encoding of the frame the simulator reaches;
both pass through the same decoder, so they differ only by the prediction error. On the
door-open task of Fig.~1 of the paper, the model predicts that CEM's cheapest candidate
swings the door open as in the goal, with the hand at its edge; in the reached frame the
door is open much less far and the hand is in front of the box (predicted cost 0.79,
real-outcome cost 1.67). The known solution reaches the goal yet is predicted at 1.00. The averaged plan of Fig.~1
(right) was executed on this task for the figure only, outside the averaging protocols, which
report the 168 new tasks (\cref{sec:supp-crossenv}); it ends the door 0.6\,cm from its goal.
On a PushT search failure (official episode 0), prediction and outcome agree for both
plans: the model sees that CEM's plan fails and that the known solution succeeds, but
CEM never samples anything as cheap as the latter. The windows Restrict and Expand select behave like
the known solution: on door-open, \DPRes's window is predicted at 1.01 and reaches 0.35
(success); on PushT, \DPExp's window is predicted at 0.17 and reaches 0.06 (success).
The decoder serves only this illustration; the plans were re-run from the frozen
records, whose costs and outcomes they reproduce.

\section{RoboCasa}
\label{sec:supp-robocasa}
\begin{table}[t]
\centering\small
\setlength{\tabcolsep}{4pt}
\begin{tabular}{lcc}
\toprule
Window tasks (90) & DINO-WM & V-JEPA-2-AC\\
\midrule
CEM (official) & 75 & 80\\
Expand$^{\dagger}$ & 74 & 80\\
Restrict$^{\dagger}$ & 81 & 82\\
CEM, warm start$^{\dagger}$ & 77 & 77\\
\DPAvg, no model$^{\dagger}$ & \textbf{88} & \textbf{88}\\
nearest window, no model$^{\dagger}$ & 88 & 88\\
search failures (probe) & 0 of 15 & 0 of 10\\
\midrule
standing still (reference) & 69 & 69\\
known solution (reference) & 90 & 90\\
\bottomrule
\end{tabular}
\caption{\textbf{RoboCasa} with world models trained on real-robot data; full success
(hand and object within 5\,cm). Both models fail only by scoring, and Expand selects no window, as the probe
predicts. $^{\dagger}$Retrieval uses the goal end-effector position, which the
image-only planner does not receive; \cref{sec:supp-goalinfo} removes it. Bold: best.}
\label{tab:supp-robocasa-main}
\end{table}

\paragraph{Setting.} RoboCasa (Nasiriany et al., RSS 2024) kitchen scenes with the PnPCounterTop
task, the two JEPA-WMs checkpoints trained on DROID (Khazatsky et al., RSS 2024) that we
could load: DINO-WM (\texttt{droid\_dino-wm\_noprop}, DINOv2 ViT-S/14, 224$\times$224)
and V-JEPA-2-AC (\texttt{droid\_vj2ac\_noprop} with the public V-JEPA-2 ViT-g encoder,
256$\times$256; every encoder weight matched). The third, JEPA-WM, needs DINOv3
weights whose licence we had not obtained. Each model uses its official RoboCasa
planner: CEM with 300 samples, 15 iterations, 10 elites, horizon 3,
variance scale 0.1, the official clipping of each action group (0.05 for position,
0.01 for rotation, 1 for the gripper), latent distance only ($\alpha{=}0$), images
from the left robot camera, each planner action held for five simulator steps. The
simulator is robosuite 1.5.1 with the authors' RoboCasa 0.2.0 fork; RoboCasa requires
MuJoCo 3.2.6, which we use. The released RoboCasa data contain 14 human
demonstrations of picking a bottle (283--597 steps), annotated with reach, pick and
place phases. In the official long-horizon protocol (goal at the end of a phase, 60
simulator steps of closed-loop MPC) one DINO-WM test episode failed; we did not run
that protocol further.

\paragraph{Window tasks.} All non-overlapping 15-step windows inside the reach or
pick phase that move the end effector by at least 3\,cm give 93 tasks (48 reach, 45
pick); three were used to develop the pipeline and are excluded, leaving 90. The
known solution converts the window's teleoperation actions to the planner's (DROID)
units, averages them over each block of five steps and clips them to the planner's
bounds; the goal is the frame reached by executing it from the official reset of the
demonstration state, so it solves the task by construction. Each executed plan
starts from a full official reset (scene XML, metadata and simulator state):
restoring only the simulator state moves the end of the known solution by 1.2\,cm,
whereas two full resets agree exactly. Success requires the hand within 5\,cm of its goal position, and full
success also the object. Retrieval uses all 15-step windows (stride 1) of the other
13 demonstrations, converted in the same way, with the key of the paper on the end
effector's position. Results by phase (full success, CEM / Expand / Restrict / warm start /
nearest window / known solution), DINO-WM: reach (47 tasks) 38 / 37 / 40 / 38 / 46 / 47,
pick (43 tasks) 37 / 37 / 41 / 39 / 42 / 43; V-JEPA-2-AC: reach 38 / 38 / 39 / 36 / 46 /
47, pick 42 / 42 / 43 / 41 / 42 / 43. In the failures the known solution costs 1.10--1.85
(DINO-WM) and 1.03--1.19 (V-JEPA-2-AC) times the cheapest CEM candidate. Against CEM:
\DPRes $+7/{-}1$ and $+4/{-}2$, \DPSeed $+4/{-}2$ and $+3/{-}6$, nearest window
$+13/{-}0$ and $+9/{-}1$. The two models agree on CEM's outcome in 79 of 90 tasks.
A V-JEPA-2-AC task took about 19 minutes (two CEM runs and nine full resets).
The single loss of \DPExp (DINO-WM) is a task in which it fell back to CEM's cheapest
candidate, whose hand ended 5.0\,cm from the goal, just outside the radius, while the
CEM mean ended just inside it.

\paragraph{What the window tasks test.} The windows were chosen by how far the hand moves, and in most
of them the object stays put: the null plan, which keeps the arm still and the gripper at its last
command, leaves the object within 1\,cm of its goal in 81 of the 90 tasks (median 0.0\,cm), so there full
success asks only for the hand within 5\,cm and the object left in place (the all-zero plan of
\cref{sec:supp-goalinfo} can open the gripper and drop a held object, and leaves it off in 11). For the
average, full and hand success agree on 89 of the 90 tasks, and for every arm of the frozen runs on 84
(DINO-WM) and 86 (V-JEPA-2-AC). The average and the nearest window solve 88 each (paired $+2/{-}2$; 95\%
interval of the difference $[-3,3]$ from 10{,}000 resamples of the 14 demonstrations), 21 each of the
23 tasks the all-zero plan fails ($+2/{-}2$, $[-3,3]$), and 89 and 90 with the hand alone ($+0/{-}1$);
their median hand errors are 2.0 and 1.9\,cm. RoboCasa thus replicates the scoring diagnosis with models
trained on real-robot data, and plans that do not rely on the model's choice beat it there, but these
tasks do not separate averaging from retrieving the nearest window. The labels do not depend on the
reference: on the 49 development tasks the known solution labels all 31 failures of the CEM mean as
scoring failures, and none of the 739 successful retrieved windows of these failures, each used as the
reference instead, is cheaper than CEM's cheapest candidate. (Computed after all results were known, from
the frozen records.)

\paragraph{Development tasks for the objective.} After the frozen evaluation above we
looked for a repair of RoboCasa's scoring failures on separate development tasks: all
non-overlapping 15-step windows in the place phase of the same 14 demonstrations that
move the end effector by at least 3\,cm (49 tasks; the evaluation uses only the reach
and pick phases). For each task we executed a candidate pool with DINO-WM---the known
solution, the CEM mean, the ten cheapest final CEM candidates, the 50 nearest retrieved
windows and a null plan that keeps the arm still and the gripper at its last
command---and stored the predicted and the re-encoded real final encodings (3,087
executions). Repeated resets of a task skipped the random kitchen that the official
reset builds and then discards; on development tasks this reproduces full resets
exactly (positions and rendered frames), also after a different plan was executed in
between, while skipping the recompilation of the scene as well does not.

\paragraph{What the pools show.} Ranking the 50 windows of a task, the native cost on
predicted outcomes puts a success before a failure with probability 0.55, the same cost
on real outcomes with 0.89: as on Metaworld, the scoring failures are prediction
errors, not an objective that cannot see success. A median 64\% (quartiles 58--69\%)
of a task's squared prediction error is shared by all its candidates. Smaller pools for
V-JEPA-2-AC (eight development tasks, 20 windows, the five cheapest CEM candidates,
run after the evaluation below) give 0.38 on predicted and 0.97 on real outcomes and a
shared share of 76\%. Within a pool, the prediction error grows with the distance of a
candidate's actions from the nearest demonstration window (rank correlation 0.39 for
DINO-WM, positive in 92\% of the pools; 0.56 for V-JEPA-2-AC, positive in all). For both models the predicted change from standing still is
about a third of the real change in size (0.40 and 0.31, medians). Offline on the
pools we compared the native cost, L1, per-patch and global cosine, per-channel
standardisation, weighting patches by how much the goal differs from the start or by
how well the null plan is predicted, subtracting the mean or dividing by the variance
of the model's errors on the known solutions of the other demonstrations, and costs on
predicted change. Only the change-relative cost
$c_\Delta(a)=\lVert(\hat z(a)-\hat z(a_0))-(z_g-z_0)\rVert^2$ ($a_0$ the null plan,
$z_0$ and $z_g$ the start and goal encodings) clearly improved the ranking (0.77;
0.64 for V-JEPA-2-AC).

\paragraph{Online development and choice.} With CEM optimising $c_\Delta$, against
the official CEM on the same development tasks and seeds, full success rose from 17 to
26 of 45 tasks on which both runs completed ($+13/{-}4$), and seeding CEM with the windows gave 32 ($+19/{-}4$); a
variant weighting patches by how much the goal differs from the start was worse
($+6/{-}9$ on 29 tasks). Under $c_\Delta$ the probe still finds scoring failures (18
of 19 CEM failures). We froze $c_\Delta$ without running V-JEPA-2-AC on development
tasks. On development tasks the null plan succeeds in 31 of 49, above the official CEM
(18): windows that move the hand by 3--5\,cm are solved by not moving, and the native
cost often moves the arm away; we therefore also report the null plan on the evaluation
tasks.

\paragraph{Evaluation of the frozen objective.} The protocol ran $c_\Delta$ on the 90
window tasks with both models (same goals and CEM seeds as the frozen evaluation; full
resets). Full success with DINO-WM, the native cost in parentheses: CEM mean 74 (75),
cheapest candidate 75 (74), \DPExp 75 (74), \DPRes 84 (81), \DPSeed 83 (77). With
V-JEPA-2-AC: CEM mean 80 (80), cheapest candidate 80 (80), \DPExp 79 (80), \DPRes 84 (82),
\DPSeed 77 (77). Standing still solves 69 with both models, with identical outcomes in
both runs, and the known solution 90. Against the
official CEM mean: CEM with $c_\Delta$ $+3/{-}4$ and $+5/{-}5$, \DPRes with $c_\Delta$
$+10/{-}1$ ($p{=}0.012$) and $+4/{-}0$, \DPSeed with $c_\Delta$ $+9/{-}1$ ($p{=}0.022$)
and $+2/{-}5$. Predictions: P1 (pooled, CEM with $c_\Delta$ wins more than it loses against the official CEM) failed ($+8/{-}9$); P2 (pooled, \DPSeed with $c_\Delta$) held ($+11/{-}6$, $p{=}0.33$); P3 (each model, CEM with $c_\Delta$ not worse) failed for DINO-WM ($+3/{-}4$) and held for V-JEPA-2-AC ($+5/{-}5$); P4, no direction: \DPRes with $c_\Delta$ against the official \DPRes $+3/{-}0$ and $+2/{-}0$, \DPExp $+5/{-}4$ and $+4/{-}5$. Under $c_\Delta$ the probe still finds only
scoring failures (0 of 16 and 0 of 10): CEM finds candidates that $c_\Delta$ prefers to
the known solution and that fail, so improving how windows are ranked did not improve
what CEM optimizes. Post hoc, on the 21 tasks where standing still fails (a stratum set
by the null plan alone), full successes with DINO-WM and V-JEPA-2-AC are: CEM 13 and 12
(native cost) and 13 and 13 ($c_\Delta$); with $c_\Delta$, \DPRes 15 and 15 and \DPSeed
15 and 10; the nearest window 20 and 20; the known solution 21 and 21. In the task with the largest loss
(V-JEPA-2-AC), CEM's choice under $c_\Delta$ costs 0.9\% less than the known solution
and ends 9.0\,cm from the goal; the official objective rates it worse than its own
choice (0.545 vs.\ 0.516), and the retrieved windows end 0.2\,cm away. Under the start-key averaging protocol of
\cref{sec:supp-goalinfo} (2026-11-10) the same 90 tasks were executed with an all-zero plan,
which solves 67; on the 23 tasks it fails, the full-key average solves 21, the warm start 12
with either model, the start-key average 1 and the nearest start-key window 4.

\paragraph{Adapting the model: recipe.} Everything but the frozen encoder is
fine-tuned (DINO-WM: predictor and action encoder, 19.8\,M parameters; V-JEPA-2-AC:
about 305\,M) with AdamW (weight decay $10^{-4}$, gradient clipping 1.0, bfloat16) on
clips built like planning inputs: the dataset frame of the planner's camera at step $t$
(pixel-identical to the simulator's rendering of the recorded state), the three planner
actions of steps $t$ to $t{+}15$ (block means in DROID units, clipped, as the known
solutions) and, as targets, the encodings of the frames at $t{+}5$, $t{+}10$ and
$t{+}15$; the loss is the mean squared error of the model's own three-step unroll.
Fold A holds the demonstrations with even index, fold B the odd ones; six train
(2,600 and 2,201 clips) and the last is held out, and the step with the lowest held-out
loss is kept. DINO-WM: batch 16, 3,000 steps, learning rate chosen per fold among
$3{\times}10^{-5}$ and $10^{-4}$ by the held-out loss ($3{\times}10^{-5}$ in both);
held-out loss 0.677$\to$0.434 (fold A, step 1,900) and 0.528$\to$0.333 (fold B, step
1,400). V-JEPA-2-AC: learning rate $3{\times}10^{-5}$, batch 8, stopped after 800 steps
without improvement, both fixed before its evaluation; best at step 200 in both folds
(held-out loss 0.226 and 0.198), after which it rose. Tasks of fold-B demonstrations
are planned with the fold-A weights and vice versa, so no task is planned with a model
that saw its demonstration. DINO-WM trained in minutes, V-JEPA-2-AC in about half an
hour per fold on a shared GPU.

\paragraph{Adapting the model: development.} Offline, on the development pools of the
other fold, the adapted DINO-WM lowers the prediction error (RMS 0.99$\to$0.89), lifts
the ranking of all candidates from 0.40 to 0.72 (windows only: 0.56$\to$0.70) and makes
the known solution cheaper than every zero-shot CEM candidate in 15 of 24 pools (none
zero-shot); for V-JEPA-2-AC (3 pools) the error falls from 0.76 to 0.59, the ranking of all candidates rises from 0.26 to 0.66, and the known solution becomes the cheapest in 3 of 3. Online,
each development task was planned with the other fold's weights. When the protocol was
frozen, 48 of the 49 had finished: CEM with the adapted DINO-WM solved 23 (zero-shot 18,
$+9/{-}4$), \DPRes 30, \DPSeed 26, \DPExp 26 and standing still 30, with no clear
difference from the zero-shot change-relative arms. The last task finished 30\,s after
the freeze and adds one success each to \DPRes, \DPSeed and standing still (31, 27
and 31 of 49). Adding $c_\Delta$ to the adapted model lowered its offline ranking (0.65).

\paragraph{Adapting the model: evaluation.} Frozen protocols for DINO-WM and, with the
recipe above, for V-JEPA-2-AC. Full success, DINO-WM / V-JEPA-2-AC (zero-shot in
parentheses): CEM mean 82 / 84 (75 / 80), cheapest candidate 82 / 84, \DPExp 82 / 83
(74 / 80), \DPRes 86 / 83 (81 / 82), \DPSeed 85 / 84 (77 / 77); standing still 69 / 69;
known solution 90 / 90. Against the zero-shot CEM: CEM $+10/{-}3$ and $+7/{-}3$
(pooled $+17/{-}6$, $p{=}0.035$), \DPSeed $+11/{-}1$ and $+7/{-}3$ ($+18/{-}4$,
$p{=}0.004$), \DPRes $+11/{-}0$ and $+5/{-}2$ ($+16/{-}2$, $p{=}0.001$), \DPExp
$+9/{-}2$ and $+6/{-}3$. Pooled McNemar tests treat the two models on the same task as
independent; with the 14 demonstrations as units (both models' net gains summed per
demonstration; computed after the fact), exact sign-flip tests give $p{=}0.047$ (CEM),
$0.055$ (\DPSeed) and $0.008$ (\DPRes). The predictions held for both models: CEM and \DPSeed with the
adapted model win more than they lose against the zero-shot CEM, and on the 21 tasks
where standing still fails CEM solves 15 vs.\ 13 and 16 vs.\ 12. Against the
change-relative arms: CEM $+12/{-}4$ and $+6/{-}2$, \DPSeed $+5/{-}3$ and $+10/{-}3$,
\DPRes $+4/{-}2$ and $+2/{-}3$. Under the adapted models the probe finds 1 search
failure among 8 and 0 among 6 CEM failures. Of the three tasks the adapted DINO-WM
loses against the zero-shot CEM, two end the hand 5.2 and 5.4\,cm from the goal
(zero-shot 5.0 and 4.7) and one pushes the object 10.7\,cm; \DPSeed and \DPRes with
the adapted model solve all three.

\section{Goal information in the key}
\label{sec:supp-goalinfo}
\paragraph{Protocols.} The key of Eq.~(2) contains the goal proprioception. The planner
receives it on PushT and Metaworld, where its cost weights it by $\alpha{=}0.1$ (on
Metaworld the term is about $2{\times}10^{-5}$ of the cost, \cref{sec:supp-metaworld}),
but not on RoboCasa, and full success also scores the agent's end position. Two
later protocols, written after all results above were known
(\cref{tab:supp-protocols}), test whether the gains come from this information; both
reuse the evaluation tasks, goals and CEM runs of the earlier protocols. They predate the
adoption of the warm start and of \DPAvg as uses of \DProp (the warm start had been run as a
baseline since 2026-10-31); a third protocol, written on 2026-11-10 after a review asked for it,
runs the average with the start key (end of this section). The first keeps
only the start term of the key: $[\,0.25\,p_t,\,0.5\,v_t\,]$ on PushT (full training
split), $[\,0.25\,p_t\,]$ on Metaworld (the task's demonstrations) and on RoboCasa
($p$ the end-effector position; the other 13 demonstrations), with $K{=}100$. On PushT,
start-key \DPExp pools the windows with the final CEM population saved by the frozen main
run, re-scored in the same process (largest relative difference 0.0e+00; the
recorded cheapest candidate stayed the cheapest), and falls back to that candidate as
before. On Metaworld no CEM is run: each task is rebuilt, and its start, goal and
known-solution cost must reproduce the frozen record (largest relative difference
0.0e+00). The RoboCasa arm (\DPRes with the zero-shot predictor and with the
predictor adapted on the other fold, and the nearest window) was launched at 14:41,
before the protocol was written at 15:15; 156 of its 180 tasks had finished and none of
their outcomes had been read. The second protocol gives CEM the goal pusher position:
endpoint-constrained CEM (\cref{sec:supp-endpoint}) on the 96 official episodes with the
official draws, objective, elites and budget, and \DPExp on its final population with
the frozen windows. Its smoke test on a public training window ended with the pusher at
its goal (constraint residual $2{\times}10^{-5}$); on the official episodes the largest
pusher error of its output is 0.00 units.

\begin{table*}[tp]
\centering\small
\setlength{\tabcolsep}{3pt}
\begin{tabular}{llcccc}
\toprule
& & \multicolumn{2}{c}{DINO-WM} & \multicolumn{2}{c}{JEPA-WM$^{\ast}$}\\
\cmidrule(lr){3-4}\cmidrule(lr){5-6}
& Arm & \W & \J & \W & \J\\
\midrule
\multirow{7}{*}{PushT}
 & CEM (official) & 79 & 62 & 80 & 68\\
 & \DPExp, full key & 86 & 78 & 88 & 82\\
 & \DPExp, start key & 80 & 65 & 85 & 72\\
 & \DPRes, start key & 49 & 15 & 54 & 18\\
 & nearest, start key & 29 & 2 & 29 & 2\\
 & endpoint CEM & 75 & 75 & 80 & 80\\
 & \DPExp on endpoint CEM & 81 & 80 & 85 & 84\\
\midrule
\multirow{5}{*}{Metaworld}
 & CEM (official) & 80 & 45 & 73 & 27\\
 & \DPRes, full key & 104 & 78 & 104 & 81\\
 & \DPRes, start key & 101 & 71 & 105 & 69\\
 & nearest, full key & 103 & 72 & 103 & 72\\
 & nearest, start key & 64 & 22 & 64 & 22\\
\midrule
\multirow{7}{*}{RoboCasa}
 & CEM, zero-shot & \multicolumn{2}{c}{75} & \multicolumn{2}{c}{80}\\
 & \DPRes, start key, zero-shot & \multicolumn{2}{c}{75} & \multicolumn{2}{c}{76}\\
 & CEM, adapted & \multicolumn{2}{c}{82} & \multicolumn{2}{c}{84}\\
 & \DPRes, full key, adapted & \multicolumn{2}{c}{86} & \multicolumn{2}{c}{83}\\
 & \DPRes, start key, adapted & \multicolumn{2}{c}{85} & \multicolumn{2}{c}{84}\\
 & nearest, full key & \multicolumn{2}{c}{88} & \multicolumn{2}{c}{88}\\
 & nearest, start key & \multicolumn{2}{c}{51} & \multicolumn{2}{c}{51}\\
\bottomrule
\end{tabular}
\caption{\textbf{Goal information in the key.} Successes on the 96 official PushT
episodes (DINO-WM, JEPA-WM), the 112 Metaworld window tasks (DINO-WM, JEPA-WM) and the
90 RoboCasa window tasks ($^{\ast}$V-JEPA-2-AC on RoboCasa; \J only). Full key: Eq.~(2); start key:
its start terms only. Nearest windows use no model. Endpoint CEM ends the pusher at its
goal position by construction.}
\label{tab:supp-goalinfo}
\end{table*}

\begin{table}[t]
\centering\small
\setlength{\tabcolsep}{3pt}
\begin{tabular}{lcc}
\toprule
Arm & \W{} / hand & \J\\
\midrule
\multicolumn{3}{l}{\emph{Metaworld, 168 new tasks}}\\
\DPAvg, full key, all tasks (frozen) & 138 & 127\\
\DPAvg, start key, all tasks & 42 & 13\\
nearest window, start key & 41 & 6\\
\DPAvg, start key, own demonstrations & 60 & 33\\
\DPAvg, full key, own demonstrations & 104 & 79\\
known solution & 168 & 168\\
\midrule
\multicolumn{3}{l}{\emph{RoboCasa, 90 window tasks}}\\
\DPAvg, full key (frozen) & 89 & 88\\
\DPAvg, start key & 63 & 60\\
nearest window, start key & 53 & 51\\
all-zero actions & 70 & 67\\
warm start, DINO-WM / V-JEPA-2-AC & -- & 77 / 77\\
\bottomrule
\end{tabular}
\caption{\textbf{Averaging without the goal term} (start-key averaging protocol, 2026-11-10;
no model in any plan but the warm starts). Object (RoboCasa: hand) and full success. Full key:
Eq.~(2); start key: its start term only; same index and $K$. The frozen full-key average was
re-executed and reproduces its record on every task.}
\label{tab:supp-startkey-avg}
\end{table}

\paragraph{Paired tests and predictions.} PushT, start-key \DPExp against CEM: $+3/{-}0$, $p{=}0.25$ (\J) and $+1/{-}0$, $p{=}1.00$ (\W) with DINO-WM, $+4/{-}0$, $p{=}0.12$ and $+5/{-}0$, $p{=}0.06$ with JEPA-WM; a window was selected in 12 and 10 episodes. Full-key against start-key \DPExp (\J): $+13/{-}0$, $p{=}2{\times}10^{-4}$ and $+10/{-}0$, $p{=}2{\times}10^{-3}$. Endpoint CEM against CEM: $+22/{-}9$, $p{=}0.03$ (\J), $+12/{-}16$, $p{=}0.57$ (\W) with DINO-WM, $+16/{-}4$, $p{=}0.01$ and $+10/{-}10$, $p{=}1.00$ with JEPA-WM. Full-key \DPExp against endpoint CEM: $+13/{-}10$, $p{=}0.68$ (\J), $+17/{-}6$, $p{=}0.03$ (\W) and $+8/{-}6$, $p{=}0.79$, $+12/{-}4$, $p{=}0.08$; \DPExp on endpoint CEM against endpoint CEM (\J): $+5/{-}0$, $p{=}0.06$ and $+5/{-}1$, $p{=}0.22$. Metaworld, start-key \DPRes against CEM (\J): $+41/{-}15$, $p{=}7{\times}10^{-4}$ and $+50/{-}8$, $p{=}2{\times}10^{-8}$; against the start-key nearest window: $+54/{-}5$, $p{=}2{\times}10^{-11}$ and $+50/{-}3$, $p{=}6{\times}10^{-12}$; full-key against start-key \DPRes: $+16/{-}9$, $p{=}0.23$ and $+19/{-}7$, $p{=}0.03$. RoboCasa, start-key \DPRes with the adapted against the zero-shot predictor: $+10/{-}0$, $p{=}2{\times}10^{-3}$ and $+9/{-}1$, $p{=}0.02$; adapted start-key \DPRes against zero-shot CEM: $+10/{-}0$, $p{=}2{\times}10^{-3}$ and $+7/{-}3$, $p{=}0.34$ (both models with the 14 demonstrations as units: sign-flip $p{=}0.03$); zero-shot start-key \DPRes against zero-shot CEM: $+7/{-}7$, $p{=}1.00$ and $+2/{-}6$, $p{=}0.29$. Predictions: G1 (PushT, pooled \J, start-key \DPExp beats CEM) held ($+7/{-}0$, $p{=}0.02$); G2 (same, \W) held ($+6/{-}0$, $p{=}0.03$); G3 (Metaworld, each model, start-key \DPRes beats CEM on \J) held and held; G4 (Metaworld, pooled, start-key \DPRes beats the start-key nearest window) held ($+104/{-}8$); G5 (RoboCasa, pooled, adapted beats zero-shot start-key \DPRes) held ($+19/{-}1$); G6 (RoboCasa, pooled, adapted start-key \DPRes beats zero-shot CEM) held ($+17/{-}3$); G7 had no direction. E1 (pooled \J, full-key \DPExp beats endpoint CEM) held in direction only ($+21/{-}16$, $p{=}0.51$); E2 (same, \W) held ($+29/{-}10$, $p{=}0.003$); E3 (pooled \J, \DPExp on endpoint CEM beats endpoint CEM) held ($+10/{-}1$, $p{=}0.01$); E4 had no direction.

\paragraph{The average without the goal term.} The averaging use was frozen after the two
protocols above, so a third, report-only protocol (\cref{tab:supp-protocols}; no predictions)
re-executed it with the start term of the key only, with the same index and $K{=}30$, on the
168 new Metaworld tasks of \cref{sec:supp-crossenv} and the 90 RoboCasa window tasks; the
frozen full-key average was re-executed in the same runs and reproduces its record on every
task (\cref{tab:supp-startkey-avg}). The average depends on the goal term more than any other
use: with the start key it solves 13 of the 168 Metaworld tasks instead of 127 ($+0/{-}114$
paired against the full key) and 60 of the 90 RoboCasa tasks instead of 88 ($+0/{-}28$); the
nearest start-key window solves 6 and 51. Without the goal term the 30 nearest windows come
mostly from other task families (on average 16\% from the task's own demonstrations, five
families among the 30), and their mean moves the hand to the wrong place: on a push-wall task
the full-key average ends the object 1.3\,cm from its goal and the start-key average, built
from push and soccer windows, 12.6\,cm. Restricting the index to the task's own
demonstrations recovers part of the loss with the start key (33) and, with the full key,
gives 79 rather than 127: the all-task index supplies closer windows. On RoboCasa the
start-key average ends the hand 5--7\,cm from its goal where the full-key average ends
2--3\,cm away, little better than not moving, and an all-zero plan solves 67 of the 90 tasks
(the null plan of \cref{sec:supp-robocasa}, which holds the gripper at its last command,
69). On the 23 tasks the all-zero plan fails, the full-key average solves 21, the warm start
12 with either model, the start-key average 1 and the nearest start-key window 4.

\section{iCEM}
\label{sec:supp-icem}
\paragraph{Planner.} The JEPA-WMs release has no iCEM, so we changed the official CEM
planner in the three ways of iCEM (Pinneri et al., CoRL 2020). Samples are colored noise
along the planning horizon, with power spectral density proportional to $1/f^{\beta}$ and
normalized to unit variance per step; the colorednoise package used by iCEM omits the DC
term in its normalization, which inflates the variance by about 40\% on a six-step
horizon, so with the exact normalization $\beta$ changes only the temporal correlation
(0.27, 0.45 and 0.51 between neighbouring steps for $\beta=1$, 2 and 2.5). Thirty percent
of each iteration's elites replace samples of the next population. The planner returns the
cheapest sequence evaluated in any iteration. The population does not decay, so the
budget equals CEM's ($300\times30$), and everything else (mean and standard-deviation
updates, clipping, the mean sample, the planner's generator) is the official code.

\paragraph{Development and protocol.} On the 64 development tasks, $\beta=1$, 2 and 2.5
gave 49, 49 and 49 full successes with DINO-WM and 47, 47 and 50 with JEPA-WM; the choice
rule, fixed before these runs (most full successes, then the lower median predicted cost
of the output, then the smaller $\beta$), selected $\beta=1$ and $\beta=2.5$. The protocol
(\cref{tab:supp-protocols}) then fixed three predictions for the 96 official episodes, for
each model: I1, iCEM's output has a lower predicted cost than CEM's cheapest final
candidate in more than half of the episodes; I2, \DPExp on iCEM, which executes the
cheapest retrieved window if it is cheaper than iCEM's output, wins more full successes
than it loses against iCEM; I3, it loses at most one.

\paragraph{Results.} Full success, CEM / iCEM / \DPExp on iCEM: 62 / 71 / 82 (DINO-WM)
and 68 / 78 / 87 (JEPA-WM); object success 79 / 85 / 90 and 80 / 90 / 93. iCEM against
CEM: $+13/{-}4$ ($p{=}0.049$) and $+12/{-}2$ ($p{=}0.013$). \DPExp on iCEM against iCEM:
$+11/{-}0$ ($p{=}0.001$) and $+9/{-}0$ ($p{=}0.004$), so I2 and I3 hold for both models;
I1 holds for JEPA-WM (51 of 96) but not for DINO-WM (46). \DPExp on iCEM against \DPExp on
CEM: $+7/{-}3$ and $+6/{-}1$; iCEM alone against \DPExp on CEM: $+7/{-}14$ and $+6/{-}10$.
Of iCEM's 25 and 18 full-success failures, the probe labels 19 and 14 as search failures,
and 11 of 11 and 8 of 9 repairs fall on them. In one DINO-WM episode (73) iCEM stalls: its
cheapest cost falls only from 1.10 to 1.02 over 30 iterations (CEM: from 1.13 to 0.11),
with the pusher driven off the canvas, where the cost is flat; \DPExp repairs it.

\section{More compute}
\label{sec:supp-compute}
\paragraph{Protocols.} Three protocols, frozen before any of their outputs was read
(\cref{tab:supp-protocols}), compare CEM with its official population and with four
times that population (1{,}200 samples; same seeds, iterations and elites, scored in
blocks of 300) on the 96 official PushT episodes and on the 112 Metaworld window tasks,
for both models. For CEM's cheapest final candidate and for its mean they record the
predicted cost and the real-outcome cost, i.e.\ the same objective on the encoding of the
frame the simulator reaches. On PushT the plans are those of the frozen $1\times$ and
$4\times$ runs of the official-episode protocol, whose success counts were already known;
their predicted and real-outcome costs were computed only after the protocol was frozen.
On Metaworld both populations were planned anew, and the $1\times$ runs reproduce the
frozen records. The predictions, each for both models and numbered as in the protocol
files, are: on PushT, C5, the real cost of the cheapest candidate is lower at $4\times$
in more than half of the episodes, and C6, the Spearman correlation over episodes between
its changes in predicted and in real cost exceeds 0.2; on Metaworld, C1, the predicted
cost of the cheapest candidate is lower at $4\times$ in more than half of the tasks, C2,
the same correlation stays below 0.2, and C3, full success of the CEM mean rises by at
most three tasks, net. Before the Metaworld protocol was frozen, a smoke test on one
development task showed that CEM at $4\times$ often lands in another local solution, so a
count of tasks whose real cost falls would sit near one half in either regime; the
protocol therefore tests the correlation (C2) instead.

\begin{figure}[t]
\centering
\includegraphics[width=\linewidth]{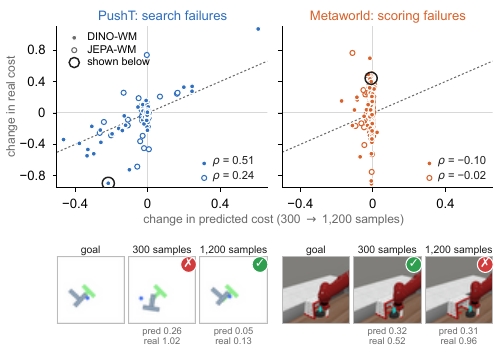}
\caption{\textbf{With more compute, the real cost follows the predicted one only in the
search regime} (reliably for DINO-WM, in direction for JEPA-WM; see the uncertainty paragraph). Top: per task, the change in predicted ($x$) and real-outcome ($y$) cost of
CEM's cheapest final candidate from 300 to 1{,}200 samples (dashed: equal change; $\rho$:
Spearman). Bottom: the circled tasks. On Metaworld the larger search lowers the predicted
cost, yet the plate stops short.}
\label{fig:supp-scale}
\end{figure}

\paragraph{Results.} PushT: the real cost of the cheapest candidate is lower at $4\times$
in 56 (DINO-WM) and 53 (JEPA-WM) of 96 episodes (C5 holds for both), and the correlations
are 0.51 and 0.24 (C6 holds for both); the median predicted cost falls from 0.067 to
0.059 and from 0.081 to 0.072, the median real cost from 0.170 to 0.126 and from 0.112 to
0.109. Metaworld: the predicted cost is lower in 108 and 108 of 112 tasks (C1 holds for
both), and the correlations are $-0.10$ and $-0.02$ (C2 holds for both); the median
predicted cost falls from 0.317 to 0.308 and from 0.345 to 0.335, while the median real
cost goes from 0.628 to 0.667 and stays at 0.664. Full success of the CEM mean goes from
45 to 43 ($+11/{-}13$; C3 holds) and from 27 to 34 ($+15/{-}8$, $p{=}0.21$; C3 fails). In
9 of JEPA-WM's 15 new successes the hand or the object ends within 0.5\,cm of the 5\,cm
threshold, and 6 of the 15 fall on faucet-open. Nine of the ten predictions hold;
\Cref{fig:supp-scale} shows the per-task changes. Only the correlations (C2, C6) separate
the regimes: C1 follows almost by construction from a larger search, the C5 counts (56
and 53 of 96) are close to one half, and C3 allows a net gain of three; the paper
therefore rests on C2 and C6, which held for both models.

\paragraph{Uncertainty (computed afterwards).} Resampling whole trajectories (PushT, 21)
and task families (Metaworld, 7), 10{,}000 times, gives 95\% intervals for the rank
correlations of $[0.35, 0.67]$ and $[-0.04, 0.50]$ on PushT and $[-0.36, 0.13]$ and
$[-0.26, 0.19]$ on Metaworld (DINO-WM, JEPA-WM). The difference between the two
environments is clear for DINO-WM ($[0.33, 0.93]$) but not for JEPA-WM ($[-0.10, 0.61]$;
it is not positive in 7.5\% of the resamples). For JEPA-WM the regimes thus differ in
direction, as predicted, but not reliably at this sample size.

\section{Retrieval by image features}
\label{sec:supp-visual}
\paragraph{Index and key.} For each PushT model, every even frame of the public training
split (1{,}173{,}620 frames: the first and last frames of the 30-step windows of the
index used on the official episodes, which spans the whole training split) is encoded with the model's own frozen encoder, exactly as the
planner encodes its observations. The $16\times16$ patch features are average-pooled to a
$4\times4$ grid (6{,}144 dimensions) and projected to 256 dimensions by a PCA fitted on
400 random training trajectories (97.6\% and 97.0\% of the variance). A window with first
frame $s$ and last frame $g$ is keyed by $[\,\lambda\,f(s),\; f(g)-f(s)\,]$ ($w$ in the protocol); the planner's start
and goal frames give the query, and the $K{=}100$ nearest windows (on development tasks excluding the task's source) replace
the proprioceptive proposals. \DPExp pools them with the frozen
final CEM population of the main run (re-scored within $3{\times}10^{-5}$ relative) and
executes the cheapest; nothing else changes.

\paragraph{Development.} On the 64 development tasks we executed all 100 windows retrieved
for $\lambda\in\{0, 0.1, 0.25, 0.5, 1\}$. With DINO-WM, 29--30 tasks had a full-success window
among them and 51--52 an object-success window; with JEPA-WM, 25--30 and 52--55. The
proprioceptive key gives 54 (full) and 60 (object). The rule fixed beforehand (most full-success tasks,
then object success, then $\lambda{=}0.5$, then the smaller $\lambda$) chose $\lambda{=}0.25$ and $\lambda{=}0$. A
linear probe locates the pusher from these features with a median error of 17.5 and 20.4
units (ridge regression, development sources held out), about the 20-unit tolerance of
full success: the image key finds windows that move the block as needed almost as often as
the proprioceptive key, but fewer that also leave the pusher at its goal.

\paragraph{A first index.} The first index fed the dataset's normalized frames to the
encoder, whose encode() normalizes again. A check built into the development runs, which
compares the features of the planner's start and goal frames with those of the same frames
in the training videos, found cosines near zero (medians 0.07 and 0.03 after projection).
The development results on that index were discarded, and the index was rebuilt from the
frames mapped back to the 0--255 range with the planner's own inverse transform. On the
rebuilt index the projected cosine has median 0.97 (DINO-WM) and 0.93 (JEPA-WM) and minimum
0.61 and 0.21. The gate set before the rebuild required a minimum above 0.5; a rule written
before the 64-row check and the development results were read accepted JEPA-WM because
every unprojected cosine exceeds 0.92 and the projected median exceeds 0.8. The lowest
frames differ only along edges (rendered versus compressed video frames). No official
episode was planned with the image key before the protocol was frozen
(\cref{tab:supp-protocols}).

\paragraph{Results.} Full success, CEM / \DPExp with the proprioceptive key / with the image
key: 62 / 78 / 72 (DINO-WM) and 68 / 82 / 78 (JEPA-WM); object success 79 / 86 / 86 and
80 / 88 / 88. Image-key \DPExp against CEM: $+11/{-}1$ ($p{=}0.006$) and $+10/{-}0$
($p{=}0.002$) in full success, $+7/{-}0$ and $+8/{-}0$ in object success (pooled
$+15/{-}0$, $p{=}6{\times}10^{-5}$); V1--V3 hold. Against the proprioceptive key:
$+2/{-}8$ and $+0/{-}4$ in full, $+3/{-}3$ and $+1/{-}1$ in object success. A window is
selected in 18 and 15 episodes, and all 11 and 10 repairs fall on search failures of the
probe. Of the 12 episodes that the proprioceptive key solves and the image key does not, 8
end with the block at its goal and a joint pusher--block error of 20.1--28.1 units (the
tolerance is 20), and in 2 no retrieved window was cheaper than CEM's candidates. The one
loss against CEM (DINO-WM, episode 75) places the block within 2 units and the joint error
at 20.1. Image-key \DPRes and the nearest window, which drop CEM's candidates, reach only
42 / 41 and 14 / 19 in full success.

\paragraph{Cost.} The world-model evaluations per plan are those of the proprioceptive key
(1.01$\times$ CEM). Building the index encodes 1{,}173{,}620 frames per model, about three
GPU-hours on shared GPUs, and stores 0.6\,GB of features per model (float16). A brute-force
top-100 query over the 893{,}345 window keys (512 dimensions, 1.8\,GB in float32) takes
about 0.64\,s on a laptop CPU (Apple M3 Pro), against about 18\,ms for the 8-dimensional
proprioceptive key with the same code; an approximate nearest-neighbour index would cut
both.
\section{Chosen mode and warm start}
\label{sec:supp-warm}
\begin{table}[t]
\centering\small
\setlength{\tabcolsep}{2.2pt}
\begin{tabular}{llcccc}
\toprule
Setting & Model & CEM & Expand & Restrict & Warm start\\
\midrule
\multirow{2}{*}{PushT} & DINO-WM & 62 & \underline{78} (0) & 62 (17) & 80 (3)\\
 & JEPA-WM & 68 & \underline{82} (0) & 62 (22) & 79 (6)\\
Wall & DINO-WM & 39 & \underline{47} (2) & -- & 59 (2)\\
\multirow{2}{*}{Metaworld} & DINO-WM & 45 & 45 (4) & \underline{78} (10) & 88 (2)\\
 & JEPA-WM & 27 & 26 (4) & \underline{81} (6) & 72 (7)\\
\multirow{2}{*}{RoboCasa} & DINO-WM & 75 & 74 (1) & \underline{81} (1) & 77 (2)\\
 & V-JEPA-2-AC & 80 & 80 (0) & \underline{82} (2) & 77 (6)\\
\midrule
\multicolumn{2}{l}{total (chosen mode)} & 396 & \multicolumn{2}{c}{529 (21)} & 532 (28)\\
\bottomrule
\end{tabular}
\caption{\textbf{Chosen mode and warm start} (main evaluations; RoboCasa zero-shot).
Full success (Wall: success) and, in parentheses, the CEM successes lost; underlined:
the mode our first protocols chose. Evaluations per plan relative to CEM: Expand 1.01 (PushT, Wall) or 1.02 (Metaworld, RoboCasa), Restrict 0.01 (PushT) or
0.02 (Metaworld, RoboCasa), warm start 1. Wall has no Restrict run.}
\label{tab:supp-warm}
\end{table}

\Cref{tab:supp-warm} lists, for every setting and model, the successes of each way of
using the windows and the CEM successes it loses. Summed over the seven, the mode our
first protocols chose solves 529 tasks and loses 21 of CEM's successes, the warm start 532 and 28.
The warm start is better on Wall (by 12), on Metaworld with DINO-WM (10) and on PushT
with DINO-WM (2), and worse on Metaworld with JEPA-WM (9), on RoboCasa (4 and 5) and on
PushT with JEPA-WM (3). Restrict, chosen in the scoring regime by those protocols, needs about 2\% of CEM's
evaluations, and Expand never loses a full success of CEM on PushT. These are the modes of our first protocols; \cref{sec:supp-crossenv} replaces Restrict by
averaging the nearest windows in the scoring regime and makes the warm start the search-regime use.

\section{Planner seeds}
\label{sec:supp-seeds}
A protocol frozen before the runs (\cref{tab:supp-protocols}; no predictions) repeats
the main run on the 96 official PushT episodes (CEM and \DPExp, with retrieval and
everything else unchanged) with two further seeds of the planner's generator. Full
success, CEM / \DPExp for seeds 0 (the frozen run), 1 and 2: DINO-WM 62 / 78, 63 / 80 and
61 / 76; JEPA-WM 68 / 82, 66 / 81 and 61 / 79. Against CEM, \DPExp gains 16, 17 and 15
and 14, 15 and 18 full successes and loses none under any seed (exact $p\le1.3{\times}10^{-4}$
each); it selects a window in 19--21 episodes. Object success, CEM / \DPExp: 79 / 86,
78 / 85 and 77 / 84; 80 / 88, 80 / 89 and 74 / 86. The spread of CEM over seeds (61--63
and 61--68 full successes) is small against the gain. The one object success that
\DPExp loses with DINO-WM is the same episode under every seed (29): CEM places the block
but leaves the pusher 155--162 units away, while the window \DPExp selects brings the
pusher within 19 units and leaves the block 36 units off; neither reaches full success.

\section{How the averaging use was found}
\label{sec:supp-sprint}
In total the warm start ties the chosen mode (\cref{sec:supp-warm}). After the main
evaluations we looked for a way of using the windows that beats it clearly. Each
attempt had a rule written before its first run, on the 28 Metaworld development tasks
per model (one development seed per task, episodes 0--3) and, for PushT, the 64
development rows.

\paragraph{Development.} (i) A small CEM around each of the ten cheapest windows, every
sample clipped to the window $\pm r$, with CEM's number of evaluations: on Metaworld it
solved 43 tasks pooled over both models against 42 for the warm start (the rule asked
for 4 more and no loss for either model; JEPA-WM 20 against 22); on PushT with DINO-WM
the warm start solved 60 of 64 and this refinement 55--59. (ii) \DPRes over the 100
windows nearest in key among all 42 tasks: 41 against 42, where the rule asked for at
least as many; a warm start with the same windows solved 37. (iii) A second opinion: the
other world model scores the warm start's plan and the window that (ii) picks, and the
cheaper one is executed: 45 against 42, where the rule asked for 4 more. All three stopped.

\paragraph{New tasks.} The development tasks and the evaluation disagreed about (ii):
paired against the warm start, it scored $+5/{-}9$ (JEPA-WM) and $+7/{-}4$ (DINO-WM) on
the 28 development tasks, but $+39/{-}12$ and $+18/{-}17$ on the 112 evaluation tasks.
Twenty-eight tasks cannot settle this, so we froze one more protocol, which records the
two failed rules and that it overrides them, and generated new window tasks for it
(seeds 3001--3007, episodes 0--15: 112 tasks, planned by both models; \cref{tab:supp-protocols}). It
predicted that \DPRes over all tasks wins more full successes than it loses against
the warm start with each model (F1), beats it pooled at one-sided $p{<}0.05$ (F2), and
does not lose against a warm start with the same windows (F3); ``clearly stronger''
required F1 and F2. All three held (\cref{tab:supp-fresh}): $+23/{-}14$ with DINO-WM,
$+31/{-}18$ with JEPA-WM, $+54/{-}32$ pooled ($p{=}0.011$), and $+66/{-}30$ against the
warm start with all-task windows. The pooled test counts each task twice, once per model;
with tasks as units (computed afterwards, summing both models' differences per task),
43 tasks favour \DPRes and 23 the warm start (sign test $p{=}0.009$), and the gain in
success rate is 9.8 points (95\% bootstrap interval over tasks 0.4--18.8).

\begin{table}[t]
\centering\small
\setlength{\tabcolsep}{3.2pt}
\begin{tabular}{lcccc}
\toprule
& \multicolumn{2}{c}{DINO-WM} & \multicolumn{2}{c}{JEPA-WM}\\
\cmidrule(lr){2-3}\cmidrule(lr){4-5}
New window tasks (112) & \W & \J & \W & \J\\
\midrule
CEM (official) & 84 & 55 & 79 & 39\\
warm start, task windows & 107 & 79 & \textbf{108} & 72\\
warm start, all-task windows & 104 & 70 & 107 & 67\\
\DPRes, task index & \textbf{109} & 84 & 103 & 69\\
\DPRes, all 42 tasks & 105 & \textbf{88} & 97 & \textbf{85}\\
\bottomrule
\end{tabular}
\caption{\textbf{New Metaworld tasks}, generated after the protocol was frozen. The warm
starts are executed as the CEM mean, as registered; their cheapest final candidates
solve 80 and 75 (task windows) and 70 and 71 (all-task windows).}
\label{tab:supp-fresh}
\end{table}

The gain is not uniform. Pooled over both models, \DPRes over all tasks wins on door
opening ($+18$ net), button pressing ($+12$) and drawer closing ($+12$), and loses on
plate sliding ($-14$) and by 1--3 on faucet, push and reach; with the seven task
families as units, the sign-flip test gives $p{=}0.27$. Door, drawer and button also
gained most on the evaluation tasks, and plate sliding also lost there. The warm start
places the object more often (215 against 202). In the 18 door tasks that \DPRes wins,
the warm start opens the door to within 3\,cm but leaves the hand 5.3--9.7\,cm from its
goal, and its cheapest candidate is always predicted cheaper than the chosen window
(DINO-WM, episode 3: 0.83 against 1.13, the window ending 1.7\,cm from the goal): the predicted cost barely sees where the hand stops, while the same
cost on the reached frames does (\cref{sec:supp-metaworld}). In the 15 plate tasks the warm start wins, the
chosen window leaves the plate 4--13\,cm off. In about half of the tasks where the two
differ, the losing plan's hand ends 5--6\,cm from its goal, just outside the radius.
\DPRes with the task index, the mode chosen in the main evaluation, solves 84 and 69
here against 78 and 81 on the evaluation tasks: differences of five to ten tasks between
arms change sign between the two task sets, and the pooled advantage of \DPRes over all
tasks is the comparison that held on both.

\paragraph{All 42 families.} Seven families are few units. We therefore froze one more
protocol (\cref{tab:supp-protocols}) on all 42 task families of the training
demonstrations, four new tasks each (seeds 3101--3142; 168 tasks, both models), with the
family as the unit of the primary test (G1: the difference in full success, summed over a
family's tasks and both models, is positive across families at exact one-sided sign-flip
$p{<}0.05$). It failed (\cref{tab:supp-family}): \DPRes over all tasks solves 96 and 87
tasks against 94 and 98 for the warm start ($+30/{-}28$ and $+25/{-}36$); 18 families
favour it and 17 the warm start, and the sign-flip test gives $p{=}0.71$ (0.78 on the 35
families not used before). G2 held for DINO-WM only and G3 held. The nearest all-task
window, which uses no model, does as well as either (96 and 96), and the model's choice
among the 100 all-task windows is no better than the nearest one ($+35/{-}35$ and
$+24/{-}33$). The probe labels every failure of CEM in the 42 families a scoring failure
(115 and 131), and Expand never selects a window, as the diagnosis predicts. The gain on
the seven families thus came from a few of them: drawer closing and door opening favour
\DPRes again ($+5$ and $+3$), while plate sliding, sliding a plate back to the side and
stick pulling favour the warm start ($-6$, $-8$ and $-6$).

\begin{table}[t]
\centering\small
\setlength{\tabcolsep}{3.2pt}
\begin{tabular}{lcccc}
\toprule
& \multicolumn{2}{c}{DINO-WM} & \multicolumn{2}{c}{JEPA-WM}\\
\cmidrule(lr){2-3}\cmidrule(lr){4-5}
All 42 families (168) & \W & \J & \W & \J\\
\midrule
CEM (official) & 86 & 53 & 82 & 37\\
\DPExp & 77 & 48 & 83 & 37\\
warm start, task windows & 124 & 94 & \textbf{131} & \textbf{98}\\
\DPRes, task index & 104 & 76 & 105 & 77\\
\DPRes, all 42 tasks & 108 & \textbf{96} & 110 & 87\\
nearest window, all tasks & \textbf{127} & \textbf{96} & 127 & 96\\
\midrule
known solution & 168 & 168 & 168 & 168\\
\bottomrule
\end{tabular}
\caption{\textbf{All 42 Metaworld task families}, four new tasks each, generated after the
protocol was frozen. \DPExp never selects a window and returns CEM's cheapest candidate.
The nearest window uses no model and is the same plan for both models.}
\label{tab:supp-family}
\end{table}

\paragraph{Averaging the nearest windows.} On the 168 tasks per model of that protocol, now
development data, we executed further uses of the windows in the simulator alone, after
replaying every stored plan to the bit. Averages of the windows nearest in key among all
tasks stood out: the mean of the 3, 10 and 30 nearest solved 109, 129 and 134 tasks against
94 and 98 for the warm start, and placed the object more often; medians, and means of the
task's own nearest windows, did worse (88--97). A rule written before the second batch of
these runs was read chose, among ten such averages, the one with most full successes over
both models, and required it to beat each warm start by at least 10: the mean of the 30
nearest windows, \DPAvg (134). Chosen among ten, this estimate is optimistic. We froze one
more protocol (\cref{tab:supp-protocols}) on new tasks of all 42 families (seeds
3201--3242, four each; 168 tasks, both models), again with the family as unit (H1: the
summed difference against each model's warm start is positive at exact one-sided
sign-flip $p{<}0.05$; H2: wins exceed losses for each model; H3: more families favour it).
All three held (\cref{tab:supp-avg}): \DPAvg solves 127 tasks against 100 and 102
($+47/{-}20$ and $+45/{-}20$); 24 families favour it and 11 the warm start ($p{=}0.014$;
0.024 on the 35 families not used before), and it places the object more often (138 against
129 and 134). Letting the model choose among the same 100 windows does worse than taking the
nearest one (80 and 83 against 93): in the scoring regime the model's choice is the
problem, not the windows. The average loses most on sliding a plate back to the side
($-8$), where it leaves the plate 22\,cm off, and on opening a faucet ($-7$), and gains most
on opening a door and on pressing a button behind a wall ($+8$ each). Its key, like those of
the other modes, uses the goal hand position (\cref{sec:supp-goalinfo}).

\begin{table}[t]
\centering\small
\setlength{\tabcolsep}{3.2pt}
\begin{tabular}{lcccc}
\toprule
& \multicolumn{2}{c}{DINO-WM} & \multicolumn{2}{c}{JEPA-WM}\\
\cmidrule(lr){2-3}\cmidrule(lr){4-5}
42 families, new (168) & \W & \J & \W & \J\\
\midrule
warm start, task windows & 129 & 100 & 134 & 102\\
warm start, cheapest cand. & 129 & 98 & 128 & 99\\
\DPRes, all 42 tasks & 96 & 80 & 106 & 83\\
nearest window, all tasks & 120 & 93 & 120 & 93\\
mean of 10 nearest & 136 & 121 & 136 & 121\\
\DPAvg (mean of 30) & \textbf{138} & \textbf{127} & \textbf{138} & \textbf{127}\\
mean of task's 10 nearest & 102 & 80 & 102 & 80\\
\midrule
known solution & 168 & 168 & 168 & 168\\
\bottomrule
\end{tabular}
\caption{\textbf{\DPAvg on new tasks of all 42 Metaworld families}, generated after the
protocol was frozen. The averages and the nearest window use no model and are the same
plans for both models; the mean of the 20 nearest solves 130.}
\label{tab:supp-avg}
\end{table}

\section{How to use demonstrations: the cross-environment test}
\label{sec:supp-crossenv}
The cross-environment protocol (\cref{tab:supp-protocols}) asks whether the use of the windows that
works follows the regime across environments. Averaging the 30 nearest windows, frozen on Metaworld
(\cref{sec:supp-sprint}), had never been run on the PushT official episodes, the Wall tasks
or the RoboCasa window tasks. Each setting keeps its frozen key and index; the averaged
plan uses no model, so one run per setting serves both models, and every episode, task and
retrieval was checked against the frozen runs (the nearest window reproduces its frozen
outcome on all 90 RoboCasa tasks). The protocol predicted that averaging solves at most
90\% of the warm start's successes on PushT (X1, each model) and Wall (X2), at least as
many on RoboCasa (X3, each model), and that following the diagnosis solves more in total
than always warm-starting and than always averaging (X4). All four held
(\cref{tab:supp-crossenv}). With goal trajectories (PushT), tasks (Wall), task families (Metaworld) and
demonstrations (RoboCasa) as units, the 95\% intervals of the average minus the warm start are $[-45,-27]$
and $[-47,-23]$ on PushT, $[-61,-50]$ on Wall, $[4,50]$ and $[2,48]$ on Metaworld and $[5,18]$ and
$[6,17]$ on RoboCasa (computed afterwards): each difference is reliable, not only its count.

\begin{table}[t]
\centering\footnotesize
\setlength{\tabcolsep}{2.5pt}
\begin{tabular}{lccccc}
\toprule
Setting (tasks), model & warm & avg.\,30 & avg.\,10 & near. & paired\\
\midrule
PushT (96), DINO & 80 & 44 & 50 & 39 & $+1/{-}37$\\
PushT (96), JEPA & 79 & 44 & 50 & 39 & $+0/{-}35$\\
Wall (64), DINO & 59 & 3 & 1 & 2 & $+0/{-}56$\\
Metaworld (168), DINO & 100 & 127 & 121 & 93 & $+47/{-}20$\\
Metaworld (168), JEPA & 102 & 127 & 121 & 93 & $+45/{-}20$\\
RoboCasa (90), DINO & 77 & 88 & 87 & 88 & $+11/{-}0$\\
RoboCasa (90), V-JEPA-2 & 77 & 88 & 87 & 88 & $+11/{-}0$\\
\bottomrule
\end{tabular}
\caption{\textbf{Cross-environment test.} Full success (Wall: success) of the warm start
and of averaging the 30 or 10 nearest windows, the nearest window, and paired wins and
losses of the 30-window average against the warm start (paired). DINO: DINO-WM, JEPA:
JEPA-WM, V-JEPA-2: V-JEPA-2-AC. Metaworld: new tasks of all 42 families (\cref{tab:supp-avg}). Following the diagnosis solves 648 tasks in total, always
warm-starting 574, always averaging 521.}
\label{tab:supp-crossenv}
\end{table}

Averaging fails where the path must be precise. On Wall it rarely reaches the door: in task
20261101, which crosses from $(14.1, 32.4)$ to $(42.5, 33.2)$, the warm start stops 0.38
from the goal and the average 17.3, because the windows pass the door at different times.
On PushT the averaged pushes blur the contact. Where the model misjudges, the average wins:
in RoboCasa task 27 the warm start leaves the hand 3.9\,cm from its goal but the object
15.5\,cm off, while the average ends with the hand at 3.4\,cm and the object at 2.4\,cm.
RoboCasa retrieval uses the goal end-effector position, which the planner does not receive;
the warm start uses the same windows.

\paragraph{Matched controls.} On Metaworld the comparison above changes two things at once:
the warm start draws its windows from the task's own demonstrations, the average from all 42
tasks. A last protocol (\cref{tab:supp-protocols}), frozen on the 168 new tasks before any
of its runs, gave every use the same 30 windows: those that \DPAvg averages. The warm start
puts them into CEM's first population (W30; the same with the 100 nearest windows, W100); the
model's choice takes the one the native cost ranks first on its prediction; and, as a
reference that a planner cannot run, every window is executed and the same cost ranks them on
the reached frame. All runs share the planner seeds of the frozen warm start, whose plans a
rerun on 42 tasks per model reproduced exactly. All five predictions held (\cref{tab:supp-matched}). With the same
windows the warm start solves fewer tasks than with the task's own (92 and 89 against 100 and
102; significant for JEPA-WM only); \DPAvg beats it by $+46/{-}11$ and $+52/{-}14$ (28 families against 5, one-sided
sign-flip $p{=}0.0002$), so its advantage does not come from the larger pool. The 30 windows
contain a full solution in 159 tasks, and the cost on real outcomes finds one in 134 and 135;
the model's choice solves 84 and 87, not more than the nearest window (93), and the
real-outcome choice beats it by $+52/{-}2$ and $+49/{-}1$. The objective recognizes the
solution once the frame is real; the predictions hide it, and a search started from the
windows does not recover it. The average, which uses no model, comes close: the
real-outcome choice beats it only by $+19/{-}12$ and $+19/{-}11$ (two-sided McNemar $p{=}0.28$ and 0.20), and it
places the object as often (138 against 137 and 138). Letting the model choose among 100 windows does not help:
of its failures, 54 and 59 pick a window beyond the 30th.

\begin{table}[t]
\centering\small
\setlength{\tabcolsep}{3.2pt}
\begin{tabular}{lcccc}
\toprule
& \multicolumn{2}{c}{DINO-WM} & \multicolumn{2}{c}{JEPA-WM}\\
\cmidrule(lr){2-3}\cmidrule(lr){4-5}
168 new tasks & \W & \J & \W & \J\\
\midrule
warm start, task windows & 129 & 100 & 134 & 102\\
warm start, the 30 windows (W30) & 127 & 92 & 122 & 89\\
warm start, 100 nearest (W100) & 122 & 91 & 126 & 89\\
model's choice among the 30 & 108 & 84 & 107 & 87\\
model's choice among 100 & 96 & 80 & 106 & 83\\
nearest window & 120 & 93 & 120 & 93\\
\DPAvg (mean of the 30) & \textbf{138} & \textbf{127} & \textbf{138} & \textbf{127}\\
\midrule
best of the 30, real outcome & 137 & 134 & 138 & 135\\
any of the 30 & 163 & 159 & 163 & 159\\
\bottomrule
\end{tabular}
\caption{\textbf{The same 30 windows, used four ways} (Metaworld, new tasks of all 42
families). The last two rows need every window executed and are references, not planners.
With its cheapest final candidate instead of its mean, W30 solves 97 and 92.}
\label{tab:supp-matched}
\end{table}

\paragraph{When the choice fails, task by task.} The executed windows show where the
choice goes wrong (an analysis made after the matched results were known). For each of the
336 task--model pairs we rank-correlate the 30 windows' predicted costs with the same cost
on their real outcomes: median Spearman 0.53, below 0.3 in 29\%. In the third of the pairs
where this correlation is lowest, the model's choice solves 42 of 112, the warm start from
the same windows 40 and the average 86 ($+46/{-}2$ against the choice); in the middle third
63, 59 and 88; in the highest third 66, 82 and 80 ($+19/{-}5$). The worse the model ranks
the windows, the more the average gains over both uses that rely on its ranking, and where
it ranks them well the warm start catches up, as \cref{prop:average} and the regime
argument expect. A large part of the prediction error is shared by all windows of a task
(median offset 0.17 against a spread of 0.19 across windows, about half of the squared
error); it shifts every cost alike and changes no choice. The part that varies across
windows is about as large as the spread of their real costs (median ratio 0.86).

\paragraph{Object or hand.} Most of the average's extra full successes over the warm start
(task windows, \cref{tab:supp-avg}) place the hand, not the object: in 26 of its 47 gains
with DINO-WM and 28 of 45 with JEPA-WM the warm start had already placed the object. On
RoboCasa, in 7 and 9 of its 11 gains the object was in place and the hand outside 5\,cm.
\Cref{tab:supp-criteria-mw} separates the criteria on the 168 new Metaworld tasks (computed after all
results were known, from the frozen records). With the object alone, the average's lead over a warm
start is within chance over families, both from the task's own windows ($+23/{-}14$ and $+19/{-}15$)
and from the same 30 windows ($+23/{-}12$ and $+26/{-}10$); with the hand alone it is large ($+49/{-}14$
and $+48/{-}16$). Against the model's choice among the same 30 windows the average wins under every
criterion, also with the object alone ($+33/{-}3$ and $+34/{-}3$). Median final errors (object, hand)
are 2.1 and 3.0\,cm for the average, 2.7 and 4.3\,cm for the warm start from the task's windows (3.0 and
4.3\,cm with JEPA-WM) and 2.7 and 3.9\,cm for the model's choice among the 30 (3.0 and 4.2\,cm). The key
carries the goal hand position (\cref{sec:supp-goalinfo}), while the model's predictions barely see
where the hand stops (\cref{sec:supp-metaworld}). Where the model misjudges, the average is the better
use than the model's choice on the object as well; what it adds over a warm start is mostly the hand.

\begin{table*}[t]
\centering\footnotesize
\setlength{\tabcolsep}{3.2pt}
\begin{tabular}{@{}llccccccccc@{}}
\toprule
& & \multicolumn{3}{c}{\J{} (object and hand)} & \multicolumn{3}{c}{\W{} (object)} & \multicolumn{3}{c}{hand}\\
\cmidrule(lr){3-5}\cmidrule(lr){6-8}\cmidrule(lr){9-11}
& the average against & solved & paired & interval & solved & paired & interval & solved & paired & interval\\
\midrule
DINO-WM & warm start, task windows & 127 / 100 & $+47/{-}20$ & $[4,50]$ & 138 / 129 & $+23/{-}14$ & $[-9,27]$ & 138 / 103 & $+49/{-}14$ & $[15,55]$\\
 & warm start, the 30 windows (W30) & 127 / 92 & $+46/{-}11$ & $[15,54]$ & 138 / 127 & $+23/{-}12$ & $[-4,26]$ & 138 / 96 & $+51/{-}9$ & $[24,61]$\\
 & model's choice among the 30 & 127 / 84 & $+48/{-}5$ & $[26,60]$ & 138 / 108 & $+33/{-}3$ & $[15,46]$ & 138 / 107 & $+42/{-}11$ & $[15,48]$\\
\midrule
JEPA-WM & warm start, task windows & 127 / 102 & $+45/{-}20$ & $[2,48]$ & 138 / 134 & $+19/{-}15$ & $[-12,20]$ & 138 / 106 & $+48/{-}16$ & $[10,53]$\\
 & warm start, the 30 windows (W30) & 127 / 89 & $+52/{-}14$ & $[18,57]$ & 138 / 122 & $+26/{-}10$ & $[0,32]$ & 138 / 98 & $+53/{-}13$ & $[22,58]$\\
 & model's choice among the 30 & 127 / 87 & $+47/{-}7$ & $[25,55]$ & 138 / 107 & $+34/{-}3$ & $[17,47]$ & 138 / 112 & $+38/{-}12$ & $[12,41]$\\
\bottomrule
\end{tabular}
\caption{\textbf{Which criterion the average wins on} (Metaworld, 168 new tasks; computed afterwards from the
frozen records). Solved: the average / the other use. Paired: wins and losses of the average. Interval: 95\%
interval of the paired difference from 10{,}000 resamples of the 42 families (seed 0). Hand: the hand within
5\,cm, whatever the object. The average uses no model, so its plans are the same for both models.}
\label{tab:supp-criteria-mw}
\end{table*}

\section{A simple error model}
\label{sec:supp-theory}
This section makes the argument of Sec.~4.3 precise in a stylized model; Propositions~1--3
of the paper are Propositions~\ref{prop:regret} and~\ref{prop:optimism}, \ref{prop:probe} and \ref{prop:average} here. A planner scores
candidates $1,\dots,N$ with real-outcome costs $r_i\ge0$ by predicted costs
$c_i=r_i+\epsilon_i$ and executes $\hat\imath=\arg\min_i c_i$; $r^\ast=\min_i r_i$ is
attained at $i^\ast$. We call $\epsilon$ \emph{$\sigma$-sub-Gaussian} if each $\epsilon_i$
has mean zero and $\mathbb{E}e^{\lambda\epsilon_i}\le e^{\lambda^2\sigma^2/2}$ for all
$\lambda$; the errors need not be independent of one another, but they are independent
of the real costs. The model is a lens, not a description of DINO-WM or JEPA-WM: their
errors are neither zero-mean nor independent of the action, $\sigma$ grows with the
distance from the training data, and CEM chooses its candidates by the predicted costs, so
the propositions hold for a candidate set fixed independently of the errors. An error shared
by all candidates shifts every cost alike and changes neither the choice nor its regret, so
$\sigma$ is to be read as the scale of the part that varies across candidates; in the
executed pools the shared part is large (about half of the squared error over Metaworld's
windows, \cref{sec:supp-crossenv}; 64\% on RoboCasa, \cref{sec:supp-robocasa}). The growth
$\sigma\sqrt{2\log N}$ of \cref{prop:optimism} needs independent Gaussian errors over
candidates equal in reality; \cref{prop:regret} needs neither.

\begin{proposition}[Regret of the cheapest candidate]\label{prop:regret}
Always $r_{\hat\imath}-r^\ast\le\epsilon_{i^\ast}-\epsilon_{\hat\imath}\le\max_i\epsilon_i-\min_i\epsilon_i$.
For $\sigma$-sub-Gaussian errors, $\mathbb{E}[r_{\hat\imath}]-r^\ast\le2\sigma\sqrt{2\log N}$.
\end{proposition}
\begin{proof}
$c_{\hat\imath}\le c_{i^\ast}$ gives $r_{\hat\imath}+\epsilon_{\hat\imath}\le r^\ast+\epsilon_{i^\ast}$.
For $N$ $\sigma$-sub-Gaussian variables $\mathbb{E}\max_i\epsilon_i\le\sigma\sqrt{2\log N}$, and the same holds for $-\min_i\epsilon_i$.
\end{proof}

\begin{proposition}[The choice looks better than it is]\label{prop:optimism}
For zero-mean errors independent of the real costs,
$\mathbb{E}[c_{\hat\imath}]\le r^\ast\le\mathbb{E}[r_{\hat\imath}]$. If all $r_i$ are equal and the
$\epsilon_i$ are i.i.d.\ $\mathcal{N}(0,\sigma^2)$, the optimism
$\mathbb{E}[r_{\hat\imath}-c_{\hat\imath}]=-\mathbb{E}\min_i\epsilon_i$ grows as
$\sigma\sqrt{2\log N}\,(1+o(1))$.
\end{proposition}
\begin{proof}
$c_{\hat\imath}\le c_{i^\ast}$ and $\mathbb{E}[c_{i^\ast}]=r^\ast$; the second inequality holds
because $r^\ast$ is the minimum. With equal $r_i$, $c_{\hat\imath}-r_{\hat\imath}=\min_i\epsilon_i$,
whose expectation for i.i.d.\ Gaussians is $-\sigma\sqrt{2\log N}(1+o(1))$.
\end{proof}

\begin{proposition}[What the probe certifies]\label{prop:probe}
Let a known solution $\bar a$ have $r(\bar a)=0$ and let every evaluated candidate fail,
$r_i\ge\Delta>0$ for $i\in\mathcal{C}$. A scoring label, $c(\bar a)\ge\min_{\mathcal{C}}c_i$,
implies $\epsilon(\bar a)-\min_{\mathcal{C}}\epsilon_i\ge\Delta$: errors as large as the gap
between the solution and the failures. For $\sigma$-sub-Gaussian errors over
$\mathcal{C}\cup\{\bar a\}$, with $\mathcal{C}$ fixed independently of the errors, the label is ``search'' with probability at least $1-\delta$
whenever $\Delta\ge\sigma\sqrt{8\log(2(|\mathcal{C}|+1)/\delta)}$.
\end{proposition}
\begin{proof}
For the minimizing $i\in\mathcal{C}$, $\epsilon(\bar a)=c(\bar a)\ge c_i=r_i+\epsilon_i\ge\Delta+\epsilon_i\ge\Delta+\min_{\mathcal{C}}\epsilon$. The event $\epsilon(\bar a)-\min\epsilon_i\ge\Delta$ requires some
$|\epsilon_j|\ge\Delta/2$; each has probability at most $2e^{-\Delta^2/(8\sigma^2)}$, and a
union bound over the $|\mathcal{C}|+1$ errors gives the claim.
\end{proof}

\begin{proposition}[More search]\label{prop:scale}
Enlarging the pool to a superset can only lower the predicted cost of the choice. As $\sigma\to0$ the
choice attains $r^\ast$, which a larger pool can only lower, so the real cost follows the
predicted one. If instead the candidates do not differ in real cost, the real cost of the
choice is unchanged while its predicted cost, the common real cost plus $\min_i\epsilon_i$,
falls as the pool grows, so the change in predicted cost says nothing about the real one.
\end{proposition}
\begin{proof}
The minimum over a superset is no larger. The limit follows from \cref{prop:regret}; the
last case is \cref{prop:optimism}.
\end{proof}
\noindent The proposition compares nested pools. The $1\times$ and $4\times$ runs of Sec.~7 are two
independent CEM runs, each adapting its samples to its own predicted costs, so the larger run's candidates
do not contain the smaller run's, and its choice need not have the lower predicted cost (it has on 108 of
the 112 Metaworld tasks with either model). The proposition motivates the compute test but does not
cover it; the test's outcome is an empirical finding.

\begin{proposition}[Choosing versus averaging windows]\label{prop:average}
Let windows $w_k=w^\ast+b+\eta_k\in\mathbb{R}^d$, $k=1,\dots,K$, scatter around a solution $w^\ast$ with a shared
offset $b$ and i.i.d.\ zero-mean deviations $\eta_k$ of covariance $\Sigma$, and let the real
cost be quadratic near the solution, $r(w)=(w-w^\ast)^\top Q(w-w^\ast)$ with $Q\succeq0$.
Then a window chosen independently of the $\eta_k$ has
$\mathbb{E}\,r=b^\top Qb+\mathrm{tr}(Q\Sigma)$, and the average $\bar w$ has
$\mathbb{E}\,r(\bar w)=b^\top Qb+\mathrm{tr}(Q\Sigma)/K$. A choice by predicted costs becomes such a choice
in the limit where errors independent of the $\eta_k$ swamp their effect on $r$; averaging
then does at least as well, and strictly better when $K>1$ and $\mathrm{tr}(Q\Sigma)>0$. With exact costs the
choice attains $\min_k r(w_k)$, which can fall below $b^\top Qb$, a level the average does
not go below in expectation.
\end{proposition}
\begin{proof}
Expand the quadratic: $\mathbb{E}\,r(w_k)=b^\top Qb+\mathbb{E}\,\eta_k^\top Q\eta_k$ and
$\mathbb{E}\,\eta_k^\top Q\eta_k=\mathrm{tr}(Q\Sigma)$; for the average the deviation is
$\frac1K\sum_k\eta_k$ with covariance $\Sigma/K$. For the last claim take $d=1$, $Q=1$:
$\min_k(b+\eta_k)^2\to0$ as $K$ grows whenever $\eta$ has mass near $-b$.
\end{proof}

\paragraph{Reading: what is proved, what is hypothesized, what is tested.} The propositions are
proved for a candidate set fixed independently of the errors, with errors independent of the real
costs. Under these conditions, \cref{prop:regret,prop:optimism} bound the regret and the optimism of
the choice; \cref{prop:probe} says what a label certifies: a scoring label always certifies errors at
least as large as the gap $\Delta$, and a search label is likely when the errors are small against it;
\cref{prop:scale} orders nested pools; and \cref{prop:average} compares a blind choice among windows
with their average: averaging cancels the windows' scatter but not their shared offset, while an exact
choice can pick a window whose deviation happens to cancel it, which needs accurate costs. None of the
conditions holds for CEM as run: CEM chooses its candidates by their predicted costs, a larger CEM run
is not a superset of a smaller one, real errors are neither zero-mean nor independent of the action,
and nothing makes the windows scatter around one solution under a quadratic cost; a scoring label does
not check these assumptions. For the planner the propositions are therefore hypotheses, which the
paper tests and does not derive: that failures are search failures where errors are small and scoring
failures where they are large; that a larger search lowers the real cost where errors are small but
mostly the predicted cost where they are large (C2, C6; \cref{sec:supp-compute}); that windows added
to the candidates are chosen where the model is accurate and rarely where it misjudges
(Secs.~6.1--6.3); and that where the errors swamp the windows' differences their average beats the
model's choice among them (M4, \cref{tab:supp-matched}), while where the model is accurate a choice
or a refinement from the windows (the warm start, which the proposition does not model) can beat the
average, which keeps the shared offset. The tests are the evidence; the propositions say why their
outcomes are expected.

\paragraph{Consistency with the measurements.} On Metaworld with DINO-WM, the median of
real minus predicted cost is $-0.58$ for the known solution and $-0.23$ for the nearest
window, neither of them chosen, $-0.14$ for the cheapest of 100 windows (Restrict) and
$+0.29$ for the cheapest of CEM's final candidates (\cref{sec:supp-metaworld}): the more
candidates the choice ranges over, and the farther from the data they lie, the more the
chosen one is overrated (\cref{prop:optimism}). In all 68 failures the known solution, whose real cost is 0, is
predicted above CEM's choice (medians 0.58 and 0.32): errors larger than the gap
(\cref{prop:probe}). A four times larger search lowers the predicted cost of the choice on
108 of 112 tasks but the real cost only on 48 and 53, while on PushT the two changes are
correlated (Spearman 0.51 and 0.24) (\cref{sec:supp-compute}; \cref{prop:scale}). On new Metaworld tasks the cheapest of
100 windows (80 and 83 of 168) does worse than the nearest window (93), which the average
of 30 beats (127): the errors swamp the windows' differences, and averaging removes scatter
(\cref{prop:average}). The direction is not universal: on the seven evaluation families the
choice among windows beat the nearest window (78 against 72 with the task's windows, 89
against 77 with all tasks; \cref{sec:supp-metaworld}), so there the errors did not swamp
the windows' differences. The proposition covers only the two limits, errors small or
large against the windows' spread. On PushT the average gains little over the nearest window (44 against
39 of 96) and falls far below the warm start (80 and 79), consistent with accurate costs and
windows that miss the block's pose by a shared offset.

\paragraph{More candidates from one source.} The compute test changes the number of candidates and where
CEM samples at once. To change only the number, we use candidates that were all executed and come from one
source, draw 400 random subsets of $n$ of them per task, let the model choose the one with the lowest
predicted cost, and record whether it succeeds, against whether any member of the subset succeeds
(coverage), the choice by the same cost on the real outcome, and a random member (Fig.~4 of the paper,
\cref{tab:supp-pool}; computed after all results were known, from the frozen records). The sources are the 900
recorded CEM candidates of the 64 PushT development tasks (DINO-WM, standard CEM; no real-outcome cost), the
30 windows of the matched protocol on the 168 new Metaworld tasks (both models), and the 50 retrieved
windows of the 49 RoboCasa development pools (DINO-WM). The curves reproduce the frozen records at their
ends: the choice among all 900 PushT candidates solves 44 tasks and 47 hold a success, as in
\cref{tab:supp-support}, and the choice among all 30 Metaworld windows solves 84 and 87, as Restrict among
them in \cref{tab:supp-matched}.

On PushT the choice keeps up with coverage at every $n$ (65\% against 67\% at $n{=}5$, 69\% against 73\% at
$n{=}900$): the cost picks nearly every success the subset holds, and failures come from what was not
sampled. On Metaworld, from one window to 30, coverage rises from 45\% to 95\% and the choice on real
outcomes to 80\%, but the model's choice peaks at 5 to 12 windows (54\% and 55\%) and falls back to 50\% and 52\%;
with the 42 families as units, its gain from $n{=}1$ to $n{=}30$ is $+8.1$ tasks $[-6.1,21.9]$ and $+11.1$
$[-0.7,22.6]$, against $+58.1$ $[47.3,68.9]$ and $+59.1$ $[48.6,69.6]$ for the choice on real outcomes. The
chosen window's predicted cost falls from 0.80 to 0.53 while its real cost falls only from 0.65 to 0.51
(DINO-WM), and its regret, the real cost above the best of the subset, grows from 0.07 at $n{=}2$ to 0.30.
On RoboCasa the model's choice falls from 57\% to 52\%, below a random window (56\%), while coverage reaches
88\% and the choice on real outcomes 80\%; the chosen windows keep the hand within 5\,cm in 95--98\% of
tasks at every $n$, so what a larger choice loses there is the object. Splitting Metaworld's tasks into
thirds by the rank correlation of the 30 windows' predicted and real costs, more windows help the choice
where the model ranks them best and hurt where it ranks them worst (\cref{tab:supp-pool}). Coverage and the
objective thus suffice in all three sources; what separates the regimes is whether the choice by predicted
cost follows them, as \cref{prop:regret} expects for errors small or large against the candidates'
differences.

\begin{table*}[t]
\centering\footnotesize
\setlength{\tabcolsep}{6pt}
\begin{tabular}{@{}llcccc@{}}
\toprule
source (tasks) & $n$ & model's choice & real outcome & coverage & random\\
\midrule
PushT, CEM (64) & 1 / 5 / 900 & 35 / 65 / 69 & -- & 35 / 67 / 73 & 35\\
Metaworld, DINO-WM (168) & 1 / 5 / 30 & 45 / 54 / 50 & 45 / 68 / 80 & 45 / 77 / 95 & 45\\
\quad worst third & 1 / 5 / 30 & 42 / 42 / 32 & 42 / 64 / 77 & 42 / 74 / 89 & 42\\
\quad best third & 1 / 5 / 30 & 47 / 61 / 63 & 47 / 68 / 77 & 47 / 75 / 95 & 47\\
Metaworld, JEPA-WM (168) & 1 / 5 / 30 & 45 / 54 / 52 & 45 / 68 / 80 & 45 / 77 / 95 & 45\\
\quad worst third & 1 / 5 / 30 & 47 / 49 / 43 & 47 / 71 / 89 & 47 / 78 / 93 & 47\\
\quad best third & 1 / 5 / 30 & 40 / 55 / 56 & 40 / 62 / 71 & 40 / 72 / 95 & 40\\
RoboCasa, DINO-WM (49) & 1 / 5 / 30 & 57 / 53 / 52 & 57 / 73 / 80 & 57 / 80 / 88 & 57 / 56 / 56\\
\bottomrule
\end{tabular}
\caption{\textbf{Choosing from more candidates of one source} (\% of tasks solved; 400 random subsets of $n$
executed candidates per task). Model's choice: lowest predicted cost; real outcome: the same cost on the
reached frame; coverage: any member succeeds; random: a random member. Thirds: Metaworld tasks split by how
well the model ranks their 30 windows (rank correlation of predicted and real cost).}
\label{tab:supp-pool}
\end{table*}

\section{Within one environment: a degraded model}
\label{sec:supp-flip}
The four environments split into two in the search regime and two in the scoring regime,
so the diagnosis might only name the environment. Here PushT, its tasks, windows,
retrieval and planner stay fixed and only the model changes.

\paragraph{Degradation.} Every linear layer of the predictor is quantized to $b$ bits,
symmetric per output row and rounded to nearest, as for low-precision deployment; the image
encoder, the goal encoding and the cost are unchanged. We use full precision and
$b=6,5,4,3$. Scoring the candidates in blocks of 100 instead of 300 reproduces every cost,
outcome and label of the full-precision development records bit for bit and was used
throughout. On the 64 development tasks, the model's rank of the known solution among the
100 windows (median) shows how the two models degrade (\cref{tab:supp-flip-dev}): DINO-WM
loses it from 6 bits on, JEPA-WM only at 3 bits.

\begin{table}[t]
\centering\footnotesize
\setlength{\tabcolsep}{2.0pt}
\begin{tabular}{lccccccccccc}
\toprule
& \multicolumn{6}{c}{DINO-WM} & \multicolumn{5}{c}{JEPA-WM}\\
\cmidrule(lr){2-7}\cmidrule(lr){8-12}
bits & rank & CEM & warm & avg. & $f$ & $s$ & rank & CEM & warm & $f$ & $s$\\
\midrule
32 & 0 & 23 & 29 & 13 & .22 & .06 & 0 & 25$^{\ast}$ & 28$^{\ast}$ & .33 & .06\\
6 & 1 & 17 & 19 & 13 & .80 & .38 & 0 & 29 & 26 & .00 & .00\\
5 & 4 & 11 & 7 & 13 & .90 & .59 & 0 & 24 & 29 & .38 & .09\\
4 & 8.5 & 0 & 1 & 13 & .91 & .91 & 0.5 & 21 & 19 & .73 & .25\\
3 & 40 & 0 & 0 & 13 & 1.0 & 1.0 & 38 & 0 & -- & 1.0 & 1.0\\
\bottomrule
\end{tabular}
\caption{\textbf{Development tasks.} Rank: median rank of the known solution among the 100
windows under the model (64 tasks; 0 = cheapest). CEM, warm start and the 30-window average
(no model, the same for both models): full success on 32 tasks ($^{\ast}$31 for JEPA-WM at
full precision, from the frozen development records). $f$: share of CEM's failures with the scoring
label (above one half, the scoring regime of Sec.~4). $s$: share of all the tasks that CEM fails
with the scoring label. Averaging solves $p_{\mathrm{avg}}{=}13/32$. The JEPA-WM warm start was not run at 3
bits, where CEM solves none.}
\label{tab:supp-flip-dev}
\end{table}

\paragraph{The rule and how it was fixed.} The rule compares how often the model's choice
fails by scoring ($s$) with how often the average of the windows fails ($1-p_{\mathrm{avg}}$, measured
directly, since averaging uses no model): it uses the warm start if $s<1-p_{\mathrm{avg}}$ and averages
otherwise. It is a heuristic, not a bound on the warm start. Seeding keeps CEM near the
data, where the model errs less, so the warm start can also avoid scoring failures: on the
Metaworld window tasks, where every CEM failure is a scoring failure, it raises full success
from 45 to 88 (DINO-WM). On PushT at full precision it repaired 16 of 17 search and 1 of 3
scoring failures of DINO-WM on the 64 development tasks of Sec.~4. The frozen protocol
justified the rule by a bound, $1-s$, on the warm start; that justification does not hold,
as Metaworld shows, so we report the rule as a heuristic. On the tasks of the main paper (development tasks for PushT and Metaworld; for Wall and
RoboCasa the evaluation tasks, read before the cross-environment protocol was frozen) it gives
the uses of the cross-environment test (\cref{tab:supp-crossenv}): $s<0.1$ on PushT and Wall, where averaging solves at most 46\%, and
$s\ge0.11$ on Metaworld and RoboCasa, where it solves 76--98\%. We wrote the rule in three
steps, all on development data and before any new episode was drawn. The first form,
written before the development runs, used the share of search labels among CEM's failures
with a threshold set on DINO-WM's development tasks. Partial development results showed
that it ignores how often CEM fails (JEPA-WM at 4 bits still solves two thirds), and the
rule above replaced it before any degraded JEPA-WM warm start was run. DINO-WM at 5 bits
then fell exactly on $s=1-p_{\mathrm{avg}}$ (19/32), and we resolved the tie toward averaging after seeing
that setting's development outcomes. The first
form, with its threshold set as planned on DINO-WM's development tasks ($0.2$), makes the
same ten choices, and so does choosing directly by the two uses' development outcomes in the
nine settings where both were run (\cref{tab:supp-flip-dev}; the JEPA-WM warm start was not
run at 3 bits). So does a simpler rule that needs neither the probe nor a run of the warm
start: use the warm start where CEM alone solves more development tasks than the average
(DINO-WM 23, 17, 11, 0 and 0, JEPA-WM 25 of 31, 29, 24, 21 and 0, against 13). It also
gives the uses of the four environments (CEM against the average: PushT 72\% and 81\%
against 41\%, Wall 61\% against 5\%, Metaworld 40\% and 24\% on the window tasks against
76\% on the new tasks, RoboCasa 83\% and 89\% against 98\%). Per setting, the probe adds
the explanation, not a better choice; what it adds per failure is below. Resampling the 32
development rows 10{,}000 times, the rule's choice is stable, the warm start being picked in
at least 95\% of resamples where it was picked and in at most 0.1\% where averaging was,
except for DINO-WM at 5 bits, the tie, where the warm start is picked in 46\%; the direct
comparison is unstable in the same places (DINO-WM at 6 bits 92\%, at 5 bits 5\%; JEPA-WM at
4 bits 94\%).

\paragraph{Regime and rule.} The rule is not the regime of Sec.~4. The regime is the majority label
among CEM's failures ($f>0.5$ in \cref{tab:supp-flip-dev}); the rule compares the share of all tasks
that CEM fails by scoring with how often the average fails. They part in two settings. DINO-WM at 6 bits
($f{=}0.80$, 12 of 15 failures) and JEPA-WM at 4 bits ($f{=}0.73$, 8 of 11) are in the scoring regime,
yet CEM fails few enough tasks by scoring ($s{=}0.38$ and $0.25$) that the warm start, which keeps most
of CEM's successes, does as well as or better than averaging on the new episodes (57 vs.\ 53, $p{=}0.62$;
67 vs.\ 53, $p{=}0.024$). The better use thus flips later than the regime (one step in point estimates;
next to the switches the differences are within chance with goal trajectories as units), and choosing by
the regime would have taken the average at JEPA-WM's 4 bits, where the warm start solves 67 against 53. This is also why the first form of the rule, a threshold on the
share of search labels among the failures, needed 0.2 rather than one half to make the same ten choices.
The regime says why the planner fails; the choice of use also weighs how often it fails and how well the
average does, which the success rates of CEM and of the average show directly.

\paragraph{Protocol.} The choices were frozen (\cref{tab:supp-protocols}) before 96 new
official episodes were drawn with the official sampler at seed 2, never run before (three of
them coincide with an episode of seed 1); goals come from the 21 validation trajectories and
retrieval uses the full training split. Per model and level the warm start was run; the
average, executed in every run, was identical in all of them. Predictions: R2, for each
model the warm start beats averaging at full precision and averaging beats it at 3 bits; R3,
in no setting does the chosen use solve significantly fewer episodes than the other (exact
McNemar, $p<0.05$), and it solves at least as many in at least 8 of the 10; R4, summed over
the settings, the rule beats always warm-starting and always averaging; R5 (direction), the
gain of averaging rises with the median rank of the known solution among the windows.

\begin{table}[t]
\centering\footnotesize
\setlength{\tabcolsep}{1.9pt}
\begin{tabular}{llccccccc}
\toprule
& bits & warm & avg. & paired & $p$ & interval & rank & search\\
\midrule
DINO & 32 & \textbf{\FlipDfull} & 53 & \FlipDfullPair & \FlipDfullP & \FlipDfullCI & 0 & \FlipDfullSearch\\
& 6 & \textbf{57} & 53 & $+16/{-}20$ & 0.62 & $[-16,8]$ & 2 & 2/39\\
& 5 & 44 & \textbf{53} & $+21/{-}12$ & 0.16 & $[-1,20]$ & 1.5 & 2/52\\
& 4 & 6 & \textbf{53} & $+49/{-}2$ & ${<}10^{-11}$ & $[34,61]$ & 7.5 & 1/90\\
& 3 & 0 & \textbf{53} & $+53/{-}0$ & ${<}10^{-15}$ & $[38,69]$ & 63 & 0/96\\
\midrule
JEPA & 32 & \textbf{\FlipJfull} & 53 & \FlipJfullPair & \FlipJfullP & \FlipJfullCI & 0 & \FlipJfullSearch\\
& 6 & \textbf{85} & 53 & $+0/{-}32$ & ${<}10^{-9}$ & $[-43,-21]$ & 0 & 4/11\\
& 5 & \textbf{86} & 53 & $+0/{-}33$ & ${<}10^{-9}$ & $[-44,-22]$ & 0 & 3/10\\
& 4 & \textbf{67} & 53 & $+10/{-}24$ & 0.024 & $[-30,2]$ & 1 & 2/29\\
& 3 & 1 & \textbf{53} & $+52/{-}0$ & ${<}10^{-15}$ & $[38,67]$ & 59 & 0/95\\
\bottomrule
\end{tabular}
\caption{\textbf{New episodes.} Full success of the warm start and the 30-window average on
96 new official episodes; bold: the rule's frozen choice. Paired: wins and losses of the
average against the warm start, exact McNemar $p$. Interval: 95\% bootstrap interval of the
average minus the warm start, resampling the 21 goal trajectories with replacement (the summed difference over a
resample, whose episode count varies, so a bound can exceed the 96 episodes). Rank: median rank of the
known solution among the windows. Search: the warm start's failures with the search label.}
\label{tab:supp-flip-new}
\end{table}

\paragraph{Results.} All four predictions held (\cref{tab:supp-flip-new}). The better use
flips between 6 and 5 bits for DINO-WM and between 4 and 3 bits for JEPA-WM, and the rule
chose it in all ten settings: \FlipRule{} episodes in total, against \FlipWarm{} for always
warm-starting and \FlipAvg{} for always averaging. The gain of averaging rises with the
measured ranking error (Spearman \FlipRho{} over the ten settings). Evidence is strong at the
ends and weak next to the switches: neither DINO-WM's 6 bits (57 vs.\ 53) nor its 5 bits (44 vs.\ 53)
differs significantly, and at JEPA-WM's 4 bits the paired test gives $p{=}0.024$ but the interval over goal
trajectories includes zero ($[-30,2]$). That the better use switches is clear; where it switches is not. Once the model degrades, the warm start's failures are nearly all
scoring failures, and it often does worse than CEM alone (DINO-WM at 5 bits on the
development tasks: 7 vs.\ 11): seeded with windows, CEM still drifts to what the model
overrates. The model's own choice among the windows degrades as well (Restrict, DINO-WM: 50,
30 and 17 at 5, 4 and 3 bits), while the average stays at 53.

Where the rule's choice lost: in episode 2 at 5 bits, DINO-WM's warm start pushed the T
block to 8.7 from its goal while the average stopped 26.3 short, the averaged pushes having
blurred the contact. On the evening of the run the host killed all our processes at once;
seven full-precision parts were rerun from the start of their sampler stream, and the 70
episodes finished before the interruption were reproduced bit for bit.

\paragraph{By criterion.} \Cref{tab:supp-flip-criteria} repeats the comparison with object success and
with the pusher alone (computed afterwards from the frozen records). With the block alone the better use
still flips: the warm start wins at full precision and at 6 and 5 bits for DINO-WM (89, 84 and 71 against
60; at 5 bits the interval reaches 0) and down to 4 bits for JEPA-WM, and averaging wins at 3 bits (60
against 11 and 27). For DINO-WM the switch moves one step, to between 5 and 4 bits, where averaging leads
by $+19/{-}10$, within chance; under this criterion the rule's choice at 5 bits would have lost (71 vs.\
60). The pusher shows why the switch comes earlier in full success: the average ends the pusher within 20
units in 88 of the 96 episodes in every setting, since the key matches the pusher's goal and the windows
end near it, while the warm start's pusher success falls with the model (DINO-WM 93, 74, 52, 13, 3). Of
the average's full-success gains over the warm start at 6, 5 and 4 bits with DINO-WM, 13 of 16, 16 of 21
and 34 of 49 come in episodes in which the warm start had already placed the block. The flip is real for
the block, and full success moves it one step earlier through the pusher.

\begin{table*}[t]
\centering\footnotesize
\setlength{\tabcolsep}{2.4pt}
\begin{tabular}{@{}llccccccccccc@{}}
\toprule
& & \multicolumn{3}{c}{\J{} (official)} & \multicolumn{3}{c}{\W{} (block)} & \multicolumn{3}{c}{pusher within 20 units} & \multicolumn{2}{c}{median error}\\
\cmidrule(lr){3-5}\cmidrule(lr){6-8}\cmidrule(lr){9-11}\cmidrule(lr){12-13}
& bits & warm/avg. & paired & interval & warm/avg. & paired & interval & warm/avg. & paired & interval & block & pusher\\
\midrule
DINO-WM & 32 & 86 / 53 & $+0/{-}33$ & $[-45,-21]$ & 89 / 60 & $+0/{-}29$ & $[-41,-18]$ & 93 / 88 & $+2/{-}7$ & $[-13,1]$ & 3.6 / 12.0 & 6.5 / 7.8\\
 & 6 & 57 / 53 & $+16/{-}20$ & $[-16,8]$ & 84 / 60 & $+3/{-}27$ & $[-35,-14]$ & 74 / 88 & $+22/{-}8$ & $[0,29]$ & 5.3 / 12.0 & 13.6 / 7.8\\
 & 5 & 44 / 53 & $+21/{-}12$ & $[-1,20]$ & 71 / 60 & $+7/{-}18$ & $[-21,0]$ & 52 / 88 & $+40/{-}4$ & $[24,48]$ & 6.8 / 12.0 & 17.7 / 7.8\\
 & 4 & 6 / 53 & $+49/{-}2$ & $[34,61]$ & 51 / 60 & $+19/{-}10$ & $[-2,20]$ & 13 / 88 & $+76/{-}1$ & $[57,95]$ & 14.4 / 12.0 & 41.2 / 7.8\\
 & 3 & 0 / 53 & $+53/{-}0$ & $[38,69]$ & 11 / 60 & $+52/{-}3$ & $[32,67]$ & 3 / 88 & $+85/{-}0$ & $[68,102]$ & 60.9 / 12.0 & 87.4 / 7.8\\
\midrule
JEPA-WM & 32 & 85 / 53 & $+1/{-}33$ & $[-45,-21]$ & 87 / 60 & $+1/{-}28$ & $[-40,-16]$ & 92 / 88 & $+3/{-}7$ & $[-11,1]$ & 2.6 / 12.0 & 4.0 / 7.8\\
 & 6 & 85 / 53 & $+0/{-}32$ & $[-43,-21]$ & 87 / 60 & $+1/{-}28$ & $[-38,-17]$ & 93 / 88 & $+2/{-}7$ & $[-13,1]$ & 2.9 / 12.0 & 4.8 / 7.8\\
 & 5 & 86 / 53 & $+0/{-}33$ & $[-44,-22]$ & 87 / 60 & $+0/{-}27$ & $[-38,-17]$ & 93 / 88 & $+3/{-}8$ & $[-13,1]$ & 4.0 / 12.0 & 6.7 / 7.8\\
 & 4 & 67 / 53 & $+10/{-}24$ & $[-30,2]$ & 81 / 60 & $+5/{-}26$ & $[-36,-8]$ & 80 / 88 & $+15/{-}7$ & $[-5,22]$ & 7.1 / 12.0 & 12.5 / 7.8\\
 & 3 & 1 / 53 & $+52/{-}0$ & $[38,67]$ & 27 / 60 & $+40/{-}7$ & $[18,48]$ & 2 / 88 & $+86/{-}0$ & $[68,105]$ & 31.2 / 12.0 & 117.7 / 7.8\\
\bottomrule
\end{tabular}
\caption{\textbf{The within-PushT test by criterion} (96 new episodes per setting; computed afterwards from the
frozen records). Successes of the warm start and of the average; paired wins and losses of the average;
95\% interval of the paired difference summed over 10{,}000 resamples of the 21 goal trajectories (seed 0); a
resample need not hold 96 episodes, so a bound can exceed 96. The \J{}
columns repeat \cref{tab:supp-flip-new}. \J, the official criterion, bounds the joint position error of
pusher and block by 20 units and the block's angle by $20^{\circ}$; \W{} bounds the block's position and
angle; the pusher criterion, defined for this analysis, bounds the pusher's position alone. Median errors
(warm / avg.) in units of the $512\times512$ canvas. The average is the same plan for both models.}
\label{tab:supp-flip-criteria}
\end{table*}

\paragraph{Failure by failure.} Per setting the label adds no choice, so we also tested
whether it tells, within a setting, which use repairs a given failure of CEM. On the
development rows of the nine settings where both uses were run, of the 67 CEM failures that exactly one of the warm
start and the average solves, the label's use (the warm start for a search label, the
average for a scoring label) solves 58: of the search-labeled ones the warm start solves 16
and the average 1, of the scoring-labeled ones the average 42 and the warm start 8 ($p{=}0.0014$
with labels permuted within settings); without 3 bits, where every label is scoring and only
the average solves anything, 45 of 54. We found this after the fact. On the 96 new episodes
CEM had not been run, and the warm start's own failures point the same way, weakly: the
average solves 1 of its 23 search-labeled failures and 201 of its 420 scoring-labeled ones
($p{=}0.077$, permuting within the eight settings that have both labels). We then froze a test (\cref{tab:supp-protocols}) and ran CEM with the probe on
the new episodes in the eight settings above 3 bits; at 3 bits CEM solves no development
task and every label is scoring, so no failure there can discriminate. The outcomes of the
warm start and the average come from the frozen runs above, fixed by a hash in the protocol;
CEM's outcomes and labels on these episodes were unknown. P1 (primary): of the failures that
exactly one use solves, the label's use solves more than labels permuted within each setting
give (one-sided $p<0.05$, 100{,}000 permutations). P2a and P2b: pooled, the warm start
solves more of the search-labeled such failures than the average, and the average more of
the scoring-labeled ones. In all 768 episodes the draw and the average's outcome matched
the frozen records, the known solution succeeded, and Restrict's outcome and cost matched.

\begin{table*}[t]
\centering\footnotesize
\setlength{\tabcolsep}{4pt}
\begin{tabular}{@{}llrrrrrrrrrrr@{}}
\toprule
& & & \multicolumn{3}{c}{search-labeled failures} & \multicolumn{3}{c}{scoring-labeled failures} & \multicolumn{2}{c}{one use solves} & \multicolumn{2}{c}{all failures}\\
\cmidrule(lr){4-6}\cmidrule(lr){7-9}\cmidrule(lr){10-11}\cmidrule(lr){12-13}
& bits & CEM & $n$ & warm & avg. & $n$ & warm & avg. & $n$ & label's use & label's use & rule's use\\
\midrule
DINO-WM & 32 & 55 & 29 & 25 & 14 & 12 & 9 & 4 & 16 & 11 (11.0) & 29 & 34\\
 & 6 & 34 & 25 & 21 & 14 & 37 & 7 & 13 & 25 & 19 (12.3) & 34 & 28\\
 & 5 & 24 & 19 & 14 & 12 & 53 & 11 & 27 & 24 & 21 (16.7) & 41 & 39\\
 & 4 & 8 & 6 & 1 & 2 & 82 & 2 & 45 & 48 & 45 (43.3) & 46 & 47\\
\midrule
JEPA-WM & 32 & 57 & 35 & 30 & 16 & 4 & 2 & 0 & 16 & 14 (14.0) & 30 & 32\\
 & 6 & 59 & 32 & 28 & 15 & 5 & 5 & 1 & 17 & 13 (13.0) & 29 & 33\\
 & 5 & 52 & 35 & 31 & 16 & 9 & 4 & 0 & 19 & 15 (15.0) & 31 & 35\\
 & 4 & 33 & 22 & 14 & 9 & 41 & 25 & 23 & 23 & 13 (10.7) & 37 & 39\\
\midrule
all & & 322 & 203 & 164 & 98 & 243 & 65 & 113 & 188 & 151 (135.9) & 277 & 287\\
\bottomrule
\end{tabular}
\caption{\textbf{CEM with the probe on the 96 new episodes.} CEM: full successes of the
official CEM. For CEM's failures by probe label: how many, and how many of them the warm start
and the average solve (frozen runs, \cref{tab:supp-flip-new}). One use solves: the failures
that exactly one of the two uses solves, and how many of them the label's use solves (the
warm start for a search label, the average for a scoring label), with its mean under labels
permuted within the setting in parentheses. All failures: successes of the label's use and
of the use the rule chose for the setting. The label needs the known solution, so the
label's use is a test of the diagnosis, not a selector one could deploy.}
\label{tab:supp-labelpred}
\end{table*}

\paragraph{Failure-level results.} All three predictions held (\cref{tab:supp-labelpred}). Of
the 188 CEM failures that exactly one use solves, the label's use solves 151, against 135.9
with labels permuted within settings (one-sided $p{<}10^{-4}$); the excess of 15.1
has a 95\% interval of $[5.9,22.6]$ when the 21 goal trajectories are resampled.
Search-labeled failures go to the warm start (71 to 5), scoring-labeled ones to the average
(80 to 32). The evidence comes from DINO-WM (96 against 83.2, $p{<}10^{-4}$); for JEPA-WM it
is not significant (55 against 52.7, $p{=}0.29$). In four settings (both models at full
precision, JEPA-WM at 6 and 5 bits) the warm start solves every such failure, so no label can
beat chance there; the excess comes from DINO-WM at 6 and 5 bits, next to its switch (19
against 12.3, 21 against 16.7), and less from both models at 4 bits (45 against 43.3, 13
against 10.7).

Two qualifications. First (post hoc), most of the average's wins over the warm start come
from the pusher alone: in 59 of the 85 failures that only the average solves, the warm start had
put the block in place (54 of the 80 scoring-labeled ones), while in 89 of the 103 that only
the warm start solves, the average left the block off. With object success instead of full
success, a search label still points to the warm start (26 to 1), but among scoring-labeled
failures the warm start places the block more often than the average (29 to 22), and the
label's use solves 48 of 78 against 43.4 ($p{=}0.048$). This fits the goal term of the key,
as on Metaworld (\cref{sec:supp-crossenv}): the key matches the pusher's goal position, and the
averaged windows end near it. Second (reported without direction, as frozen), the label's use
is not a better choice than the setting's rule: of the 446 failures it solves 277 and the
rule's use 287 (151 and 161 of the 188 that one use solves), and it wins only at DINO-WM's 6
and 5 bits (34 against 28, 41 against 39). The label needs the known solution; it informs
single failures, clearly for DINO-WM, but does not select a use.

\paragraph{By criterion and output.} Stratifying the 188 failures that exactly one use solves by whether
the two uses agree on the block, with labels permuted within setting and stratum (computed afterwards),
the label's use still solves 151 against 136.3 ($p{<}10^{-4}$). Where both uses placed the block and only
the pusher differed (73 failures), it solves 67 against 59.5 ($p{=}0.001$; DINO-WM $p{<}0.001$, JEPA-WM
$p{=}0.50$): search labels go to the warm start 13 to 5, scoring labels to the average 54 to 1. Where
exactly one use placed the block (115), it solves 84 against 76.8 ($p{=}0.003$; DINO-WM $p{=}0.010$,
JEPA-WM $p{=}0.26$): search labels go to the warm start 58 to 0, but scoring labels go to the warm start
more often than to the average (31 to 26). The search label thus points to the use that places the
block; the scoring label points to the average only through the pusher. With the pusher alone as the
criterion, the label's use solves 136 of 160 against 127.6 ($p{=}0.006$; interval of the excess
$[-1.6,18.3]$), nearly all of it scoring labels going to the average (118 to 4; search labels 18 to 20).
With CEM's cheapest final candidate as its output instead of its mean, the test is unchanged (151 of 188
against 135.9): the two differ in full success in only 3 of the 768 episodes (DINO-WM at full precision 1,
JEPA-WM at 4 bits 2), none of which exactly one use solves. Permuting the labels within setting and third of
CEM's block error, of its pusher error or of the nearest window's key distance, the label's use still solves
151 against 136.2, 138.5 and 137.8 ($p{<}10^{-4}$, $p{=}0.0002$, $p{=}0.0003$; DINO-WM $p{\le}0.0003$, JEPA-WM
$p{\approx}0.2$): the association is not the geometry of the task or of CEM's miss. Nor is it the reference:
where the nearest window succeeds (187 of the 446 failures), using it as the reference instead of the known
solution keeps the label in 170.

Where the label missed: in episode 78 at 4 bits, JEPA-WM's CEM failed with a search label
(the known solution costs 0.3825, the cheapest of its candidates 0.4162); the warm start put the
block 8.0 from its goal but left the pusher 20.7 off, while the average, with the block as
close, stopped the pusher 8.6 off and succeeded. In episode 53 at full precision, DINO-WM's
CEM failed with a scoring label (0.652 against 0.220), yet the warm start succeeded (block 3.75
off) where the average left the block 48 off, presumably because, seeded with windows, CEM
stays near the data, where the model errs less.

\section{A goal-conditioned policy as a reference}
\label{sec:supp-bc}
Every planner here is training-free; a natural question is how far they are from a policy
trained on the same demonstrations. We therefore trained, after all other results were
fixed, a goal-conditioned behaviour-cloning policy per environment and report it as a
reference line, like the known solution, with no directional prediction
(\cref{tab:supp-protocols}). The policy is an MLP ($8\to512\to512\to512\to$ action chunk,
GELU) that maps the agent's proprioception now and at the goal, the same information as the
retrieval key, to the 30-step action chunk. It is trained on every window of the retrieval
index (PushT: the 893{,}345 windows of the training split, targets the relative actions;
Metaworld: the 441{,}000 windows of the 42 tasks' demonstrations), with inputs and targets
standardized, Adam at $10^{-3}$ with cosine decay, batches of 4{,}096, a fixed 30 epochs and
seed 0; 2\% of the trajectories are held out for a validation loss that is reported but
never used for selection (standardized MSE 0.207 on PushT, 0.703 on Metaworld; the Metaworld
chunks vary far more between demonstrations of the same start and goal). The policy is
executed open loop on the 96 official PushT episodes and on the 168 new Metaworld tasks of
the averaging protocol, where the stored averaged plan is re-executed and must reproduce its
frozen outcome. We did not train a policy on RoboCasa's 14 demonstrations. A smoke run of
one PushT rank and one Metaworld task preceded the freeze; their outcomes carry no
prediction.

\paragraph{Results.} On PushT the policy solves \BCptFull{} of 96 episodes in full and
\BCptObj{} in object success, against 80 and 79 for the warm starts (paired $+0/{-}39$ and
$+0/{-}38$; cluster intervals $[-53,-26]$ and $[-55,-23]$ over the 21 goal trajectories) and
44 for averaging ($+9/{-}12$, $[-15,8]$). On Metaworld it solves \BCmwFull{} of 168
(\BCmwObj{} object), against 100 and 102 for the warm starts (\BCmwPairWarm{} and $+35/{-}9$;
intervals $[9,49]$ and $[10,44]$ over the 42 families), 127 for averaging (\BCmwPairAvg,
$[-16,17]$) and 80 for the model's choice among the same windows ($+57/{-}9$, $[29,68]$). The
reference thus falls on the side the diagnosis predicts: in the search regime a search started
from the demonstrations beats a policy trained on them, in the scoring regime the two
model-free plans tie and both beat the model's choice.

\section{Engineering log}
\label{sec:supp-log}
All runs used shared GPUs. The following engineering events occurred; none changed a
frozen method or protocol choice, and all but the interruption of the within-PushT test happened before any outcome of the
affected run had been read.
\begin{itemize}
\item The audited DINO-WM loader refuses GPUs with less than 32\,GiB free memory;
our processes use about 18\,GiB. For the MPC and search runs we executed the
loader's unchanged source with only that threshold lowered to 22\,GiB, and record
the source hash and the substitution in each receipt.
\item The first closed-loop smoke test failed before planning because the goal was
not passed to the agent, and a second one failed when replacing an action in the
planner's immutable result type; both were fixed before any task of the evaluation
run. The one task repeated between the smoke test and the evaluation run reproduces
bitwise-identical executed action sequences in both the baseline and the method
episode.
\item The Wall configuration lacks a top-level frame-skip entry that the PushT
pipeline sets; it was added before the first Wall task.
\item On Wall, the official CEM mean and its cheapest candidate can differ in
outcome, which never happened on PushT (0 of 258 task--variant pairs). The frozen
rule falls back to the cheapest candidate; we report it as specified and, separately
and labelled as not pre-registered for Wall, the closed-loop protocol's fallback to
the planner output.
\item At 19:55 the platform evicted our GPU container after the machine's local
storage was exhausted (our runs write only to shared storage, and their logs show no
storage error). All running processes were killed mid-task: 16 of 65 MPC tasks, 18 of
65 search-baseline tasks and 49 of 64 Wall tasks had finished, and no unfinished task
had produced an outcome. The replacement machine has NVIDIA H20 instead of H200 GPUs and
a different system image, so we rebuilt the software stack on shared storage at the
recorded package versions (with the official PyTorch 2.7.0 release in place of NVIDIA's
2.7.0 build) and froze an amendment before resuming (\cref{tab:supp-protocols}): finished
tasks are kept as recorded, unfinished tasks run once with identical inputs and seeds, and
all arms of a task run in one process, so every paired comparison stays on one machine.
Before resuming we checked the new machine: 1{,}200 previously executed action sequences
reproduce their recorded physical outcomes bitwise; on two finished tasks a CEM restart
reproduces the recorded H200 population bitwise, and recomputed planner costs differ by at
most $8\times10^{-6}$ (relative) with unchanged selections; one finished Wall task
reproduces all its outcomes and selections; and one finished MPC task reproduces its
baseline episode bitwise and its method episode to within $2.5\times10^{-4}$ in the
executed actions, with the same proposal wins and the same outcomes.
\item Two JEPA-WM Metaworld processes ran out of GPU memory on their first task
(other users occupied about 72\,GiB of those GPUs) before producing any outcome; they
were re-run with identical inputs on other GPUs and the failed attempts are archived.
\item In the goal-information runs (\cref{sec:supp-goalinfo}) two processes ran out of GPU memory while other users occupied the GPUs: one RoboCasa V-JEPA-2-AC process, whose unfinished tasks the relaunch script re-ran, and the first endpoint-CEM JEPA-WM process, which failed before its first episode finished; it was archived and re-run with identical inputs.
\item In the compute runs (\cref{sec:supp-compute}), two JEPA-WM Metaworld processes ran
out of GPU memory on their first task while other users occupied the GPUs, before producing
any outcome; they were archived and re-run with identical inputs. The launcher first
detected completion only from a log message that the Metaworld compute script does not
print, which delayed some launches; no outcome is affected.
\item The iCEM protocol builder first also expected development runs of the official CEM,
which had been dropped before launch; it stopped before writing anything and was rerun
without them, with the choice rule unchanged.
\item The analysis of the Metaworld seeding protocol for JEPA-WM filled its verdict field
with the criteria of the DINO-WM seeding protocol (M1, M2) instead of its own (J1, J2). The
counts are unaffected; a separate script evaluates J1 and J2 as registered on them
(\cref{tab:supp-ledger}).
\item The first image-feature index fed already normalized frames to the encoder, which
normalizes again; the development runs' render-versus-video check caught it, and the index
was rebuilt and the development runs repeated before the protocol was frozen
(\cref{sec:supp-visual}).
\item Several protocol files carry an approximate local creation time in their text;
\cref{tab:supp-protocols} lists the actual server write times, which, except for the
RoboCasa arm of the start-key protocol, precede every
run of each protocol.
\item During the averaging protocol the shared storage filled up (not with our files): one
result file was left empty and one launcher stopped. The affected parts, 4 DINO-WM tasks
and 92 JEPA-WM tasks, were re-run with identical inputs, writing to a local disk; the
240 results written before were complete and were kept, and no result had been read.
In the 42-family protocol one DINO-WM part ran out of GPU memory before writing any
result and was re-run.
\item In the within-PushT test (\cref{sec:supp-flip}) the host killed all our processes at
once while the full-precision runs were in progress (2026-11-09, 23:45--23:49). We had read
the outcomes of the 946 finished episodes when we reran the seven unfinished parts from the
start of their sampler streams; the 70 episodes finished in both runs are identical bit for
bit, and no choice depended on the interruption.
\item On 2026-11-10 the Metaworld environment could no longer be built on the shared machine: the OSMesa
library that MuJoCo renders with had disappeared from the container after the restart of the previous night (the
data volume was unaffected). We reinstalled the same system package before running the BC reference and the
frames of Fig.~1; no frozen result was recomputed with the new library.
\item In the start-key averaging protocol (\cref{sec:supp-goalinfo}), five of the six RoboCasa parts exited
silently about ten seconds after loading the predictor (no error, no out-of-memory record), while the part on
another GPU ran normally, and one of the 42 Metaworld parts stopped in the same minutes after writing its first
episode. We restarted the five on that GPU 13 minutes later and the three missing episodes the next morning; no
outcome had been read, and the written episode reproduces bit for bit when re-executed.
\end{itemize}

\end{document}

%% file: preamble.tex
\input{sections/flip_numbers}

\input{sections/bc_numbers}

%% file: sections/flip_numbers.tex
\newcommand{\FlipDfull}{86}      %
\newcommand{\FlipJfull}{85}      %
\newcommand{\FlipDfullPair}{$+0/{-}33$}
\newcommand{\FlipJfullPair}{$+1/{-}33$}
\newcommand{\FlipRule}{678}
\newcommand{\FlipWarm}{517}
\newcommand{\FlipAvg}{530}
\newcommand{\FlipRho}{0.96}
\newcommand{\FlipDfullP}{${<}10^{-9}$}
\newcommand{\FlipJfullP}{${<}10^{-8}$}
\newcommand{\FlipDfullCI}{$[-45,-21]$}
\newcommand{\FlipJfullCI}{$[-45,-21]$}
\newcommand{\FlipDfullSearch}{3/10}
\newcommand{\FlipJfullSearch}{6/11}

%% file: sections/bc_numbers.tex
\newcommand{\BCptFull}{41}
\newcommand{\BCptObj}{42}
\newcommand{\BCmwFull}{128}
\newcommand{\BCmwObj}{144}

\newcommand{\BCmwPairWarm}{$+42/{-}14$}
\newcommand{\BCmwPairAvg}{$+18/{-}17$}

%% file: sections/abstract.tex
\begin{abstract}
Visual world models such as DINO-WM and JEPA-WM plan with an optimizer that searches for actions whose predicted outcome has the lowest cost, its distance to a goal image.
A plan can thus fail in two ways: the optimizer never tries a good action that the cost would prefer (a search failure), or the cost, computed on predictions, ranks a bad action first (a scoring failure).
The two need different remedies, more search or a better model, but a planner does not record which one occurred.
We tell them apart by executing the optimizer's candidates, scoring a known solution with the planner's own cost, and comparing costs on predicted and real outcomes.
An error model explains when each arises and so how to use retrieved demonstration windows (\DProp): start the search from them (a warm start) when prediction errors are small, and average them without the model when large errors make the optimizer pick overrated actions.
Across four simulated environments and three world models, on short tasks whose goals the planner can represent, PushT and Wall fail mostly by search, while Metaworld and RoboCasa fail by scoring because the cost, sound on real outcomes (AUROC 0.94 on Metaworld), is at chance on predictions (0.54).
The diagnosis tells which remedy pays off: a four times larger search lowers the real-outcome cost only where failures come from search (clearly for DINO-WM), and where they come from scoring, averaging the windows without the model solves 127 of 168 new Metaworld tasks, against at most 87 when the model chooses among them.
Within PushT, degrading only the model flips which use is better, and our rule, frozen in advance, follows the flip and solves 678 of 960 new episodes, against 530 for the best fixed use.
\end{abstract}

%% file: sections/introduction.tex
\section{Introduction}
\label{sec:intro}

World models built on frozen self-supervised visual features can plan without rewards.
They predict the latent outcome of a candidate action sequence, and a sampling
optimizer, usually the cross-entropy method (CEM)~\cite{mc_cem,pets}, searches for the
sequence whose predicted outcome is closest to a goal
image~\cite{mc_dinowm,mc_jepawms,vjepa2}. When such a planner fails, the usual responses
are to search more or to improve what it optimizes: a better encoder, predictor or
objective~\cite{mc_objective_mismatch,mc_jepawms,mc_limits}.

Each response assumes a cause. The cost may rank a wrong action first, a \emph{scoring}
failure, or it may rank the right action first while the optimizer never samples it, a \emph{search} failure. The cost is computed on the
model's predicted outcome, so a scoring failure can occur even when the same cost ranks real outcomes well. The two need different remedies, yet they are rarely told apart, because
a planner does not record whether a successful action was among its candidates or whether its
cost would prefer a solution it never tried.

We record both (\cref{sec:diagnosis}). On PushT~\cite{mc_diffusion_policy}, with the
DINO-WM~\cite{mc_dinowm} checkpoint released by the JEPA-WMs study~\cite{mc_jepawms}, we execute in
the simulator the 900 candidates that CEM evaluates in three of its 30 iterations. We also
score the expert action sequence that generated each goal with the planner's own cost, a
\emph{known-solution probe} in the spirit of search-error tests in machine
translation~\cite{stahlberg2019}. Choosing better among the executed candidates could
rescue at most 3 of CEM's 20 failures, yet in 17 of them the cost prefers the known
solution to every candidate (\cref{fig:teaser,fig:diag}). The same holds for both models on the
official evaluation episodes, whose trajectories neither model saw, and on Wall navigation. These
planners fail by search. On Metaworld~\cite{metaworld} the probe finds the opposite: with either
model, the known solution is never cheaper than CEM's candidates. The cost itself is sound there:
applied to the frame the simulator reaches, it separates successes from failures (area under the
ROC curve, AUROC, 0.94), but applied to the model's prediction of that frame, it does not (0.54). Metaworld fails by scoring, and
so do two world models trained on real-robot data in the simulated RoboCasa kitchen~\cite{robocasa}.

We call these the \emph{search} and the \emph{scoring regime}. Both come from the same
source: a world model predicts well near its training data and worse away from it. While its
errors are small, a failure means that the right action was never sampled. Once they grow,
CEM selects the actions the model overrates most, and a larger search exploits these errors further
(\cref{sec:errormodel}). The regime
therefore suggests which remedies can work. It is read from the failures alone, so choosing
between warm-starting and averaging also depends on how often CEM fails (\cref{sec:method}). We wrote down what the regime predicts before measuring (\cref{sec:predict}). The regime also depends on the model: when only the PushT models are quantized, their failures
become scoring failures.

The diagnosis also suggests how demonstrations should enter the planner. We test this with
retrieved action windows whose start and end match the agent's current and goal proprioception (its own state, such as
position; \DProp, \cref{sec:method}). Where the model is right, the windows should help it search, either as
CEM's starting population (a \emph{warm start}) or added to its final candidates (\emph{Expand}).
Where the model misjudges, its choice is the problem even among demonstrations, so the average of
the nearest windows, executed without the model (\emph{Average}), should beat the window it ranks
first (\emph{Restrict}). Expand raises full success on the 96 official PushT
episodes from 62 to 78 for DINO-WM and from 68 to 82 for JEPA-WM, keeping all of CEM's full successes, and
has little effect on Metaworld and RoboCasa. On those two, Average beats the
model's choice among the windows, also in object success, and its advantage over the warm start
comes mostly from placing the hand. On PushT and Wall the warm start wins.
Within PushT, lower precision first turns failures into scoring failures and later flips the better use.

\noindent Our contributions are:
\begin{itemize}
  \item Three measurements separating search from scoring failures: executing CEM's candidates,
    scoring a known solution, and comparing predicted and real-outcome costs.
  \item Evidence in four environments and three world models that failure type depends on
    setting and model, explained by an error model checked on a fixed candidate set.
  \item Frozen tests of the diagnosis: a larger search lowers the real-outcome cost only in the search
    regime (clearly for DINO-WM), and the better use of retrieved demonstrations (\DProp) follows the regime.
\end{itemize}

%% file: sections/related.tex
\section{Related Work}
\label{sec:related}

\paragraph{Planning with visual world models.}
Latent dynamics models have long served model-predictive control from
pixels~\cite{planet,pets,mc_visual_foresight}. DINO-WM~\cite{mc_dinowm} plans by CEM
on frozen DINOv2~\cite{dinov2} patch features, JEPA-WMs~\cite{mc_jepawms} studied
encoders, predictors and planners for such models and released the checkpoints we use,
and V-JEPA~2~\cite{vjepa2} and Navigation World Models~\cite{nwm} also plan by sampling. These works report how often the system succeeds; we ask
why it fails.

\paragraph{Where model-based planning fails.}
A dynamics model trained for prediction need not rank actions by task
return~\cite{mc_objective_mismatch}, and optimizers that search a learned model exploit
its errors~\cite{mbpo}. Offline model-based methods counter this with
pessimism~\cite{mopo,morel,coms}; more generally, the best of many noisy estimates is
biased toward overestimation, the optimizer's curse~\cite{optimizerscurse}. In machine
translation, scoring the reference output under the model separates search errors from
model errors~\cite{stahlberg2019}. For latent world models, imagined action effects can
deviate from matched interventions~\cite{mc_gap}, terminal latent scoring can misdiagnose
model quality~\cite{vesessook2026hidden}, and a better objective repairs most failures on
TwoRooms with distant goals~\cite{singh2026objective}.
Concurrent work finds rankings reliable only when the goal lies within the imagined
horizon~\cite{mc_limits}, as in our tasks, and attributes lost decision
quality to prediction, metric or search, with a dominant source that changes with the
regime~\cite{li2026where}. These studies attribute a planner's average loss; we label each failure, judged by task success,
and test which remedies the labels say can help.

\paragraph{Proposals and warm starts for planning.}
Improved samplers such as iCEM~\cite{icem} and model predictive path integral control
(MPPI)~\cite{mppi} change how candidates are drawn. Learned policies serve as proposal distributions or warm starts for model-based
planning~\cite{poplin,mbop,tdmpc,tdmpc2}: GPC ranks diffusion-policy proposals with a
world model~\cite{gpc}, and SAGE refines subgoal-conditioned proposals with a frozen
one~\cite{sage}. Trajectory libraries warm-start optimization from stored
solutions~\cite{lightning,memmo}, and regularizing trajectory optimization toward the data
keeps a planner where its model is accurate~\cite{boney2019}. Closest to \DProp's warm start,
IMWM~\cite{imwm} shows that finite-budget CEM fails even with perfect dynamics and
initializes the planner from a retrieved demonstration through a learned intuition model.
Nearest-neighbor retrieval of demonstrations is a strong imitation baseline~\cite{vinn}. \DProp's warm
start does IMWM's initialization without training and leaves the cost unchanged.

%% file: sections/setup.tex
\section{Setting}
\label{sec:setup}

\paragraph{Environment and tasks.}
We diagnose short tasks with known solutions in four environments (Tab.~S1 in the supplement); here we
describe PushT.
PushT~\cite{mc_diffusion_policy} is a planar task in which a position-controlled
circular pusher pushes a T-shaped block to a goal pose. An action is a 2-D displacement of
the pusher's target. Following DINO-WM and JEPA-WMs, five simulator steps form one
planning step, and a plan has six planning steps (30 simulator steps). A \emph{window
task} is a 30-step window of a demonstration whose goal comes from the official
\emph{replay-expert} procedure: the window's actions are executed from its
start state, and the final frame, proprioception and state become the goal. The window's
actions are therefore a \emph{known solution} of the task. Our main evaluation uses the
\emph{official evaluation episodes} of JEPA-WMs~\cite{mc_jepawms}: 96 windows drawn with the official
sampler and seeds from the 21 trajectories of the validation split, which neither world model was trained
on. For development we use 64 training windows~\cite{mc_dinowm}, split by
demonstration source before any outcome was seen and used for every analysis and design decision.

\paragraph{World models and cost.}
We use the public PushT checkpoints of DINO-WM~\cite{mc_dinowm} and
JEPA-WM released by the JEPA-WMs study~\cite{mc_jepawms}, without any training. Both encode
$224\times224$ images with a frozen DINOv2 ViT-S/14~\cite{dinov2} into $16\times16$ patch
tokens and predict future tokens and proprioception from actions. The native planning
cost of an action sequence $a$ is the official terminal latent distance
\begin{equation}
  c(a) = \bigl\| \hat z^{\mathrm{vis}}_{H}(a) - z^{\mathrm{vis}}_{g} \bigr\|^2
       + \alpha \bigl\| \hat z^{\mathrm{prop}}_{H}(a) - z^{\mathrm{prop}}_{g} \bigr\|^2,
  \quad \alpha = 0.1,
\end{equation}
averaged over feature dimensions, where $\hat z_H$ is the prediction after $H{=}6$
planning steps and $z_g$ encodes the goal observation. All planners here use this
\emph{native cost} unchanged. Applied to the encoding of the reached observation instead of the prediction, it gives the \emph{real-outcome cost} $c_{\mathrm{real}}(a)$.

\paragraph{Planners.}
The official planner is CEM~\cite{mc_cem} with 300 samples, 30 iterations, 10 elites and
the previous mean inserted as a sample, i.e., 9{,}000 cost evaluations per plan. The official
open-loop planner executes all six planning steps of CEM's \emph{final mean}. We distinguish it from CEM's \emph{cheapest candidate}, the lowest-cost member of the
final population. We also use
an \emph{endpoint-constrained CEM} that samples only plans ending with the pusher at its
goal position (Supp.~S2).

\paragraph{Metrics.}
\emph{Object success} (\W) requires the block to end within 20 units (on the $512\times512$ canvas) and $20^{\circ}$ of its goal pose. \emph{Full success}
(\J), the official PushT criterion, bounds the joint position error of pusher and block
by 20 units, so the pusher must also end at its goal. Tasks are paired: $+a/{-}b$ means that one planner solves $a$ tasks the
other does not, and the other $b$. Where tasks share goal trajectories, task families or
demonstrations, 95\% intervals use these as units; otherwise, exact McNemar tests.

%% file: sections/diagnosis.tex
\section{Diagnosing Failures}
\label{sec:diagnosis}

Let $\mathcal{C}$ be the candidates the optimizer evaluated on a failed task and $\bar a$
its known solution. The failure is a \emph{search} failure if
$c(\bar a)<\min_{a\in\mathcal{C}}c(a)$, that is, if the cost prefers a solution the optimizer
never tried. Otherwise it is a \emph{scoring} failure: the cost, computed on the model's predictions,
ranks a tried candidate above a solution. $\mathcal{C}$ holds every candidate CEM evaluated, except
on the PushT development tasks, where we recorded 900, and on Wall, where we recorded the final
population. The label is local to the reference and to these candidates. A search label means that the
cost would have selected a known successful plan had the optimizer sampled it; it does not guarantee that
a larger search will succeed or that all candidates are ranked correctly. A scoring label does not rule
out a missed solution. A label describes one failure. A setting is in the search or the scoring
\emph{regime} when most of its failures carry that label. The regime summarizes the type of the failures;
how demonstrations should enter also depends on how often CEM fails (\cref{sec:method}). The environment alone does not fix the regime: a goal twice as far
turns half of DINO-WM's PushT failures into scoring failures (Supp.~S4), and so does a degraded model
(\cref{sec:predict}).

\subsection{Executing the candidate pool}
\label{sec:support}
On the 64 development tasks with DINO-WM, CEM evaluates 9{,}000 candidates per plan. We record the full populations of iterations
1, 15 and 30 and execute all 900 of them from the task's start state, for both CEM
variants (115{,}200 executions; Supp.~S4). The executions bound what any rescoring of these candidates can gain. With standard CEM
the cheapest candidate succeeds on 53 (\W) and 44 (\J) of 64 tasks, and an oracle that
picks a successful candidate whenever one exists reaches only 57 and 47. With
endpoint-constrained CEM no failure has a successful candidate, so the oracle gains
nothing. No reranker trained on the executed candidates beats the cost's choice either (Supp.~S4).

\subsection{Scoring the known solution}
\label{sec:expert}
Executing the known solution reproduces the goal on all 64 tasks. Among the 20 full-success failures of standard CEM, the known solution is cheaper than all
900 recorded candidates in 17; among the 8 failures of endpoint-constrained CEM, in all 8.
For object success the numbers are 8 of 11 and 8 of 8. CEM thus converged to a region that is worse under its own cost. The three scoring failures are exactly the three tasks that the
oracle rescues, so the label matches the executed candidates on all 28 failures.

\begin{figure}[t]
\centering
\includegraphics[width=\linewidth]{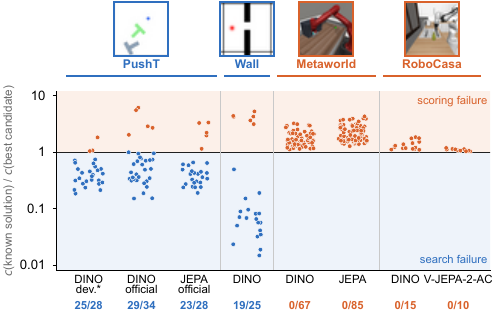}
\caption{\textbf{The known-solution probe across settings.} One dot per full-success failure of CEM, on a
log scale: the cost of the known solution divided by that of the cheapest candidate the optimizer evaluated, both
on predicted outcomes. Below 1 the cost prefers a solution the optimizer missed (search failure); at or above 1 it
prefers the optimizer's failing choice (scoring failure). Counts: search failures / all failures.
Candidates: all evaluated, except the 900 recorded on PushT development tasks (dev.; $^{\ast}$both CEM
variants) and Wall's final population. Official: official evaluation episodes. DINO, JEPA: DINO-WM, JEPA-WM.}
\label{fig:diag}
\end{figure}

\paragraph{Unseen trajectories, a second model, a second environment.}
On the official evaluation episodes, whose trajectories neither model saw, the probe gives the same answer:
the known solution is cheaper in 29 of 34
full-success failures with DINO-WM and in 23 of 28 with JEPA-WM (\cref{fig:diag}). Wall
(\cref{sec:wall}) samples its goals, so we script a known solution through the door. It
reaches the goal on all 64 tasks and is cheaper than CEM's cheapest candidate in 19 of 25
failures. The label also holds with
other successful plans as references. It is unchanged on all 20 failures, 49 of 50 on Metaworld
and 170 of 187 with quantized PushT models, and none of 739 successful RoboCasa windows turns a scoring label into a search one (Supp.~S4, S8, S9, S19).

\subsection{Why the two regimes arise}
\label{sec:errormodel}
Write the native cost, computed on predicted outcomes, as
$c(a)=c_{\mathrm{real}}(a)+\epsilon(a)$, the real-outcome cost plus a prediction error,
and let the planner execute $\hat a=\arg\min c$ over $N$ candidates. Assume that the errors
have zero mean, are $\sigma$-sub-Gaussian and are independent of the real-outcome costs, and that the
candidate set is fixed independently of the errors. Under these assumptions, three propositions relate
the error scale to the choice, to the probe and to averaging windows (proofs in Supp.~S18, Props.~S1--S2,
S3 and S5). An error shared by all candidates changes no choice, so $\sigma$ is the scale of the part
that varies across them.

\begin{proposition}[Choice]\label{prop:choice}
$\mathbb{E}\,c_{\mathrm{real}}(\hat a)-\min c_{\mathrm{real}}\le2\sigma\sqrt{2\log N}$ and
$\mathbb{E}\,c(\hat a)\le\min c_{\mathrm{real}}\le\mathbb{E}\,c_{\mathrm{real}}(\hat a)$: the
choice is near-optimal for small $\sigma$, and when the candidates differ little in reality
it looks better than it is, by about $\sigma\sqrt{2\log N}$ for independent Gaussian errors (Prop.~S2).
\end{proposition}

\begin{proposition}[Probe]\label{prop:probe-main}
Let a known solution $\bar a$ have real cost 0 and every evaluated candidate in
$\mathcal{C}$ real cost at least $\Delta$. A scoring label, $c(\bar a)\ge\min_{\mathcal{C}}c$,
implies $\epsilon(\bar a)-\min_{\mathcal{C}}\epsilon\ge\Delta$; a search label holds with
probability at least $1-\delta$ once $\Delta\ge\sigma\sqrt{8\log(2(|\mathcal{C}|+1)/\delta)}$.
\end{proposition}

\begin{proposition}[Choose or average]\label{prop:avg}
Let windows $w_k=w^\ast+b+\eta_k$ scatter around a solution $w^\ast$ with a shared offset $b$
and i.i.d.\ zero-mean deviations of covariance $\Sigma$, and let the real cost be
$(w-w^\ast)^\top Q(w-w^\ast)$ near $w^\ast$. The mean of $K$ windows has expected real cost
$b^\top Qb+\mathrm{tr}(Q\Sigma)/K$. A choice that ignores the deviations has expected real cost
$b^\top Qb+\mathrm{tr}(Q\Sigma)$, while a choice by the real cost itself can fall below $b^\top Qb$.
\end{proposition}

\noindent These propositions assume a fixed candidate set. Because CEM adapts its samples, we treat them
as predictions for the planner, not guarantees, and test them in \cref{sec:experiments,sec:predict}. Where $\sigma$ is small against the gap between good and
bad candidates, a failure means that no good candidate was tried (\cref{prop:choice}), and the probe says
so (\cref{prop:probe-main}). The remedy is then better candidates: a larger search, or demonstrations
added to the pool or placed at its start, where CEM's refinement can also correct the offset that their
average keeps (\cref{prop:avg}). Where $\sigma$ is large, the cheapest of many candidates is mostly the
most overrated one, the optimizer's curse~\cite{optimizerscurse}. Added candidates are then rarely
chosen, a larger search lowers mostly the predicted cost, and choosing among more candidates from the
same set stops helping (\cref{fig:pool}). Plans that stay with the demonstrations escape part of this: their average cancels part of the windows'
scatter without any choice (\cref{prop:avg}), and a search started from them stays where $\sigma$ is
smaller (cf.~\cite{boney2019}).

%% file: sections/method.tex
\section{DemoProp: Testing the Diagnosis with Demonstrations}
\label{sec:method}

\begin{figure*}[t]
\centering
\includegraphics[width=\linewidth]{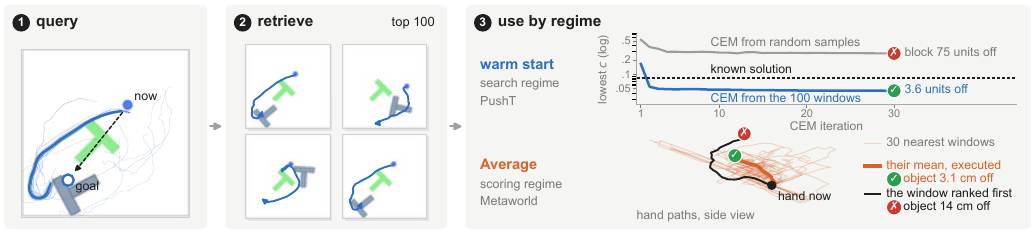}
\caption{\textbf{DemoProp.} (1) Query: the pusher now and at the goal. (2) Four of the 100 retrieved windows (PushT task 51). (3) Search regime (official PushT episode
0, DINO-WM): CEM started from the windows finds a plan with a lower predicted cost than the known
solution, and the block ends
3.6 units from its goal on the 512-unit canvas, against 75 from random samples. Scoring regime (Metaworld pick-out-of-hole, new
task): the mean of the 30 nearest windows leaves the object 3.1\,cm from its goal, the window ranked first 14\,cm. \checkmark/$\times$: full success.}
\label{fig:method}
\end{figure*}

The diagnosis says whether the model can be trusted to choose, and so how demonstrations might
enter the planner. To test this, \DProp retrieves windows from demonstrations and uses them as each regime suggests. Two further uses of them serve as controls (\cref{fig:method};
Supp.~S1).

\paragraph{Retrieval.}
We index every window of $T$ consecutive actions in the demonstrations (stride 2;
893{,}345 windows for $T{=}30$ on PushT). A window starting at time $t$ has the
proprioceptive key
\begin{equation}
  k(w) = \bigl[\,0.25\,p_t,\; p_{t+T}-p_t,\; 0.5\,v_t,\; 0.5\,v_{t+T}\,\bigr],
  \label{eq:key}
\end{equation}
where $p$ and $v$ are the agent's position and velocity (on Wall and Metaworld, position terms only). The query is the same expression with
the current and the goal proprioception, and the $K{=}100$ nearest windows in Euclidean
distance are the proposals $\mathcal{P}_{\mathrm{demo}}$. The planner sees both
inputs on PushT, Wall and Metaworld, where its cost weights goal proprioception by
$\alpha{=}0.1$; image-only RoboCasa models lack it.

\paragraph{Uses of the windows.}
In the search regime the windows help CEM search: a \emph{warm start} swaps 100 samples of
CEM's first population for them, at the same 9{,}000 evaluations. In the scoring regime the model's
choice is unreliable, so \DPAvg executes the mean of the 30 nearest windows (on Metaworld, from all
42 tasks) and uses no model. The two controls use the model's cost. \DPExp scores the windows with the
unchanged native cost $c$, pools them with the 300 final candidates and executes
$a^\star=\arg\min_{a\in\mathcal{P}_{\mathrm{CEM}}\cup\mathcal{P}_{\mathrm{demo}}}c(a)$, at
1--2\% more evaluations. If no window has the lowest cost, Expand executes CEM's cheapest candidate.
It therefore never alters CEM's search and can help only where the cost prefers a window to every CEM
candidate; the probe's label alone does not tell whether that happens (Supp.~S4, S11). \DPRes executes the window that $c$ ranks first, the model's choice among
demonstrations, which the average should beat where the model misjudges. No use trains anything, uses a reward or object
state, or sees the known solution of an evaluation task: retrieval excludes all demonstrations from its
source trajectory (on PushT, 101 noise variants).

\paragraph{Choosing the use.}
We read the regime from development tasks on PushT and Metaworld and, before the use was frozen,
from the evaluation tasks on Wall and RoboCasa (Supp.~S17, S19). PushT and Wall are in the search regime,
Metaworld and RoboCasa in the scoring regime, and JEPA-WM inherits each environment's regime. The regime
alone cannot fix the use for a setting: the warm start keeps most of CEM's successes, whereas Average
replaces CEM on every task. Let $s$ be the share of all development tasks that CEM fails by scoring
(the regime looks only at failures) and $p_{\mathrm{avg}}$ the share that Average solves. We use the warm start if $s<1-p_{\mathrm{avg}}$, a heuristic
(Supp.~S19). This rule follows the regime in the four environments but not for two
degraded PushT models (\cref{sec:predict}). Comparing CEM's success rate with Average's, or trying
both uses, picks the same use in every setting we ran, so the diagnosis is no better way to choose a use.
It instead explains why the better use switches, which remedies do little, and which use tends to repair each failure.

%% file: sections/experiments.tex
\section{Experiments}
\label{sec:experiments}

This section diagnoses each environment; \cref{sec:predict} tests predictions across settings. The work proceeded in rounds (Supp.~S3): Expand and Restrict were
registered first, the warm start entered as a baseline, and Average was found on Metaworld tasks that then
became development tasks. Except for one arm, all protocols were frozen before their runs; that arm
started 34 minutes early, and its outcomes had not been read when its protocol was frozen (Supp.~S10). Confirmations therefore come from episodes and tasks never used
before: Expand on the official PushT episodes, Average and its matched controls on new Metaworld tasks, and
the within-PushT test on new episodes. The cross-environment and compute tests reuse evaluation tasks, from
which the regimes of Wall and RoboCasa were also read.

\subsection{Official evaluation episodes}
\label{sec:main-results}

\begin{table}[t]
\centering\small
\setlength{\tabcolsep}{2.2pt}
\begin{tabular}{lccccc}
\toprule
& & \multicolumn{2}{c}{DINO-WM} & \multicolumn{2}{c}{JEPA-WM}\\
\cmidrule(lr){3-4}\cmidrule(lr){5-6}
Planner & WM evals. & \W & \J & \W & \J\\
\midrule
CEM (official) & 1$\times$ & 79 & 62 & 80 & 68\\
CEM, $4\times$ population & 4$\times$ & 87 & 70 & 81 & 70\\
CEM, endpoint$^{\ast}$ & 1$\times$ & 75 & 75 & 80 & 80\\
iCEM$^{\dagger}$ & 1$\times$ & 85 & 71 & 90 & 78\\
\midrule
nearest window, no model & 0 & 43 & 39 & 43 & 39\\
\DPRes & 0.01$\times$ & 70 & 62 & 76 & 62\\
CEM, warm start$^{\ddagger}$ & 1$\times$ & \textbf{91} & 80 & 86 & 79\\
\DPExp & 1.01$\times$ & 86 & 78 & 88 & 82\\
\DPExp{} + endpoint CEM$^{\ast}$ & 1.01$\times$ & 81 & 80 & 85 & 84\\
\DPExp{} + iCEM$^{\dagger}$ & 1.01$\times$ & 90 & \textbf{82} & \textbf{93} & \textbf{87}\\
\midrule
BC policy (reference) & 0 & \BCptObj & \BCptFull & \BCptObj & \BCptFull\\
known solution (reference) & 0 & 96 & 96 & 96 & 96\\
\bottomrule
\end{tabular}
\caption{\textbf{Official PushT evaluation episodes}: successes out of 96 per model (\W: object,
\J: full success; validation goals; protocol frozen before any plan).
Bold: best planner per column. WM evals.: world-model evaluations per plan relative to
CEM; all world-model planners use the same cost. Nevergrad, MPPI and gradient descent:
Supp.~S3. $^{\ast}$Ends the pusher at its goal, which needs its known dynamics
(\cref{sec:goalinfo}). $^{\dagger}$Colored-noise sampling tuned on development tasks.
$^{\ddagger}$\cref{sec:method}. BC: behavior cloning on the windows (Supp.~S20).}
\label{tab:main}
\end{table}

\DPExp raises full success from 62 to 78 with DINO-WM and from 68 to 82 with JEPA-WM, and
it loses none of CEM's full successes ($+16/{-}0$ and $+14/{-}0$; $p<10^{-3}$ with trajectories
as units; Supp.~S3, S15). Counted afterwards, all 30 repairs fall on search failures, and none of the
10 scoring failures is repaired. Retrieval alone does not explain the gain: the nearest window solves 39,
and Restrict 62.
\label{sec:search}iCEM~\cite{icem} and a four times larger
population beat CEM but stay below \DPExp in full success (Supp.~S11, S12); the warm start
matches \DPExp but loses 9 of CEM's successes.

\paragraph{Wall.}\label{sec:mpc}\label{sec:wall}
On Wall~\cite{mc_dinowm,mc_jepawms} a 2-D agent must cross a wall through its single
door to a goal sampled by the environment (64 tasks, DINO-WM). CEM solves
39: it often stays on the start side, although the scripted path through the door is
cheaper under its own cost in 19 of 25 failures (\cref{sec:expert}). \DPExp raises success
to 47 ($+8/{-}0$ against its fallback), all 8 repairs on search failures, and the warm start
reaches 59; averaged windows, which pass the door at different times, solve 3 (Supp.~S7, S17).

\subsection{Metaworld}
\label{sec:metaworld}
On the 3-D manipulation benchmark Metaworld~\cite{metaworld} we use the checkpoints of JEPA-WMs~\cite{mc_jepawms}
with their official CEM (300 samples, 15 iterations, horizon 6) and seven of the 42 tasks in their training
data, from reach to faucet-open; the 300 demonstrations per task form the index, keyed by hand position and
gripper opening.

\paragraph{Diagnosis.} The official protocol places the goal 100 steps ahead of a 6-step planning
horizon. No plan can reach it, so the cost cannot rank plans by what matters~\cite{mc_limits}: in the 44
failures the expert's first 30 actions are never cheaper than CEM's candidates, and \DPExp changes
nothing. A diagnosis needs tasks whose solution the planner can represent, so we build window tasks as for
PushT, from the 30-step window of the expert's rollout in which the object moves most. Success requires
the object (\W), and also the hand (\J), to end within 5\,cm. With DINO-WM, CEM's final mean reaches object
success on 80 of 112 tasks and full success on 45. In each of its 67 full-success failures, at least one candidate CEM evaluated costs no more than the
known solution. As predicted, \DPExp selects a window in one task, none with JEPA-WM, and otherwise
differs from CEM only by falling back to the cheapest candidate (Supp.~S8).

\paragraph{Why the cost prefers CEM's choice.} We measure the real-outcome cost of every
executed plan (Supp.~S8). CEM's cheapest
candidate fails 68 tasks in full success, one more than its final mean, and in all 68 the real-outcome
cost exceeds the predicted one, by a median factor of 2.3. The known solution,
whose real-outcome cost is 0, is predicted at 0.58, above CEM's choices at 0.32, so the errors are larger
than the gap, as a scoring label requires (\cref{prop:probe-main}). On real outcomes the cost separates
successes from failures (AUROC 0.94 over DINO-WM's executed plans, without the known solution); on
predictions it is at chance (0.54). Within a task it orders 98\% of success--failure pairs correctly on
real outcomes and 25\% on predictions, and 99\% and 34\% where only the hand differs (Supp.~S8). For
the hand, then, the cost is informative on real outcomes but not on predicted ones. CEM's choice exploits
these errors, as the decoded prediction in Fig.~\ref{fig:teaser} shows.

\paragraph{Windows.} Restrict raises full success from 45 to 78 of 112 with DINO-WM and from 27
to 81 with JEPA-WM, but trails the warm start on new tasks of all 42 families (\cref{sec:predict}). The
warm start solves 88 with DINO-WM, beating the prediction we had registered, and 72 with JEPA-WM. It
reaches the predicted cost of unconstrained CEM but stays in the demonstrated region, where low cost goes
with success.

\subsection{RoboCasa with real-robot world models}
\label{sec:robocasa}

RoboCasa~\cite{robocasa}, a photorealistic kitchen simulator, tests two world models trained on
real-robot DROID data~\cite{droid}, a DINO-WM and the 1-billion-parameter V-JEPA-2-AC~\cite{vjepa2}, each
with its official planner. The 14 released demonstrations of picking a bottle give 90 window tasks of the
reach and pick phases, built as for PushT but 15 steps long (Supp.~S9); full success requires the hand and
the object to end within 5\,cm of their goals. Zero-shot, CEM solves 75 and 80 of them, and standing still
solves 69. The object has to move by more than 1\,cm in only 9 tasks, and many
windows move the hand by only 3--5\,cm, so the tasks mostly test reaching. The probe finds only scoring
failures: Expand selects no window, and \DPRes and the warm start gain little. On development tasks the
cost ranks executed windows well on real outcomes (AUROC 0.89 and 0.97) but near or below chance on
predictions (0.55 and 0.38), so a choice helps little (\cref{fig:pool}). Average and
the nearest window, which use no model, each solve 88 ($+2/{-}2$ against each other); on the 23 tasks that an
all-zero plan fails, each solves 21, against 12 for the warm start with either model. Both retrieve with the goal end-effector position, which the planner never
receives. RoboCasa thus replicates the scoring diagnosis with real-robot models but cannot
separate averaging from retrieval.

\subsection{What the key needs}
\label{sec:goalinfo}
The key contains the goal proprioception, which full success also scores. Three protocols separate this information from the demonstrations (Supp.~S2, S10).
Without the goal term, \DPRes keeps most of its Metaworld gain (71 and 69 instead of 78 and 81), while
the nearest window falls from 72 to 22. Average needs the goal term most: with the start key it solves
13 of the 168 new Metaworld tasks instead of 127, and 60 instead of 88 on RoboCasa (Supp.~S10). In most of
Average's extra full successes over the warm start, the warm start had already placed the object and only
the hand was off: in 26 of 47 with DINO-WM and 28 of 45 with JEPA-WM on Metaworld, and in 7 and 9 of 11
with DINO-WM and V-JEPA-2-AC on RoboCasa. The goal term supplies the hand's goal, which the predictions
miss. On Metaworld, Average therefore clearly leads the warm start in full success and in hand success,
judged on the hand alone ($+49/{-}14$ with DINO-WM and $+48/{-}16$ with JEPA-WM). In object success its
lead cannot be told from chance with task families as units ($+23/{-}14$ and $+19/{-}15$), yet against
Restrict, the model's choice among the same 30 windows, it leads in object success too ($+33/{-}3$ and
$+34/{-}3$; Supp.~S17).

%% file: sections/predict.tex
\section{What the Diagnosis Predicts}
\label{sec:predict}

\paragraph{More compute.} By \cref{prop:choice}, a larger pool lowers the real-outcome cost of the choice
when $\sigma$ is small but selects larger errors when $\sigma$ is large (Supp.~S18, Prop.~S4). A four times
larger CEM is a new run rather than a larger pool, so we test it directly. Before measuring any real-outcome cost of the larger search,
we predicted that the rank correlation between the changes in predicted and real-outcome cost would exceed
0.2 on PushT and stay below it on Metaworld, for each model (Supp.~S12). All four predictions held: 0.51 and 0.24
on PushT, $-0.10$ and $-0.02$ on Metaworld, where the larger search lowers the predicted cost on 108 of
112 tasks. With trajectories and families as units, the environments differ clearly for DINO-WM; for JEPA-WM
the difference has the predicted sign, but its interval includes zero ($[-0.10,0.61]$). Full success follows the costs: it rises on PushT
from 62 to 70 with DINO-WM and from 68 to 70 with JEPA-WM, and falls on Metaworld from 45 to 43 with
DINO-WM. With JEPA-WM on Metaworld it rises from 27 to 34 ($p{=}0.21$), above the net gain of three we had allowed.

\begin{figure}[t]
\centering
\includegraphics[width=\linewidth]{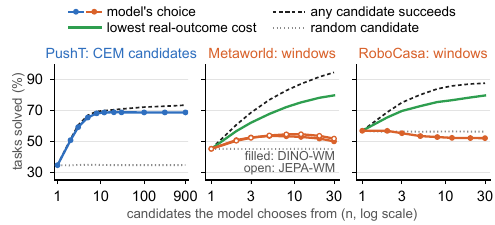}
\caption{\textbf{Choosing from more candidates of the same set} (post hoc; 400 random subsets of $n$
per task). Curves: tasks solved (full success) by the model's choice (lowest predicted cost), by the lowest real-outcome
cost, by any candidate of the subset, and by a random one. PushT: 900
recorded CEM candidates of 64 development tasks (DINO-WM). Metaworld: 30 windows of 168 new tasks. RoboCasa: 50 windows of 49 development tasks (DINO-WM). Blue: search regime;
orange: scoring regime.}
\label{fig:pool}
\end{figure}

\paragraph{More candidates from the same set.} A larger CEM run also samples elsewhere. To change only the
number of candidates, we let the model choose among random subsets of one fixed set of executed candidates: CEM's recorded ones or the retrieved windows (\cref{fig:pool}; post hoc, Supp.~S18). On PushT the choice finds nearly every success that a subset holds,
69\% of tasks with all 900 candidates against 73\% with a success among them, so what fails there is the
search. On Metaworld and RoboCasa, 30 windows hold a success in 95\% and 88\% of tasks, and the same cost on
real outcomes picks one in 80\%. Yet the model's choice stays close to a random window: on Metaworld it
solves 50\% of tasks with DINO-WM and 52\% with JEPA-WM, against 45\% for a random window, and on RoboCasa
52\% with DINO-WM, against 56\%. Where the model ranks a Metaworld task's windows best, more windows help (47\% to 63\% with DINO-WM); where it ranks them worst, they hurt (42\% to 32\%),
as \cref{prop:choice} expects.

\paragraph{Which use of the windows works.} A protocol frozen before averaging was run on PushT, Wall or
RoboCasa (Supp.~S17) predicted that averaging loses to the warm start by at least 10\% in the search regime
and does not lose in the scoring regime. All four predictions held, each reliably with trajectories, tasks,
families or demonstrations as units. In full success, Average solves 44 episodes against 80 and 79 for the
warm start on PushT, 3 against 59 on Wall, 127 against 100 and 102 on Metaworld's new tasks, and 88 against
77 on RoboCasa.
Averaging loses on PushT and Wall, where the path must be precise, and wins on Metaworld and RoboCasa,
where the model misjudges, as \cref{prop:avg} suggests. One protocol gave every use the same 30 windows on the 168 new Metaworld tasks (Supp.~S17): a warm start
from them solves 92 and 89, Restrict 84 and 87, and their average 127.

\begin{table}[t]
\centering\footnotesize
\setlength{\tabcolsep}{2.7pt}
\begin{tabular}{lcccc|cccc}
\toprule
& \multicolumn{4}{c|}{development (shares)} & \multicolumn{4}{c}{new episodes (of 96)}\\
& \multicolumn{2}{c}{DINO-WM} & \multicolumn{2}{c|}{JEPA-WM} & \multicolumn{2}{c}{DINO-WM} & \multicolumn{2}{c}{JEPA-WM}\\
\cmidrule(lr){2-3}\cmidrule(lr){4-5}\cmidrule(lr){6-7}\cmidrule(lr){8-9}
predictor & $f$ & $s$ & $f$ & $s$ & warm & Avg. & warm & Avg.\\
\midrule
full precision & .22 & .06 & .33 & .06 & \textbf{\FlipDfull} & 53 & \textbf{\FlipJfull} & 53\\
6 bits & .80 & .38 & .00 & .00 & \textbf{57} & 53 & \textbf{85} & 53\\
5 bits & .90 & .59 & .38 & .09 & 44 & \textbf{53} & \textbf{86} & 53\\
4 bits & .91 & .91 & .73 & .25 & 6 & \textbf{53} & \textbf{67} & 53\\
3 bits & 1.0 & 1.0 & 1.0 & 1.0 & 0 & \textbf{53} & 1 & \textbf{53}\\
\bottomrule
\end{tabular}
\caption{\textbf{Within PushT, the failures and the better use move with the model.} Left, on 32
development tasks: $f$, the share of CEM's failures with the scoring label (above one half: scoring
regime), and $s$, the share of all tasks that CEM fails by scoring; averaging solves
$p_{\mathrm{avg}}{=}0.41$ of these tasks. Right: full successes of the warm start and of Average on 96 new
episodes; bold: the use the rule chose beforehand ($s<1-p_{\mathrm{avg}}$).}
\label{tab:flip}
\end{table}

\paragraph{Within one environment.} So far each environment showed a single regime, so the regime
might only name the environment. We therefore kept PushT, its tasks, windows and planner fixed and
degraded only the model, quantizing its predictor to 6, 5, 4 and 3 bits (Supp.~S19). On the development tasks, most failures
become scoring failures from 6 bits on for DINO-WM and from 4 bits on for JEPA-WM ($f$ in \cref{tab:flip}).
On 96 new official episodes, the better use flips at a lower precision. With goal trajectories as
units, the warm start is reliably better at full precision with both models and at 6 and 5 bits with
JEPA-WM, and Average is reliably better at 4 and 3 bits with DINO-WM and at 3 bits with JEPA-WM. Next to the switches, the two uses cannot be told apart reliably: the warm start solves
57 and 44 episodes against Average's 53 at 6 and 5 bits with DINO-WM, and 67 against 53 at 4 bits with
JEPA-WM. The results show the reversal but do not locate it precisely. The warm start keeps CEM's
successes, so it holds up in the scoring regime while CEM fails few tasks by scoring (small $s$). The rule of \cref{sec:method}, which weighs $s$, was frozen before the
new episodes. It picks the better use by point estimate in all ten settings, reliably in seven, and solves
\FlipRule{} episodes in total, against \FlipWarm{} for always warm-starting and \FlipAvg{} for always
averaging. With object success alone the flip remains, for DINO-WM at a lower precision in point estimates
(Supp.~S19).

\paragraph{Failure by failure.} We also froze a test of whether the label tells which use repairs a
given failure of CEM: the warm start for a search label, Average for a scoring label. On the 96 new
episodes it does. Among the 188 failures that exactly one use repairs, the use the label indicates repairs
151, against 135.9 when labels are permuted within settings (interval of the excess $[5.9,22.6]$ with goal
trajectories as units; DINO-WM $p{<}0.001$, JEPA-WM $p{=}0.29$). The result holds when labels are permuted
only within thirds of CEM's block or pusher error. Of these failures, the warm start repairs 71
search-labeled ones and Average 5; Average repairs 80 scoring-labeled ones and the warm start 32. That
difference comes from the pusher (post hoc): where both uses placed the block, Average repairs 54 and the
warm start 1; where only one did, 26 and 31.
Choosing the use by the label repairs 277 of 446 failures, against 287 for the rule (Supp.~S19). The label
thus helps explain single failures, clearly for DINO-WM, but does not choose uses better than the rule.

%% file: sections/discussion.tex
\section{Conclusion}
\label{sec:discussion}
World-model planners fail either because CEM misses solutions that its cost prefers or because it picks
actions that the model overrates. A few tasks with known solutions tell which, and so which remedies can
help: more search and more candidates in the first case, plans that stay with the demonstrations or a
better model in the second. With retrieved demonstrations, the better use follows the regime across our
environments and, within PushT, flips once scoring failures are frequent. Acting on the regime pays off:
where the model misjudges, averaging without it solves 127 of 168 new Metaworld tasks, against at most 87
for its choice among the same windows, and within PushT our rule solves \FlipRule{} of 960 new episodes,
against \FlipAvg{} for the best fixed use. Choosing by success rates gives
the same uses, so the diagnosis explains these decisions rather than making them better.

\paragraph{Limitations.} The search regime rests on two simple 2-D tasks, PushT and Wall (Wall with
DINO-WM only); all environments are simulated, and all models but V-JEPA-2-AC share DINOv2 features.
Labels are relative to a reference and to the candidates compared (\cref{sec:diagnosis}), and the probe
needs a known solution for each diagnosed task, here a demonstration window or, on Wall, a script. Metaworld is
tested on window tasks, not in its official long-horizon protocol (Supp.~S8). The warm start and Average
were run open loop, one plan per task; only Expand was also tested in closed loop (Supp.~S7). Plans that work in the
scoring regime need dense demonstrations and the goal position in the key. Most of Average's advantage
over the warm start comes from placing the hand; on RoboCasa, Average ties the nearest window, and on
Metaworld a policy trained on the demonstrations matches it (Supp.~S9, S20).

%% file: sections/llm_statement.tex
We used OpenAI Codex and Anthropic Claude Code for English language polishing of the
manuscript and to write the experiment scripts. All ideas, experimental design, and
conclusions are the authors' own, and the authors take full responsibility for the
content of this paper.

%% file: supp_predictions_table.tex
\begin{table*}[tp]
\centering\footnotesize
\setlength{\tabcolsep}{3pt}
\begin{tabular}{@{}p{0.130\textwidth}p{0.040\textwidth}p{0.550\textwidth}p{0.240\textwidth}@{}}
\toprule
Protocol & Label & Prediction & Outcome\tabularnewline
\midrule
\raggedright Metaworld & \raggedright A1 & \raggedright Among tasks where the CEM mean fails FULL, the expert window is cheaper than every CEM candidate evaluated in all 15 iterations in fewer than half. & \raggedright held (0 of 67)\tabularnewline
\raggedright & \raggedright A2 & \raggedright The frozen rule (task) selects a proposal in at most 10\% of tasks, and its pooled OBJ and FULL successes differ from the CEM mean by at most 3. & \raggedright held (1 of 112; net $-1$ \W, 0 \J)\tabularnewline
\raggedright & \raggedright A3 & \raggedright task\_only has more FULL successes than the CEM mean: pooled gains exceed losses (exact two-sided McNemar reported). & \raggedright held ($+43/{-}10$)\tabularnewline
\raggedright & \raggedright A4 & \raggedright task\_only versus task\_nn and all\_only versus all\_nn are reported as paired comparisons whatever their direction; development suggested a small advantage for the world-model ranking. & \raggedright no direction ($+16/{-}10$; all tasks $+28/{-}16$)\tabularnewline
\raggedright & \raggedright A5 & \raggedright A3 and A4 are repeated with task\_only\_x and task\_nn\_x. & \raggedright held (A3: $+46/{-}10$; A4, no direction: $+18/{-}10$)\tabularnewline
\raggedright & \raggedright B1 & \raggedright The expert prefix is cheaper than every evaluated CEM candidate in fewer than half of the failed episodes. & \raggedright held (0 of 44)\tabularnewline
\raggedright & \raggedright B2 & \raggedright Proposals win at most 10\% of replanning steps and pooled success changes by at most 3 episodes. & \raggedright held (0 of 392 steps; 12 vs.\ 12)\tabularnewline
\midrule
\raggedright Metaworld, JEPA-WM & \raggedright J1 & \raggedright Among tasks where the CEM mean fails FULL, the expert window is cheaper than every CEM candidate evaluated in fewer than half. & \raggedright held (0 of 85)\tabularnewline
\raggedright & \raggedright J2 & \raggedright The PushT rule (task) selects a proposal in at most 10\% of tasks and changes pooled OBJ and FULL by at most 3. & \raggedright held (0 of 112; net $+1$ \W, $-1$ \J)\tabularnewline
\raggedright & \raggedright J3 & \raggedright task\_only has more FULL successes than the CEM mean (gains exceed losses; exact two-sided McNemar reported). & \raggedright held ($+60/{-}6$)\tabularnewline
\raggedright & \raggedright J4 & \raggedright task\_only versus task\_nn and all\_only versus all\_nn are reported whatever their direction. & \raggedright no direction ($+15/{-}6$; all tasks $+30/{-}8$)\tabularnewline
\raggedright & \raggedright J5 & \raggedright J3 and J4 with task\_only\_x and task\_nn\_x. & \raggedright held (J3: $+60/{-}6$; J4, no direction: $+15/{-}7$)\tabularnewline
\midrule
\raggedright Official PushT episodes & \raggedright P1 & \raggedright per model: among episodes where the official CEM output fails, the known solution costs less than every candidate CEM evaluated in more than half (PushT stays search-limited on unseen trajectories) & \raggedright held for both models (29 of 34 and 23 of 28)\tabularnewline
\raggedright & \raggedright P2 & \raggedright pooled over both models (192 paired episodes): \DPExp J $>$ CEM output J, exact one-sided McNemar $p<0.05$ & \raggedright held ($+30/{-}0$, $p{=}9{\times}10^{-10}$)\tabularnewline
\raggedright & \raggedright P3 & \raggedright pooled: \DPExp J $>$ CEM cheapest-final J, one-sided McNemar $p<0.05$ & \raggedright held ($+30/{-}0$, $p{=}9{\times}10^{-10}$)\tabularnewline
\raggedright & \raggedright P4 & \raggedright pooled: \DPExp J $>$ nearest window J ($p<0.05$) and Expand wins $\geq$ losses against Restrict & \raggedright held (nearest window $+82/{-}0$; Restrict $+39/{-}3$)\tabularnewline
\raggedright & \raggedright P5 & \raggedright pooled: \DPExp wins $\geq$ losses (J) against each matched-budget baseline: warm, mppi, ng, gd; cem4x reported without a prediction & \raggedright held (warm $+10/{-}9$; MPPI $+157/{-}0$; NG $+80/{-}6$; GD $+142/{-}0$)\tabularnewline
\raggedright & \raggedright P5b & \raggedright pooled: the same against the best-scoring selection rule of each baseline (output, cheapest final candidate, MPPI mean), chosen after the fact, which favours the baseline & \raggedright held (warm $+10/{-}9$; MPPI cheapest final candidate $+86/{-}2$; NG $+80/{-}6$; GD $+142/{-}0$)\tabularnewline
\raggedright & \raggedright P6 & \raggedright sign-flip permutation over the 21 validation trajectories of the pooled Expand $-$ CEM-output J difference, exact one-sided $p<0.05$ & \raggedright held ($p{=}5{\times}10^{-4}$)\tabularnewline
\raggedright & \raggedright P7 & \raggedright conditional: for a model where P1 fails, Restrict wins $\geq$ losses against Expand (the probe chooses the remedy) & \raggedright not applicable (P1 held for both models)\tabularnewline
\midrule
\raggedright Wall probe & \raggedright W1 & \raggedright among tasks where the frozen CEM (native argmin) fails and the scripted path succeeds, the scripted path costs less than CEM's cheapest final candidate in more than half (Wall is search-limited, consistent with \DPExp helping there). & \raggedright held (19 of 25)\tabularnewline
\midrule
\raggedright Metaworld decomposition & \raggedright D1 & \raggedright in most CEM J failures CEM's pick is optimistic: real-outcome cost $>$ predicted cost & \raggedright held (68 of 68)\tabularnewline
\raggedright & \raggedright D2 & \raggedright the real-outcome cost separates J success from failure better than the predicted cost (higher AUROC) & \raggedright held (AUROC 0.94 real vs.\ 0.54 predicted)\tabularnewline
\bottomrule
\end{tabular}
\caption{Every prediction in the protocols of \cref{tab:supp-protocols}, with its outcome. Protocols are named and ordered as in \cref{tab:supp-protocols}; labels are those of the protocol files, and some recur across protocols. Predictions are quoted from the files, where J and FULL denote full success, W and OBJ object success (labels such as J1 and W1 are protocol labels), and warm-start or seeded CEM, called DemoProp-Seed or Seed in some protocols, is written warm-start CEM. Outcomes give gains/losses of the first-named arm on paired tasks; two values joined by `and' are for DINO-WM and JEPA-WM (V-JEPA-2-AC on RoboCasa), in this order. \emph{No direction}: the protocol registered the comparison without predicting its direction. The compute protocols number their predictions C1--C3, C5 and C6; there is no C4.}
\label{tab:supp-ledger}
\end{table*}

\begin{table*}[tp]
\ContinuedFloat
\centering\footnotesize
\setlength{\tabcolsep}{3pt}
\begin{tabular}{@{}p{0.130\textwidth}p{0.040\textwidth}p{0.550\textwidth}p{0.240\textwidth}@{}}
\toprule
Protocol & Label & Prediction & Outcome\tabularnewline
\midrule
\raggedright Metaworld seeding, DINO-WM & \raggedright M1 & \raggedright warm-start CEM (mean) does not significantly change full success J against the frozen CEM mean (two-sided exact McNemar $p\geq0.05$) & \raggedright failed ($+45/{-}2$, $p{=}2{\times}10^{-11}$)\tabularnewline
\raggedright & \raggedright M2 & \raggedright frozen \DPRes beats warm-start CEM (mean) on J with one-sided exact McNemar $p<0.05$ & \raggedright failed ($+11/{-}21$)\tabularnewline
\midrule
\raggedright Metaworld seeding, JEPA-WM & \raggedright J1 & \raggedright warm-start CEM (mean) raises full success J over the frozen CEM mean, one-sided exact McNemar $p<0.05$ & \raggedright held ($+52/{-}7$, $p{=}7{\times}10^{-10}$)\tabularnewline
\raggedright & \raggedright J2 & \raggedright warm-start CEM (mean) has at least as many J wins as losses against the frozen \DPRes & \raggedright failed ($+13/{-}22$)\tabularnewline
\midrule
\raggedright Wall seeding & \raggedright S1 & \raggedright warm-start CEM (mean) has more successes than the frozen CEM mean, with more wins than losses & \raggedright held (59 vs.\ 39; $+22/{-}2$)\tabularnewline
\midrule
\raggedright RoboCasa windows & \raggedright R1 & \raggedright per model, report the share of CEM-mean FULL failures in which the known solution is cheaper than every evaluated candidate (search failures); no direction is predicted & \raggedright no direction (0 of 15 and 0 of 10 are search failures)\tabularnewline
\raggedright & \raggedright R2 & \raggedright conditional, per model: if search failures are the majority, Expand has at least as many FULL wins as losses against CEM mean and against Restrict; if scoring failures are the majority, Restrict has at least as many FULL wins as losses against Expand & \raggedright held for both models (majority scoring failures; Restrict vs.\ Expand $+8/{-}1$ and $+4/{-}2$)\tabularnewline
\raggedright & \raggedright R3 & \raggedright per model, \DPSeed (mean) has at least as many FULL wins as losses against CEM mean & \raggedright held for DINO-WM ($+4/{-}2$), failed for V-JEPA-2-AC ($+3/{-}6$)\tabularnewline
\midrule
\raggedright RoboCasa objective & \raggedright P1 & \raggedright pooled over the two models: CEM mean with the new objective has more FULL wins than losses against the official CEM mean (direction only; the exact two-sided $p$ is reported, no threshold is claimed in advance: CEM already solves 75 and 80 of 90, so at most 25 pairs can be won) & \raggedright failed ($+8/{-}9$)\tabularnewline
\raggedright & \raggedright P2 & \raggedright pooled: warm-start CEM (mean) with the new objective (windows in the first population of CEM with the new objective) has more FULL wins than losses against the official CEM mean & \raggedright held ($+11/{-}6$, $p{=}0.33$)\tabularnewline
\raggedright & \raggedright P3 & \raggedright per model: CEM mean with the new objective has at least as many FULL wins as losses against the official CEM mean & \raggedright failed for DINO-WM ($+3/{-}4$), held for V-JEPA-2-AC ($+5/{-}5$)\tabularnewline
\raggedright & \raggedright P4 & \raggedright no direction predicted: Expand and Restrict with the new objective against their official-objective counterparts; the share of search failures (probe under the new objective) among the failures of CEM mean with the new objective; the null plan as a reference & \raggedright no direction (Restrict $+3/{-}0$ and $+2/{-}0$; Expand $+5/{-}4$ and $+4/{-}5$; search failures 0 of 16 and 0 of 10; null plan 69 and 69)\tabularnewline
\midrule
\raggedright RoboCasa adaptation, DINO-WM & \raggedright A1 & \raggedright CEM mean with the adapted model has more FULL wins than losses against the zero-shot CEM mean (direction; exact $p$ reported) & \raggedright held ($+10/{-}3$)\tabularnewline
\raggedright & \raggedright A2 & \raggedright warm-start CEM (mean) with the adapted model has more FULL wins than losses against the zero-shot CEM mean & \raggedright held ($+11/{-}1$)\tabularnewline
\raggedright & \raggedright A3 & \raggedright on the 21 tasks where standing still fails (stratum fixed by the null plan in RCASA\_\allowbreak REPAIR\_\allowbreak PROTOCOL\_\allowbreak v1), CEM mean with the adapted model has at least as many FULL wins as losses against the zero-shot CEM mean & \raggedright held (15 vs.\ 13; $+4/{-}2$)\tabularnewline
\raggedright & \raggedright A4 & \raggedright no direction predicted: the adapted arms against the change-relative arms of RCASA\_\allowbreak REPAIR\_\allowbreak PROTOCOL\_\allowbreak v1 (CEM, Restrict, warm start), and the share of search failures under the adapted model & \raggedright no direction (CEM $+12/{-}4$, warm start $+5/{-}3$, Restrict $+4/{-}2$; 1 search failure among 8 CEM failures)\tabularnewline
\midrule
\raggedright RoboCasa adaptation, V-JEPA-2-AC & \raggedright V1 & \raggedright CEM mean with the adapted model has more FULL wins than losses against the zero-shot CEM mean (direction; exact $p$ reported) & \raggedright held ($+7/{-}3$)\tabularnewline
\raggedright & \raggedright V2 & \raggedright warm-start CEM (mean) with the adapted model has more FULL wins than losses against the zero-shot CEM mean & \raggedright held ($+7/{-}3$)\tabularnewline
\raggedright & \raggedright V3 & \raggedright on the 21 tasks where standing still fails (stratum fixed by the null plan in RCASA\_\allowbreak REPAIR\_\allowbreak PROTOCOL\_\allowbreak v1), CEM mean with the adapted model has at least as many FULL wins as losses against the zero-shot CEM mean & \raggedright held (16 vs.\ 12; $+6/{-}2$)\tabularnewline
\raggedright & \raggedright V4 & \raggedright no direction predicted: the adapted arms against the change-relative arms of RCASA\_\allowbreak REPAIR\_\allowbreak PROTOCOL\_\allowbreak v1 (CEM, Restrict, warm start), and the share of search failures under the adapted model & \raggedright no direction (CEM $+6/{-}2$, warm start $+10/{-}3$, Restrict $+2/{-}3$; 0 search failures among 6 CEM failures)\tabularnewline
\bottomrule
\end{tabular}
\caption{Every prediction in the protocols of \cref{tab:supp-protocols}, with its outcome (continued).}
\end{table*}

\begin{table*}[tp]
\ContinuedFloat
\centering\footnotesize
\setlength{\tabcolsep}{3pt}
\begin{tabular}{@{}p{0.130\textwidth}p{0.040\textwidth}p{0.550\textwidth}p{0.240\textwidth}@{}}
\toprule
Protocol & Label & Prediction & Outcome\tabularnewline
\midrule
\raggedright Goal information, start key & \raggedright G1 & \raggedright PushT, pooled over both models, J: start-key Expand wins more than it loses against CEM (official output) & \raggedright held ($+7/{-}0$, $p{=}0.02$)\tabularnewline
\raggedright & \raggedright G2 & \raggedright PushT, pooled, W: start-key Expand wins more than it loses against CEM & \raggedright held ($+6/{-}0$, $p{=}0.03$)\tabularnewline
\raggedright & \raggedright G3 & \raggedright Metaworld, each model, J: start-key Restrict wins more than it loses against CEM (official output) & \raggedright held for both models ($+41/{-}15$ and $+50/{-}8$)\tabularnewline
\raggedright & \raggedright G4 & \raggedright Metaworld, pooled, J: start-key Restrict wins more than it loses against the start-key nearest window & \raggedright held ($+104/{-}8$)\tabularnewline
\raggedright & \raggedright G5 & \raggedright RoboCasa, pooled, FULL: start-key Restrict with the adapted predictor wins more than it loses against start-key Restrict with the zero-shot predictor & \raggedright held ($+19/{-}1$)\tabularnewline
\raggedright & \raggedright G6 & \raggedright RoboCasa, pooled, FULL: start-key Restrict with the adapted predictor wins more than it loses against zero-shot CEM & \raggedright held ($+17/{-}3$)\tabularnewline
\raggedright & \raggedright G7 & \raggedright no direction predicted: every start-key arm against the frozen full-key arm of the same mode, per model (the share of each gain that the goal terms of the key account for) & \raggedright no direction (full vs.\ start key, \J: PushT Expand $+13/{-}0$ and $+10/{-}0$; Metaworld Restrict $+16/{-}9$ and $+19/{-}7$)\tabularnewline
\midrule
\raggedright Goal information, endpoint CEM & \raggedright E1 & \raggedright pooled over both models, J: the frozen \DPExp (official CEM) wins more than it loses against the endpoint-CEM mean & \raggedright held in direction only ($+21/{-}16$, $p{=}0.51$)\tabularnewline
\raggedright & \raggedright E2 & \raggedright pooled, W: the frozen \DPExp wins more than it loses against the endpoint-CEM mean & \raggedright held ($+29/{-}10$, $p{=}0.003$)\tabularnewline
\raggedright & \raggedright E3 & \raggedright pooled, J: \DPExp on the endpoint-CEM final population wins more than it loses against the endpoint-CEM mean & \raggedright held ($+10/{-}1$, $p{=}0.01$)\tabularnewline
\raggedright & \raggedright E4 & \raggedright no direction predicted: endpoint-CEM mean against the official CEM mean, J and W, per model & \raggedright no direction (\J $+22/{-}9$ and $+16/{-}4$; \W $+12/{-}16$ and $+10/{-}10$)\tabularnewline
\midrule
\raggedright Compute, PushT & \raggedright C5 & \raggedright each model: the real-outcome cost of CEM's cheapest final candidate is lower at 4x than at 1x in more than half of the 96 episodes & \raggedright held for both models (56 and 53 of 96)\tabularnewline
\midrule
\raggedright Compute, PushT addendum & \raggedright C6 & \raggedright each model: across the 96 official episodes, the Spearman correlation between the 1x$\to$4x change in the predicted cost of CEM's cheapest final candidate and the change in its real-outcome cost is above 0.2 (in the search regime predicted improvements carry over to the real outcome) & \raggedright held for both models ($\rho{=}0.51$ and $0.24$)\tabularnewline
\midrule
\raggedright Compute, Metaworld & \raggedright C1 & \raggedright each model: the predicted cost of CEM's cheapest final candidate is lower at 4x than at 1x in more than half of the 112 tasks & \raggedright held for both models (108 and 108 of 112)\tabularnewline
\raggedright & \raggedright C2 & \raggedright each model: across the 112 tasks, the Spearman correlation between the 1x$\to$4x change in the predicted cost of CEM's cheapest final candidate and the change in its real-outcome cost is below 0.2 (a predicted improvement does not carry over to the real outcome; on PushT, the search regime, SCALE\_\allowbreak PUSHT\_\allowbreak ADDENDUM\_\allowbreak v1 predicts a correlation above 0.2) & \raggedright held for both models ($\rho{=}{-}0.10$ and ${-}0.02$)\tabularnewline
\raggedright & \raggedright C3 & \raggedright each model: full success of the CEM mean (official output) rises by at most 3 net tasks at 4x (gains minus losses $\leq$ 3) & \raggedright held for DINO-WM ($+11/{-}13$), failed for JEPA-WM ($+15/{-}8$)\tabularnewline
\bottomrule
\end{tabular}
\caption{Every prediction in the protocols of \cref{tab:supp-protocols}, with its outcome (continued).}
\end{table*}

\begin{table*}[tp]
\ContinuedFloat
\centering\footnotesize
\setlength{\tabcolsep}{3pt}
\begin{tabular}{@{}p{0.130\textwidth}p{0.040\textwidth}p{0.550\textwidth}p{0.240\textwidth}@{}}
\toprule
Protocol & Label & Prediction & Outcome\tabularnewline
\midrule
\raggedright iCEM, PushT & \raggedright I1 & \raggedright each model: the predicted cost of iCEM's output is lower than that of CEM's cheapest final candidate in more than half of the 96 episodes & \raggedright held for JEPA-WM (51 of 96), failed for DINO-WM (46 of 96)\tabularnewline
\raggedright & \raggedright I2 & \raggedright each model: \DPExp on iCEM wins more full successes than it loses against iCEM & \raggedright held for both models ($+11/{-}0$ and $+9/{-}0$)\tabularnewline
\raggedright & \raggedright I3 & \raggedright each model: \DPExp on iCEM loses at most one full success of iCEM & \raggedright held for both models (no loss)\tabularnewline
\raggedright & \raggedright -- & \raggedright reported without direction: iCEM against CEM (official output) and against \DPExp on CEM, full and object success; probe labels of iCEM's failures (known solution cheaper than every iCEM candidate) & \raggedright no direction (\J: iCEM vs.\ CEM $+13/{-}4$ and $+12/{-}2$, vs.\ \DPExp on CEM $+7/{-}14$ and $+6/{-}10$; \W: $+8/{-}2$ and $+10/{-}0$, $+5/{-}6$ and $+5/{-}3$; search failures 19 of 25 and 14 of 18)\tabularnewline
\midrule
\raggedright Image key, PushT & \raggedright V1 & \raggedright each model: visual-key \DPExp wins more full successes than it loses against CEM (official output) & \raggedright held for both models ($+11/{-}1$ and $+10/{-}0$)\tabularnewline
\raggedright & \raggedright V2 & \raggedright each model: visual-key \DPExp loses at most one full success of CEM & \raggedright held for both models (1 and 0 losses)\tabularnewline
\raggedright & \raggedright V3 & \raggedright pooled over both models: visual-key \DPExp wins more object successes than it loses against CEM & \raggedright held ($+15/{-}0$, $p{=}6{\times}10^{-5}$)\tabularnewline
\raggedright & \raggedright -- & \raggedright reported without direction: visual-key against proprioceptive-key \DPExp; visual-key \DPRes and nearest window; proposals selected; repairs on probe-labelled search failures & \raggedright no direction (\J\ against the proprioceptive key $+2/{-}8$ and $+0/{-}4$; \DPRes 42 and 41, nearest window 14 and 19; selected 18 and 15; repairs on search failures 11 of 11 and 10 of 10)\tabularnewline
\bottomrule
\end{tabular}
\caption{Every prediction in the protocols of \cref{tab:supp-protocols}, with its outcome (continued).}
\end{table*}

\begin{table*}[tp]
\ContinuedFloat
\centering\footnotesize
\setlength{\tabcolsep}{3pt}
\begin{tabular}{@{}p{0.130\textwidth}p{0.040\textwidth}p{0.550\textwidth}p{0.240\textwidth}@{}}
\toprule
Protocol & Label & Prediction & Outcome\tabularnewline
\midrule
\raggedright Metaworld, new tasks & \raggedright F1 & \raggedright each model: \DPRes with the all-task index wins more full successes than it loses against warm-start CEM with the task windows (mean output, the registered baseline) & \raggedright held for both models ($+23/{-}14$ and $+31/{-}18$)\tabularnewline
\raggedright & \raggedright F2 & \raggedright pooled over both models: \DPRes with the all-task index beats that warm start in full success, one-sided exact McNemar $p<0.05$ & \raggedright held ($+54/{-}32$, $p{=}0.011$)\tabularnewline
\raggedright & \raggedright F3 & \raggedright pooled: it wins at least as many full successes as it loses against warm-start CEM with the all-task windows (mean output) & \raggedright held ($+66/{-}30$)\tabularnewline
\raggedright & \raggedright -- & \raggedright reported without direction: object success of all arms; cheapest-final-candidate outputs; \DPRes with the task index; CEM; paired counts per task family; sign-flip over the seven families & \raggedright no direction (object success 202 vs.\ 215; families $p{=}0.27$; \cref{sec:supp-sprint})\tabularnewline
\midrule
\raggedright Metaworld, 42 families & \raggedright G1 & \raggedright primary, unit = task family: the full-success difference \DPRes (all-task index) minus warm start, summed over a family's 4 tasks and both models; the sum over the 42 families is positive, exact one-sided sign-flip $p<0.05$ & \raggedright failed ($-9$; 18 vs.\ 17 families; $p{=}0.71$)\tabularnewline
\raggedright & \raggedright G2 & \raggedright each model: \DPRes (all-task index) wins more full successes than it loses against the warm start & \raggedright held for DINO-WM ($+30/{-}28$), failed for JEPA-WM ($+25/{-}36$)\tabularnewline
\raggedright & \raggedright G3 & \raggedright more families with a positive than with a negative difference & \raggedright held (18 vs.\ 17)\tabularnewline
\raggedright & \raggedright -- & \raggedright reported without direction: the 35 families not used before; against the nearest all-task window; \DPRes with the task index; Expand; probe labels; object success & \raggedright no direction (35 families $p{=}0.78$; nearest window $+35/{-}35$, $+24/{-}33$; \cref{sec:supp-sprint})\tabularnewline
\midrule
\raggedright Metaworld, averaging & \raggedright H1 & \raggedright primary, unit = task family: the full-success difference \DPAvg minus each model's warm start, summed over a family's 4 tasks and both models; the sum over the 42 families is positive, exact one-sided sign-flip $p<0.05$ & \raggedright held ($+52$; 24 vs.\ 11 families; $p{=}0.014$)\tabularnewline
\raggedright & \raggedright H2 & \raggedright each model: \DPAvg wins more full successes than it loses against the warm start & \raggedright held for both models ($+47/{-}20$, $+45/{-}20$)\tabularnewline
\raggedright & \raggedright H3 & \raggedright more families with a positive than with a negative difference & \raggedright held (24 vs.\ 11)\tabularnewline
\raggedright & \raggedright -- & \raggedright reported without direction: other averages, the nearest window, \DPRes over all tasks, object success, the 35 families not used before & \raggedright no direction (35 families $p{=}0.024$; \cref{tab:supp-avg})\tabularnewline
\midrule
\raggedright Cross-environment & \raggedright X1 & \raggedright PushT, each model: averaging the 30 nearest windows solves at most 90\% of the warm start's full successes & \raggedright held (44 vs.\ 80 and 79)\tabularnewline
\raggedright & \raggedright X2 & \raggedright Wall: the same & \raggedright held (3 vs.\ 59)\tabularnewline
\raggedright & \raggedright X3 & \raggedright RoboCasa, each model: averaging solves at least as many full successes as the warm start & \raggedright held (88 vs.\ 77 and 77)\tabularnewline
\raggedright & \raggedright X4 & \raggedright summed over the seven setting--model pairs, following the diagnosis (warm start in PushT and Wall, averaging in Metaworld and RoboCasa) solves more than always warm-starting and than always averaging & \raggedright held (648 vs.\ 574 and 521)\tabularnewline
\midrule
\raggedright Within PushT & \raggedright R2 & \raggedright each model: the warm start beats averaging at full precision, and averaging beats it at 3 bits & \raggedright held (\FlipDfull{} and \FlipJfull{} vs.\ 53; 53 vs.\ 0 and 1)\tabularnewline
\raggedright & \raggedright R3 & \raggedright in no setting is the rule's use significantly worse than the other (exact McNemar), and it is at least as good in at least 8 of 10 & \raggedright held (10 of 10, none worse)\tabularnewline
\raggedright & \raggedright R4 & \raggedright summed over the ten settings, the rule solves more than always warm-starting and than always averaging & \raggedright held (\FlipRule{} vs.\ \FlipWarm{} and \FlipAvg{})\tabularnewline
\raggedright & \raggedright R5 & \raggedright direction: averaging's gain rises with the median rank of the known solution among the windows & \raggedright held (Spearman \FlipRho{})\tabularnewline
\midrule
\raggedright BC reference & \raggedright -- & \raggedright reference line (goal-conditioned policy trained on the demonstration windows); no prediction & \raggedright reported (\BCptFull{} of 96 PushT episodes, \BCmwFull{} of 168 Metaworld tasks; \cref{sec:supp-bc})\tabularnewline
\bottomrule
\end{tabular}
\caption{Every prediction in the protocols of \cref{tab:supp-protocols}, with its outcome (continued).}
\end{table*}

\begin{table*}[tp]
\ContinuedFloat
\centering\footnotesize
\setlength{\tabcolsep}{3pt}
\begin{tabular}{@{}p{0.130\textwidth}p{0.040\textwidth}p{0.550\textwidth}p{0.240\textwidth}@{}}
\toprule
Protocol & Label & Prediction & Outcome\tabularnewline
\midrule
\raggedright Start-key averaging & \raggedright -- & \raggedright report only: \DPAvg{} with the start term of the key, its nearest window and the task's own demonstrations; no prediction & \raggedright reported (\cref{sec:supp-goalinfo})\tabularnewline
\midrule
\raggedright Matched controls & \raggedright M1 & \raggedright primary, unit = task family: the full-success difference \DPAvg minus warm-start CEM seeded with the same 30 windows (W30), summed over a family's 4 tasks and both models; the sum over the 42 families is positive, exact one-sided sign-flip $p<0.05$ & \raggedright held ($+73$; 28 vs.\ 5 families; $p{=}0.0002$)\tabularnewline
\raggedright  & \raggedright M2 & \raggedright each model: \DPAvg wins more full successes than it loses against W30 & \raggedright held for both models ($+46/{-}11$, $+52/{-}14$)\tabularnewline
\raggedright  & \raggedright M3 & \raggedright each model: \DPAvg wins more than it loses against the same warm start with the 100 nearest all-task windows & \raggedright held for both models ($+52/{-}16$, $+50/{-}12$)\tabularnewline
\raggedright  & \raggedright M4 & \raggedright each model: \DPAvg wins more than it loses against the window of the 30 that the native cost ranks first on the model's prediction & \raggedright held for both models ($+48/{-}5$, $+47/{-}7$)\tabularnewline
\raggedright  & \raggedright M5 & \raggedright each model: the window of the 30 that the same cost ranks first on its real outcome wins more than it loses against the model's choice & \raggedright held for both models ($+52/{-}2$, $+49/{-}1$)\tabularnewline
\raggedright  & \raggedright -- & \raggedright reported without direction: object success; the real-outcome choice against \DPAvg; any of the 30; the model's choice among 100 windows; the registered warm start & \raggedright no direction (real-outcome choice vs.\ \DPAvg $+19/{-}12$, $+19/{-}11$; \cref{tab:supp-matched})\tabularnewline
\midrule
\raggedright Failure-level labels & \raggedright P1 & \raggedright primary: of the CEM failures on the 96 new episodes (8 settings above 3 bits) that exactly one of the warm start and \DPAvg{} solves, the label's use (warm start for a search label, \DPAvg{} for a scoring label) solves more than labels permuted within each setting give, one-sided $p<0.05$ & \raggedright held ($151$ of $188$ against $135.9$; $p{<}10^{-4}$)\tabularnewline
\raggedright  & \raggedright P2a & \raggedright pooled, among search-labeled such failures the warm start solves more than \DPAvg{} & \raggedright held ($71$ vs.\ $5$)\tabularnewline
\raggedright  & \raggedright P2b & \raggedright pooled, among scoring-labeled such failures \DPAvg{} solves more than the warm start & \raggedright held ($80$ vs.\ $32$)\tabularnewline
\raggedright  & \raggedright -- & \raggedright reported without direction: per setting and per model; per-model $p$; $T$ minus its null mean with a goal-trajectory interval; the label's use against the rule's use among CEM's failures; agreement of \DPRes{} with the frozen records & \raggedright no direction (DINO-WM $p{<}0.001$, JEPA-WM $p{=}0.29$; label's use 277, rule's use 287 of 446; \cref{tab:supp-labelpred})\tabularnewline
\bottomrule
\end{tabular}
\caption{Every prediction in the protocols of \cref{tab:supp-protocols}, with its outcome (continued).}
\end{table*}

\begin{table*}[tp]
\ContinuedFloat
\centering\footnotesize
\setlength{\tabcolsep}{3pt}
\begin{tabular}{@{}p{0.130\textwidth}p{0.060\textwidth}p{0.530\textwidth}p{0.240\textwidth}@{}}
\toprule
Protocol & Label & Pass criterion & Outcome\tabularnewline
\midrule
\raggedright Method (DINO-WM) & \raggedright decision.\allowbreak pass & \raggedright Neither arm's W nor J degrades against either strong reference, and pooled W or J gain over both references is positive. & \raggedright held (H1: \W $+10/{-}0$, \J $+6/{-}0$)\tabularnewline
\midrule
\raggedright H2 confirmation & \raggedright decision.\allowbreak pass & \raggedright Neither arm's W nor J degrades against either strong reference, and pooled W or J gain over both references is positive (same gate as for H1). & \raggedright held (\W $+2/{-}0$, \J $+2/{-}0$)\tabularnewline
\midrule
\raggedright JEPA-WM replication & \raggedright decision.\allowbreak pass & \raggedright Pooled over the 65 tasks: neither arm's W nor J degrades against either strong reference, and pooled W or J gain is positive. & \raggedright held (\W $+15/{-}0$, \J $+10/{-}0$)\tabularnewline
\midrule
\raggedright Closed-loop MPC & \raggedright decision.\allowbreak pass & \raggedright Pooled over 65 tasks: method W and J not below the MPC baseline and pooled W or J gain positive. & \raggedright held (\W $+10/{-}0$, \J $+12/{-}0$)\tabularnewline
\midrule
\raggedright Search baselines & \raggedright claim\_\allowbreak rule & \raggedright The paper may state that the method is at least as successful as best-of-$K$ CEM only if the method's pooled W and J successes are both $\geq$ those of best-of-$K$; otherwise it reports the shortfall. & \raggedright held (best-of-4: \W 60 vs.\ 57, \J 52 vs.\ 51; best-of-2: 60 vs.\ 56, 52 vs.\ 48)\tabularnewline
\midrule
\raggedright Wall & \raggedright decision.\allowbreak pass & \raggedright Method success not below either reference and positive gain over both. & \raggedright held (47 vs.\ 39 for both references; $+10/{-}2$ vs.\ CEM mean, $+8/{-}0$ vs.\ native argmin)\tabularnewline
\midrule
\raggedright Planner seeds & \raggedright -- & \raggedright No predictions: CEM and \DPExp per seed and their range are reported. & \raggedright reported (\cref{sec:supp-seeds})\tabularnewline
\bottomrule
\end{tabular}
\caption{Every prediction in the protocols of \cref{tab:supp-protocols}, with its outcome (continued): the first six protocols of \cref{tab:supp-protocols} state no separate predictions; this block lists their pass criteria and claim rule, and the planner-seed protocol, which states neither.}
\end{table*}